\documentclass{article}
\usepackage[table,dvipsnames]{xcolor}

\PassOptionsToPackage{numbers}{natbib}
\usepackage[preprint]{neurips_2026}

\usepackage[utf8]{inputenc} 
\usepackage[T1]{fontenc}    
\usepackage{hyperref}       
\usepackage{url}            
\usepackage{booktabs}       
\usepackage{amsfonts}       
\usepackage{nicefrac}       
\usepackage{microtype}      
\usepackage{xcolor}

\usepackage{amsmath}
\usepackage{amssymb}
\usepackage{mathtools}
\usepackage{amsthm}

\usepackage{graphicx}
\usepackage{float}
\usepackage{algorithm}
\usepackage{algpseudocode}

\usepackage{bm} 

\usepackage[capitalize,noabbrev]{cleveref}
\usepackage{amsmath}
\usepackage{amssymb}
\usepackage{mathtools}
\usepackage{physics}

\usepackage{makecell}
\usepackage{xparse}
\usepackage{dsfont}

\theoremstyle{plain}
\newtheorem{theorem}{Theorem}
\newtheorem{proposition}[theorem]{Proposition}
\newtheorem{lemma}[theorem]{Lemma}

\theoremstyle{definition}

\theoremstyle{remark}

\newcommand{\best}[1]{\cellcolor{green!50}\ensuremath{\bm{#1}}}  %
\newcommand{\almostbest}[1]{\cellcolor{green!25}\ensuremath{\bm{#1}}}

\newcommand*\diff{\mathop{}\!\mathrm{d}}
\newcommand{\distreq}{\stackrel{\rm distr}{=}}

\usepackage{amsthm}
\theoremstyle{plain}
\newtheorem*{rep@theorem}{\rep@title}
\newcommand{\newreptheorem}[2]{%
	\newenvironment{rep#1}[1]{%
		\def\rep@title{#2 \ref{##1} (restated)}%
		\begin{rep@theorem}}%
		{\end{rep@theorem}}}
\newreptheorem{theorem}{Theorem}

\newtheorem*{rep@lemma}{\rep@title}
\newcommand{\newreplemma}[2]{%
	\newenvironment{rep#1}[1]{%
		\def\rep@title{#2 \ref{##1} (restated)}%
		\begin{rep@lemma}}%
		{\end{rep@lemma}}}
\newreplemma{lemma}{Lemma}

	\newcommand{\R}{\mathbb{R}}

	\newcommand{\N}{\mathbb{N}}
	
\NewDocumentCommand{\1}{g}{%
		\ensuremath{\mathds{1}%
		\IfNoValueTF{#1}
			{}
			{_{#1}}
		}%
	}

\renewcommand{\d}{\ensuremath{\mathrm{d}}}
	
	\RenewDocumentCommand{\Pr}{g}{%
		\ensuremath{\operatorname{\mathbb{P}}%
		\IfNoValueTF{#1}
			{}
			{\left[ #1 \right]}
		}%
	}
	\NewDocumentCommand{\Prhat}{g}{%
		\ensuremath{\hat{\operatorname{\mathbb{P}}}%
		\IfNoValueTF{#1}
			{}
			{\left[ #1 \right]}
		}%
	}

	\NewDocumentCommand{\E}{m g}{%
		\ensuremath{\operatorname{\mathbb{E}}%
		\IfNoValueTF{#2}
			{}
			{_{#1}}
		\left[%
		\IfNoValueTF{#2}
			{#1}
			{#2}
		\right]%
		}%
	}

	\NewDocumentCommand{\Var}{m g}{%
		\ensuremath{\operatorname{Var}%
		\IfNoValueTF{#2}
			{}
			{_{#1}}
		\left[%
		\IfNoValueTF{#2}
			{#1}
			{#2}
		\right]%
		}%
	}

	\NewDocumentCommand{\Cov}{m g}{%
		\ensuremath{\operatorname{Cov}%
		\IfNoValueTF{#2}
			{}
			{_{#1}}
		\left(%
		\IfNoValueTF{#2}
			{#1}
			{#2}
		\right)%
		}%
	}

	\newcommand{\iid}{i.i.d\@ifnextchar.{}{.\@\xspace}}

	\newcommand{\Bernoulli}[1]{\operatorname{Bernoulli}\left( #1 \right)}

	\NewDocumentCommand{\Normal}{m  m g}{%
	\ensuremath{\operatorname{\mathcal{N}}%
		\IfNoValueTF{#3}
			{}
			{_{#1}}
		\left(%
		\IfNoValueTF{#3}
			{#1,\, #2}
			{#2,\, #3}
		\right)%
		}%
		}

\newcommand{\Oh}[1]{\ensuremath{\mathcal{O}\left(#1\right)}} 

\newcommand{\Th}[1]{\ensuremath{\Theta\left(#1\right)}}

\newcommand{\mc}{\mathcal}

\newcommand{\lsp}{\mkern-1mu}

\title{Permutation-Invariant Spectral Learning via Dyson Diffusion}

\author{%
Tassilo Schwarz$^{1,2}$ \quad Cai Dieball$^{2}$ \quad Constantin Kogler$^{3}$
\quad Renaud Lambiotte$^{1}$\\
\textbf{Arnaud Doucet${^4}$ \quad Aljaz Godec$^{2,5}$ \quad George Deligiannidis$^{4}$}\\
$^1$Mathematical Institute, University of Oxford \\
$^2$Mathematical bioPhysics Group, Max Planck Institute for Multidisciplinary Sciences \\
$^3$School of Mathematics, Institute for Advanced Study \\
$^4$Department of Statistics, University of Oxford \\
$^5$Mathematical Physics and Stochastic Dynamics, University of Freiburg\\
\texttt{tassilo.schwarz@maths.ox.ac.uk, deligian@stats.ox.ac.uk}
}

\begin{document}

\maketitle

\begin{abstract}
  Diffusion models are central to generative modeling and have been adapted to graphs by diffusing adjacency matrix representations. The challenge of having up to $n!$ such representations for graphs with $n$ nodes is only partially mitigated by using permutation-equivariant learning architectures. Despite their computational efficiency, existing graph diffusion models struggle to distinguish certain graph families and their spectra, unless graph data are augmented with ad hoc features.
This shortcoming stems from enforcing the inductive bias within the learning architecture.
In this work, we leverage random matrix theory to analytically extract the spectral properties of the diffusion process, allowing us to push most of the inductive bias from the architecture into the dynamics. Building on this, we introduce the Dyson Diffusion Model, which employs Dyson's Brownian motion to capture the spectral dynamics of an Ornstein-Uhlenbeck process on the adjacency matrix.
Furthermore, conditioned on the spectral dynamics, we formulate a Lie group diffusion, appropriately modeling the remaining degrees of freedom. Strikingly, the resulting learning problem becomes permutation invariant at the Lie algebra level.
We demonstrate that the Dyson Diffusion Model learns graph spectra accurately and outperforms existing graph diffusion models.
\end{abstract}

\section{Introduction}\label{sec:introduction}

Diffusion models are a key class of generative models based on noising data with stochastic differential equations (SDEs) and learning their time reversal \citep{sohl-dicksteinDeepUnsupervisedLearning2015, songScoreBasedGenerativeModeling2021, hoDenoisingDiffusionProbabilistic2020}. They provide state-of-the-art results in many domains such as audio \citep{zhangSurveyAudioDiffusion2023} and vision \citep{croitoruDiffusionModelsVision2023}. Generalizing diffusion models from Euclidean space to graphs offers promising applications in numerous areas, such as %
biology \citep{watsonNovoDesignProtein2023} or combinatorial optimization \citep{sunDIFUSCOGraphbasedDiffusion2023}. However, while diffusing adjacency matrix representations is straightforward and popular \citep{niuPermutationInvariantGraph2020,joScorebasedGenerativeModeling2022,vignacDiGressDiscreteDenoising2022}, this approach faces a major obstacle: Each graph with $n$ vertices has up to $n!$ representations as adjacency matrices. Therefore,
if one aims to use 
a diffusion model on the space of matrices, one must learn
$n!$ %
representations \emph{per graph}.~This is not feasible.
Previous works tackled this problem by shifting the inductive bias of permutation invariance to the learning algorithm: If the neural network was permutation equivariant, training on one of the (up to $n!$ many) matrix representations would suffice. For example, \citet{niuPermutationInvariantGraph2020} and \citet{joScorebasedGenerativeModeling2022} use message-passing graph neural networks (GNNs) while ConGress \citep{vignacDiGressDiscreteDenoising2022} applied graph transformers. However, these learning architectures have a ``blind spot'' detailed below.

\noindent
\textbf{Theoretical Limitations.}~
The ``blind spot'' arises from the limited ability of these models to solve Graph Isomorphism ($\mathrm{GI}$): determining if two graphs are structurally identical regardless of node labeling. While permutation equivariance ensures that the model produces consistent outputs for different representations of the same graph, it does not guarantee that the model can differentiate between two structurally different (non-isomorphic) graphs. The failure is a result of how these architectures aggregate information: permutation-symmetric operations -- message passing in GNNs and self-attention in graph transformers -- can collapse distinct graphs with similar neighborhoods to the same representation, treating them as identical. More formally, $\mathrm{GI}$ is a challenging problem in algorithm theory, and it remains unknown whether $\mathrm{GI} \in \mathsf{P}$ \citep{babaiGraphIsomorphismQuasipolynomial2016}. Since a polynomial time (learning) algorithm perfectly distinguishing all graphs would prove $\mathrm{GI} \in \mathsf{P}$, contemporary (polynomial-time) graph learning algorithms must compromise on expressivity.  

\noindent
\textbf{Extracting Permutation-Invariant Information from Graph Diffusion.}~When diffusing an entire adjacency matrix, state-of-the-art work pushes the entire inductive bias into the learning algorithm \citep{niuPermutationInvariantGraph2020,joScorebasedGenerativeModeling2022} with possible data augmentation \citep{huangGraphGDPGenerativeDiffusion2022, vignacDiGressDiscreteDenoising2022, xuDiscretestateContinuoustimeDiffusion2024}.
However, as we show below, this is neither necessary nor desirable (see the previous discussion on blind spots). Using techniques from random matrix theory, we show that an Ornstein--Uhlenbeck (OU) diffusion on the graph can be \emph{dissected} into diffusion of the (permutation invariant) spectrum and diffusion of the (permutation-dependent) eigenvectors.
Our method therefore allows us to learn the spectrum while preserving all remaining information. Moreover, since the spectrum is inherently permutation-invariant, we can parameterize the score using a much broader range of learning architectures, expanding the scope to architectures able to distinguish between graphs that are equivalent in the Weisfeiler-Leman (WL) sense \citep{morrisWeisfeilerLemanGo2019,xuHowPowerfulAre2018}.\footnote{We give a theoretical discussion in form of the WL-equivalence class in \Cref{sec:theory-and-result} and demonstrate the challenge of those architectures empirically in \Cref{fig:wl-demonstration}.}

\begin{figure*}[t]
  \centering
  \includegraphics[width=\linewidth]{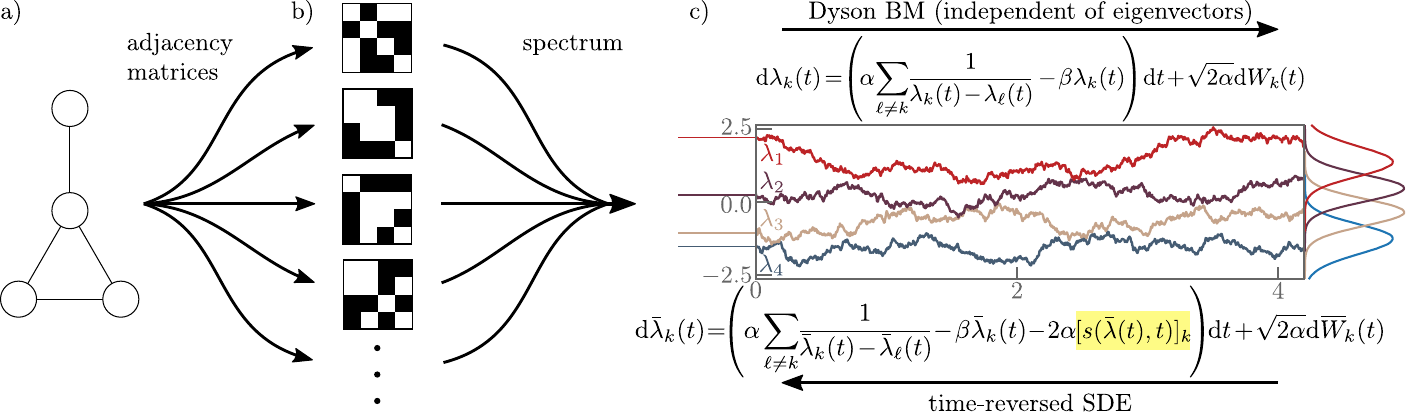}
  \caption{
  Dyson Diffusion Model and its application to graph spectra: A graph on $n$ vertices (a) has up to $n!$ representations as adjacency matrices (b). For an OU-driven diffusion on \emph{any} adjacency matrix, the permutation-invariant spectrum (c) evolves according to the same SDE \eqref{eq:SDE-spectrum}. A sample path of the $n$ non-intersecting eigenvalues is shown. The marginals of the invariant density for the $\lambda_k$ are depicted on the far right. The DyDM diffusion model learns the score $s(\lambda, t)$ (highlighted in yellow) to generate spectra via the time-reversed SDE \eqref{eq:dyson-timereversal}.
  }
  \label{fig:mainfig}
\end{figure*}

\noindent
\textbf{Information in the Spectrum.}~The spectrum of a graph encodes key structural features including connectivity, expansion, and subgraph patterns. Moreover, non-isomorphic WL-equivalent graphs typically have distinct spectra \citep{huangSpectrumRandomDregular2024}. An idea, therefore, is to augment the graph data based on spectral information \citep{vignacDiGressDiscreteDenoising2022,xuDiscretestateContinuoustimeDiffusion2024}. Our work is based on an entirely different method, exploiting analytical insight from random matrix theory to dissect the spectral from the remaining information, allowing to push the inductive bias from the architecture to the dynamics.

\noindent
\textbf{Dyson's Brownian Motion.}~For an OU~process on the space of symmetric matrices, the eigenvalues follow a well-characterized stochastic differential equation, the so-called Dyson Brownian motion (DBM), which is inherently permutation invariant. Thus -- in contrast to \cite{niuPermutationInvariantGraph2020,joScorebasedGenerativeModeling2022,vignacDiGressDiscreteDenoising2022} -- the score of DBM %
can be parameterized with any (not necessarily permutation invariant) neural network. Moreover, contrary to \cite{luoFastGraphGeneration2024}, the remaining information is not lost (see \Cref{thm:eigenvector-SDE}).

\noindent
\textbf{Contributions.}~Our main contributions are:
\begin{itemize}
    \item Based on analytical insight, we demonstrate an inherent struggle of GNN-based graph diffusion models. 
    \item We introduce the Dyson Diffusion Model (DyDM) in \Cref{sec:dy-DM}, extracting the spectral dynamics from an OU-driven diffusion. 
    DyDM learns graph spectra without GNNs or graph transformers; the accompanying conditional eigenvector SDE provides a principled route to modeling the remaining degrees of freedom, which we validate in controlled experiments.
    \item %
    We demonstrate in \Cref{sec:exp} that building on this SDE, DyDM is more effective than existing GNN-based and graph-transformer-based methods for learning graph spectra.
    \item We also derive the eigenvector SDE and its time reversal. Expressing the dynamics on the Lie group $O(n)$—and its corresponding tangential Lie algebra—enables tractable analysis and efficient simulation. We prove a symmetry exponentially reducing the learning space in the dimension and translate the equivariant learning problem for the eigenvectors into an invariant learning problem on the Lie algebra, opening the field to new architectures.
\end{itemize} 
Beyond graphs, our framework generally applies to symmetric matrices where spectral information is often key. For instance, in statistics, it can encode the importance of principal components \citep{jamesIntroductionStatisticalLearning2023}, while in dynamical systems,\footnote{For instance, the generator of Markov dynamics under detailed balance is self-adjoint, see \cite{pavliotisStochasticProcessesApplications2014}.} it reflects the stability and timescales of linear operators.

\textbf{Notation.}~We work on the set of symmetric real matrices %
$\mathrm{Sym}(\R^{n\times n}):= \{A \in \R^{n \times n}: A^T= A\}$.
The positive integers up to $n$ are denoted by $[n]:= \{1, \ldots n\}$.
We consider undirected graphs $G = (V,E,w)$ where $V$ is a finite set with edges $E\subseteq \{S \subseteq V: 1 \leq |S| \leq 2 \}$ allowing for self-loops and weights $w: E \to \R$. 
The family of such graphs of size $n$ is $\mc G^n:= \{G=(V,E,w): |V| = n\}$. For a graph $G \in \mc G^n$ with $\R^V = \{f:V \to \R \}$ being the space of functions from $V$ to $\R$, we %
let $A \equiv A(G): \R^V \to \R^V$ be the adjacency \emph{operator}, defined %
for $f \in \R^V$ and $v \in V$ as the weighted sum of %
$f$ applied to the neighbors of $v$ as $ Af(v) \coloneqq \sum_{\{ u,v \} \in E} w\left(\{ u,v \}\right)f(u).$  
As the graphs are undirected, the adjacency operator is self-adjoint with respect to the standard inner product on $\R^V$. The operator $A$ therefore has $n$ real eigenvalues $\lambda_1 \geq \lambda_2 \geq \ldots \geq \lambda_n$. We denote the ordered spectrum of the adjacency operator by 
$\lambda(G) \coloneqq \lambda(A(G)) \coloneqq \{ (\lambda_1,\lambda_2, \ldots , \lambda_n) \,:\, \lambda_1 \geq \ldots \geq \lambda_n\}$.
Unless defined explicitly otherwise, we use $n\in \N$ for the (vertex) size of the graph, and $N \in \N$ for the number of samples. We denote by $ s \sim \mc N(0,I_d)$ that the $d$-dimensional random vector $ s$ has multivariate normal distribution with $0$ mean and unit covariance $I_d$.

\section{Limitations of Existing Diffusion Models} \label{sec:theory-and-result}

We consider the matrix-valued OU-SDE starting from some data $M\lsp(\lsp0\lsp)\!\in\!\mathrm{Sym}\lsp(\lsp\R^{n\times n}\lsp)$ given by
\begin{align}
    \d M_{ji}(t) = \d M_{ij}(t) = - \beta M_{ij}(t) \d t + D_{ij} \d B_{ij}(t), \quad 1\!\leq\!i\!\leq\!j\!\leq\!n  \label{eq:SDE-mtx} 
\end{align}
with $D_{ij} \coloneqq \sqrt{(1+\delta_{ij}) \alpha }$ for any constants $\alpha, \beta \in \R^+$,
where $B_{ij}(t)=B_{ji}(t)$ are independent Brownian motions 
and $\delta_{ij} = 1$ if and only if $i = j$, and $0$ otherwise. We consider \eqref{eq:SDE-mtx} on the space of symmetric matrices to represent undirected graphs $G \in \mc G^n$.
\Cref{eq:SDE-mtx} is an %
entry-wise OU process that preserves the symmetry of the matrix. Standard diffusion models run \eqref{eq:SDE-mtx} until $T>0$ from data samples. To %
fix notation, denote by $p_t$ the distribution of $M(t)$ induced by \eqref{eq:SDE-mtx} for $t \geq 0$. 

The time-reversal of \eqref{eq:SDE-mtx} on time interval $[0,T]$ is a diffusion initialized from ${M(T)\!\sim\! p_T}$ satisfying
\begin{align}\label{eq:backward}
	\d M_{ij}(t) =&-\left\{\beta M_{ij}(t) + D_{ij}^2 \left[s(M(t),t)\right]_{ij} \right\}dt +D_{ij}\d \bar B_{ij}(t), \quad 1\leq i \leq j \leq n
\end{align}
for independent %
Brownian motions $\bar B_{ij}(t)$ \citep{andersonReversetimeDiffusionEquation1982, songScoreBasedGenerativeModeling2021}. In \eqref{eq:backward}, $s(M,t)$ represents the score matrix at time $t\in (0,T]$, i.e., $s(M,t) \coloneqq \grad_M \ln p_t(M)$.
This score is intractable but, as the OU process has tractable Gaussian transition densities, we can obtain an estimate $s_\theta(M,t)$ of it by minimizing the denoising score matching loss \citep{songScoreBasedGenerativeModeling2021}
\begin{align}
	L{\mkern-1mu}(\theta)\! =\! \mathbb{E}\!\!\left[\!\norm{ s_\theta(M(t),t)\!\! -\!\! \grad_{\! M(t)}\! \ln p_{t|0}(M\!(t)|M\!(0))}^2_2\!\right]\!\!, \label{eq:OU-loss}
\end{align}  
where the expectation is over $t \sim \mathcal{U}[0,T],M(0) \sim p_0, M(t) \sim p_{t \mid 0}(\cdot|M(0))$. Samples of $p_0$ can then be obtained by simulating an approximation of \eqref{eq:backward} obtained by sampling $M(T)$ from the invariant distribution of \eqref{eq:SDE-mtx},
${M_{ij}^{\rm inv} \sim \mc{N}(0,\alpha(1+\delta_{ij})/2\beta)}$, and using $s_\theta(M,t)$ in place of~$s(M,t)$.

\textbf{Challenges posed by graphs.}~When working with graphs, to obtain an adjacency matrix for a given weighted graph $G = (V,E,w) \in \mathcal{G}^n$, we need to fix an ordering $(v_1,\ldots , v_n)$ of all vertices. In fact, given the ordering $(v_1,\ldots , v_n)$, the associated matrix $M = (m_{ij})_{1 \leq i,j \leq n} \in \mathrm{Sym}(\R^{n\times n})$  has entries $m_{ij} = w(\{v_i,v_j\})$ if $\{v_i, v_j\} \in E$ and $0$ otherwise.
The challenge is that the adjacency operator $A(G)$ for a given graph $G \in \mathcal{G}^n$ admits up to $n!$ distinct adjacency matrices. For instance, in \Cref{fig:mainfig} we see four different matrix representations of the same graph.

\noindent
\textbf{Why should we enforce this inductive bias?}~One could think that the reason for using the inductive bias stems from wanting that graph generative models assign uniform probability to each of the (up to) $n!$ many representations. However, one may easily apply a random independent permutation to the output of the generative model.
Instead, the challenge stems from the following problem: In diffusion models, we learn a function (the score) on a set of graphs, say\footnote{Importantly, $\Omega$ refers to \emph{the set of graphs} and \emph{not} the set of (permutation-sensitive) adjacency matrices.} $\Omega$, rather than the distribution directly. Learning on the adjacency representations would correspond to learning on $\Omega\!\times\!S_n$. We demonstrate on a toy example in  \Cref{cor:avg-mse:orders} that \emph{not} leveraging the inductive bias, i.e.\ learning on a space of size $\Omega\!\times\! S_n$, leads to an explosion of the average mean squared error; i.e., learning a function on a fixed number of $k$ objects (say unweighted graphs on $n$ nodes) from $N$ samples yields a mean squared error of order $\Theta(n!/N)$. In contrast, using the inductive bias provides an average mean squared error of $\Theta(1/N)$. Given that $n! > 3 \cdot 10^6$ even for $n\!=\!10$ nodes, utilizing the inductive bias is essential.

\begin{figure}[th!]
    \centering
    \includegraphics[width=1.0\linewidth]{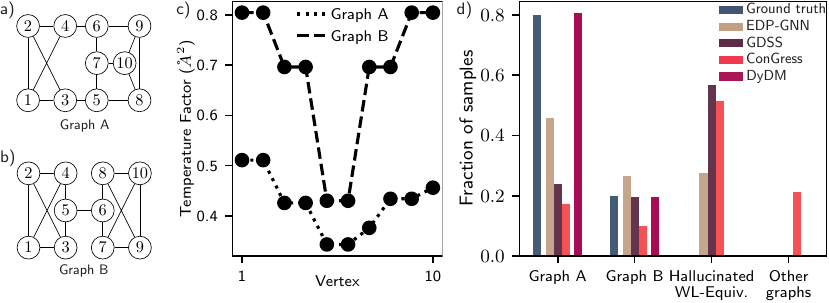} \caption{Struggle of GNN-based and graph-transformer-based models with two WL-equivalent graphs: Graph A and Graph B are WL-equivalent, but non-isomorphic. Also physically, they have very different properties, such as different temperature factors (c) and a different cut size. Upon training on an $80 \%$ Graph A and $20\%$ Graph B dataset, state-of-the-art GNN-based (EDP-GNN, GDSS) and graph-transformer-based (ConGress) models learn the WL-equivalence class quickly but fail to generate the underlying distribution among the two graphs, with some even predominantly hallucinating WL-equivalent but non-isomorphic graphs (d).}
    \label{fig:wl-demonstration}
\end{figure}
\noindent 
\textbf{Cost of pushing inductive bias entirely into architecture.}~One solution would be to impose the inductive bias in the learning architecture. This is what state-of-the-art graph diffusion models do \citep{niuPermutationInvariantGraph2020,joScorebasedGenerativeModeling2022,vignacDiGressDiscreteDenoising2022}.
However, using these architectures comes at a cost.~As argued in the introduction, since~$\mathrm{GI} \in \mathsf{P}$ remains unknown, we expect some limitations. 
In the case of GNNs this compromise has been 
precisely characterized: GNNs cannot distinguish between the large families of %
so-called Weisfeiler-Leman (WL) equivalent graphs \citep{morrisWeisfeilerLemanGo2019,xuHowPowerfulAre2018}. For example, %
all $k$-regular graphs on $n$ vertices for any fixed $k\in \N$ are WL-equivalent (see \Cref{lem:wl-equivalence-regular} below and its proof in App.~\ref{sec:app:wl-k-reg-proof}) and therefore indistinguishable by these architectures. This limitation may lead to hallucination and is also evident in the learned spectra.

 \newcommand{\stateLemmaKReg}{
 For every fixed $n,k \in \mathbb{N}$, all $k$-regular graphs $G \in \mc G^n$ are WL equivalent. Moreover, every graph $G \in \mc G^n$ that is WL equivalent to a $k$-regular graph is $k$-regular.
 }
\begin{lemma}[WL-equivalence of $k$-regular graphs] \label{lem:wl-equivalence-regular}
	\stateLemmaKReg
\end{lemma} 
This is a vast class, since e.g.\ on $n=20$ vertices, there are $510{,}489$ such non-isomorphic, connected 3-regular and thereby WL-equivalent graphs \citep{meringerFastGenerationRegular1999}. 
In particular, we demonstrate on a simple example in \Cref{fig:wl-demonstration} that both GNN- and graph-transformer-based methods fail to learn a distribution on two particular WL-equivalent graphs: during sampling, they either fail to learn the distribution on both WL-equivalent graphs or hallucinate other, WL-equivalent but non-isomorphic, graphs (Fig.~\ref{fig:wl-demonstration}d). This can also be observed during learning: The EDP-GNN model learns the WL-equivalence class quickly (after 500 epochs) but then keeps hallucinating non-isomorphic but WL-equivalent graphs, and in particular struggles to learn the right distribution between graphs $A$ and $B$ for the remaining 4500 epochs (see App.~\ref{sec:app:learning-dynamics} for details). Importantly, those graphs are very different. For instance, if the graphs represented Gaussian Network Models for macromolecules \citep{tirionLargeAmplitudeElastic1996}, physical observables such as the temperature factors in X-ray scattering \citep{halilogluGaussianDynamicsFolded1997} would be clearly distinct, see Fig.~\ref{fig:wl-demonstration}c.
Therefore, a diffusion model based on GNNs will suffer from this expressivity blind spot. More generally, any diffusion model relying on a graph-specific learning algorithm will have 
limited expressivity.

\section{Dyson Diffusion Model} \label{sec:dy-DM}

\subsection{Dyson Brownian Motion}\label{sec:dyson-BM}

\citet{dysonBrownianMotionModelEigenvalues1962} showed that the spectrum of \eqref{eq:SDE-mtx} behaves as $n$ positively charged %
particles in a one-dimensional Coulomb gas. These particles exhibit Brownian motion, but with a pairwise repulsion force proportional to their inverse distance so that their paths do not cross (see Fig.~\ref{fig:mainfig}c). More precisely, Dyson proved that the spectrum of the entry-wise Ornstein--Uhlenbeck process from \eqref{eq:SDE-mtx} follows the SDE given in \Cref{thm:eigenvalue-SDE}. Without loss of generality, we restrict the domain to the \emph{Weyl Chamber} $
 	C_n := \left\{  \lambda \in \R^n: \lambda_1 > \ldots > \lambda_n \right\}.$

\newcommand{\stateThmEigenvalues}{
    Denote by $ \lambda (t) = (\lambda_1(t), \ldots , \lambda_n(t))$ the ordered spectrum of the matrix-valued Ornstein-Uhlenbeck process $M(t)$ of \eqref{eq:SDE-mtx}. Then assuming that the initial matrix $M(0)$ has simple spectrum, $ \lambda(t)$ satisfies for all  $1 \leq k \leq n$ the stochastic differential equation 
        \begin{align}
        \d \lambda_k(t) =
        &\left(\alpha \sum_{\ell \neq k} \frac{1}{\lambda_k(t) - \lambda_\ell(t)}- \beta \lambda_k(t) \right) \d t  + \sqrt{2 \alpha } \d W_k(t),  \tag{Dyson-BM} \label{eq:SDE-spectrum} 
     \end{align}
        for $W_1, \ldots , W_n$ independent standard Brownian motions. Moreover, 
        the unique stationary distribution of \eqref{eq:SDE-spectrum} has density $p_{\mathrm{inv}}( \lambda)  = \exp(-U( \lambda))/Z$ for $U( \lambda) = \frac{\beta}{2 \alpha } \sum_k \lambda_k^2 - \sum_{k<\ell } \ln|\lambda_k-\lambda_\ell|$,  $ \lambda \in C_n$, and $Z$ a normalizing constant so that $p_{\mathrm{inv}}$ corresponds to a probability measure.
 }

 \newcommand{\stateThmEigenvaluesNoTag}{
    Denote by $ \lambda (t) = (\lambda_1(t), \ldots , \lambda_n(t))$ the ordered spectrum of the matrix-valued Ornstein-Uhlenbeck process $M(t)$ of \cref{eq:SDE-mtx}. Then assuming that the initial matrix $M(0)$ has simple spectrum, $ \lambda(t)$ satisfies for all  $1 \leq k \leq n$ the stochastic differential equation 
        \begin{align}
        \d \lambda_k(t) = \left(\alpha \sum_{\ell \neq k} \frac{1}{\lambda_k(t) - \lambda_\ell(t)}- \beta \lambda_k(t) \right) \d t+ \sqrt{2 \alpha } \d W_k(t),  \tag{Dyson-BM}
     \end{align}
        for $W_1, \ldots , W_n$ independent standard Brownian motions. Moreover 
        the unique stationary distribution of \eqref{eq:SDE-spectrum} has density $p_{\mathrm{inv}}( \lambda)  = \exp(-U( \lambda))/Z$ for $U( \lambda) = \frac{\beta}{2 \alpha } \sum_k \lambda_k^2 - \sum_{k<\ell } \ln|\lambda_k-\lambda_\ell|$,  $ \lambda \in C_n$, and $Z$ a normalizing constant so that $p_{\mathrm{inv}}$ corresponds to a probability measure.
 }

\begin{theorem}[Eigenvalue SDE \cite{dysonBrownianMotionModelEigenvalues1962}]\label{thm:eigenvalue-SDE}
   \stateThmEigenvalues
\end{theorem}
For completeness, a full proof of \Cref{thm:eigenvalue-SDE} is given in App.~\ref{sec:app:eigenvalue-SDE-proof}, where we generalize a well-known proof to arbitrary coefficients $\alpha, \beta$. We note that the assumption of \Cref{thm:eigenvalue-SDE} that $M_0$ has simple spectrum is minor. Indeed, generic random graphs or matrices have simple spectra, and in the case of eigenvalues with higher multiplicity, we can perturb the spectrum to be simple. From \eqref{eq:SDE-spectrum} we see that the eigenvalues perform a Brownian motion in a confining potential with a repulsion force: once a pair of eigenvalues $\lambda_k, \lambda_l$ comes too 
close, 
they become repelled with a force $\alpha/ (\lambda_k - \lambda_l)$ inversely proportional to their separation.
A remarkable property of \Cref{thm:eigenvalue-SDE} is that the evolution of the spectrum is decoupled from all other information about the matrix: the spectral SDE (\Cref{thm:eigenvalue-SDE}) is independent of the eigenvectors. This is the key analytical insight that motivates our approach.
Furthermore, conditioned on the eigenvalues, the remaining information captured in form of the eigenvectors can be deduced as we show in \Cref{thm:eigenvector-SDE}. This generalizes a statement of \citet{allezEigenvectorsGaussianMatrices2015}, and we give a proof in App.~\ref{sec:app:eigenvector-SDE-proof}.

\newcommand{\stateThmEigenvectors}{
    Denote by $ (v_1(t), \ldots , v_n(t))$ the orthonormal eigenvectors associated to the eigenvalues of \Cref{thm:eigenvalue-SDE}. Assuming that the initial matrix $M(0)$ has simple spectrum, $ v_k(t)$ satisfies for $ k \in [n]$ the stochastic differential equation 
        \begin{align}
        &\d v_k(t) = -\frac{\alpha}{2} \sum_{\ell \neq k} \frac{v_k(t) }{(\lambda_k(t) - \lambda_\ell(t))^2}\d t  + \sqrt{\alpha} \sum_{\ell \neq k } \frac{v_\ell (t) }{\lambda_k(t) - \lambda_\ell(t)} \d w_{\ell k}(t) \tag{Eigenvector-SDE} \label{eq:SDE-eigenvectors} 
     \end{align}
        for $\{w_{ij: i \neq j}\}$ standard Brownian motions independent of the eigenvalue trajectories, with $w_{ji} = w_{ij}$.
 }

 \newcommand{\stateThmEigenvectorsNoLabel}{
    Denote by $ (v_1(t), \ldots , v_n(t))$ a tuple of orthonormal eigenvectors associated to the eigenvalues of \Cref{thm:eigenvalue-SDE}. Assuming that the initial matrix $M(0)$ has simple spectrum, $ v_k(t)$ satisfies for $ k \in [n]$ the stochastic differential equation 
        \begin{align}
        \d v_k(t) = -\frac{\alpha}{2} \sum_{\ell \neq k} \frac{1}{(\lambda_k(t) - \lambda_\ell(t))^2} v_k(t) \d t  + \sqrt{\alpha} \sum_{\ell \neq k } \frac{1}{\lambda_k(t) - \lambda_\ell(t)} v_\ell (t) \d w_{\ell k}(t) \tag{Eigenvector-SDE}
     \end{align}
        for $\{w_{ij: i \neq j}\}$ standard Brownian motions also independent of the eigenvalue trajectories, with $w_{ji} = w_{ij}$.
 }

 \begin{theorem}[Eigenvector SDE] \label{thm:eigenvector-SDE}
    \stateThmEigenvectors
 \end{theorem}

\subsection{From Dyson SDE to Spectral Diffusion Model}
\label{sec:dyson-sde-to-diffusion-model}

The analytical expression in \Cref{thm:eigenvalue-SDE} allows learning the spectrum of a matrix-diffusion following~\eqref{eq:SDE-mtx}. Despite its advantages, constructing a diffusion model based on \eqref{eq:SDE-spectrum} poses several challenges, e.g.\ dealing with a singular drift, non-Gaussian conditional density, etc. (see App.~\ref{sec:app:challenges-which-we-all-solved} for details). As described below, with DyDM we overcome these obstacles and design an efficient diffusion model for the spectrum, which can distinguish between spectra of graphs that GNNs are blind to (Fig.~\ref{fig:wl-demonstration}) and which does \emph{not} require ad~hoc data augmentation.

 The time-reversal of the \eqref{eq:SDE-spectrum} in the sense of \cite{andersonReversetimeDiffusionEquation1982} reads 
 \begin{align}
 	\d \bar  \lambda_k(t) \!=\!& \Bigg[\!\alpha \!\sum_{\ell  \neq k}\!\frac{1}{\bar \lambda_k(t) \!\!-\!\! \bar \lambda_\ell (t)}\!-\! \beta \bar \lambda_k(t) \!-\! 2\alpha [ s(\bar \lambda(t) ,t)]_k \!\Bigg] \!\d t + \sqrt{2 \alpha } \d \overline W_k(t) \label{eq:dyson-timereversal},
 \end{align}
where we aim to learn the score $s(\lambda ,t) \coloneq \grad_\lambda \ln p_t(\lambda)$. As the coefficients in \eqref{eq:SDE-spectrum} are non-Lipschitz, the applicability of \cite{andersonReversetimeDiffusionEquation1982} is not immediate. Accordingly, in App.~\ref{sec:app:time-reversal} we present a proof of existence and uniqueness of a strong solution and verify that Anderson’s time reversal applies.

\textbf{Making the loss tractable.}~Learning the loss function $s(\lambda ,t)$ as in \eqref{eq:OU-loss} is not feasible for the \eqref{eq:SDE-spectrum}, since the conditional distribution $p_{t \mid 0}$ is not available in closed form in contrast to the OU process. To overcome this, we follow a derivation in the style of \citep{debortoliRiemannianScoreBasedGenerative2022} to obtain the loss function -- up to constants in $\theta$ -- for any $h \in \R^+$
\begin{align}
    \tilde L({\mkern-1mu}\theta{\mkern-1mu})\!\! =\!\!\mathbb{E}\!\!\left[\!\norm{{\mkern-1mu} s_\theta(\lambda_{ t{\mkern-1mu}+{\mkern-1mu}h },\!t\!+\!h) \!\!- \!\!\grad_{\lambda_{t{\mkern-1mu}+{\mkern-1mu}h}}\!\!\ln p_{t{\mkern-1mu}+{\mkern-1mu}h | t}(\lambda_{t{\mkern-1mu}+{\mkern-1mu}h}|\lambda_t){\mkern-1mu}}^2_2\!\right]\!,\! \label{eqn:loss}
\end{align}
the expectation being over $t \sim \mathcal{U}[0,T],\lambda_t \sim p_t, \lambda_{t+h} \sim p_{t+h \mid t}(\cdot|\lambda_t)$. We approximate $p_{t+h\mid t}$ with the Gaussian transition of the Euler--Maruyama step, which is exact for $h\!\to\!0$ (see App.~\ref{sec:app:loss}).

\textbf{Handling singularities.}~Numerical solutions of \eqref{eq:SDE-spectrum} with a fixed step size are not practical, since \eqref{eq:SDE-spectrum} is singular at the boundary of the Weyl Chamber. A fixed step size leads to inaccuracies at the boundaries and may overshoot the singularity, leaving the Weyl Chamber.  To overcome this, we implement
an adaptive step-size algorithm which conditions on an event of probability $1$ (non-crossing) and hence does not change marginal densities.\footnote{The adaptive step-size is only used in the forward simulation. The objective is evaluated on a fixed grid for correctness.} The step-size controller is described in Algorithm \ref{alg:stepsize_fwd} and in App.~\ref{sec:app:stepsize-controller:forward} in detail.

\textbf{Schedule.}~Dyson's conjecture states that for $\alpha\!=\!\frac{1}{n}$ and $\beta\!=\!\frac{1}{2}$  \eqref{eq:SDE-spectrum}
converges to the invariant distribution in time $\Theta(1)$, while the majority of eigenvalues mixes \emph{locally} already in time $\Theta(1/n)$ \citep{yangTopicsRandomMatrix}.
Hence, the fine-grained structure will be mixed in the time interval $\left(0,\Theta(1/n)\right)$. We show in App.~\ref{sec:app:time-rescaling} through time change that this conjecture can be applied to any choice of $\alpha, \beta$. It is thus sensible to choose an exponential schedule. The particular choices are specified in App.~\ref{sec:app:time-grid}.

\newcommand{\dysonTrainingFwd}{
\Require{spectral samples $ \lambda^{(1)}\!\!\!\!\!,\lsp\ldots\lsp, \lsp \lambda^{(N)}\!\in\! \R^n\!\!\!,$ schedule $\mc T\!\!=\!\!\{\lsp t_j\lsp\}$} %
  \For{each sample $i \in [N]$}
  	\State{$t\gets 0$}
  	\State{$ \lambda(t) \gets \lambda^{(i)}$}
  	\While{$t<T$} \Comment{Diffuse sample}
  	\State{Let $ u \sim \mc N(0,I_n)$} 
  	\State{$\delta t \gets\text{ForwardStepsizeController}( \lambda(t), u)$} \Comment{Conditions on non-intersection, see~\ref{sec:app:stepsize-controller:forward}.}
  	\If{$\delta t < \delta t_\text{min}$}
  	\State{Skip step} \Comment{Rare event: Skip tiny steps}  %
  	\EndIf
    \State{$t_\text{fix} = \min \{t_j \in \mc T: t_j > t \}$} 
  	\State{$\delta t \gets \min \{\delta_t,t_\text{fix}-t\}$}
    \State{$
    \begin{aligned}
    \lambda(t\!+\!\delta t)\!\gets{}\!&
    \text{Euler-Maruyama step of \eqref{eq:SDE-spectrum} with step size $\delta t$ and noise $u$} 
    \end{aligned}
    $}
  	\State{$t \gets t + \delta t$}
  	\EndWhile
\EndFor
\State{Update $s_\theta$ using loss $\tilde L(\theta)$ along paths $ \lambda^{(1)}(t), \ldots,  \lambda^{(N)}(t)$ on schedule $\mc T$ using Eq.~\eqref{eq:pathwise-loss}}
 
 \Ensure {$s_\theta$.}
}

\begin{figure}[t]
    \begin{algorithm}[H]
  \caption{DyDM training}
      \label{alg:dyson-training-fwd}
      \begin{algorithmic}[1]
        \dysonTrainingFwd
      \end{algorithmic}
    \end{algorithm}
  \caption{Dyson Diffusion Model (training): The \eqref{eq:SDE-spectrum} is evolved forward in time with an adaptive step size ensuring that the paths remain in the Weyl Chamber. 
  The step-size controller conditions on the probability $1$ event of non-crossing as detailed in App.~\ref{sec:app:stepsize-controller:forward} and \Cref{alg:stepsize_fwd}.}
\end{figure}

\textbf{Equilibrium shooting mechanism.}~In the backward dynamics, imperfect learning may erroneously lead to crossing of the Weyl Chamber. To overcome this, we introduce a repulsion force deduced from the equilibrium distribution, acting close to the boundary. For details, see App.~\ref{sec:app:shooting}.

\textbf{Sampling from Invariant Distribution.} To sample from the invariant distribution of $\lambda(t)$ given by $p_{\mathrm{inv}}$ we exploit the connection between $\lambda(t)$ and $M(t)$: We first sample from the Gaussian invariant distribution of $M(t)$ and then perform an eigendecomposition.

\textbf{Comparison to direct simulation.}~There are clear efficiency and precision reasons for why we do not sample from \eqref{eq:SDE-mtx} in combination with eigendecompositions, see App.~\ref{app:comparision:why-direct-simulation-is-unwise}. Moreover, our SDE implementation renders the training runtime on the order of a simulation-free diffusion model.

\subsection{Eigenvector Time Reversal and Numerical Scheme} \label{sec:dysondiff-eigenvector}
While DyDM focuses on learning spectra, a defining feature is that the eigenvector information remains accessible. We now present the time reversal and numerical scheme for the \eqref{eq:SDE-eigenvectors}, requiring novel methods that go beyond existing work on diffusions on the Lie group \citep{debortoliRiemannianScoreBasedGenerative2022,bertoliniDiffusionGenerativeModeling2025}.

 From \eqref{eq:SDE-eigenvectors}, we derive in App.~\ref{sec:app:DiffusionModelEigenvectorSDE} that  $X_t = (v_1(t), \dots, v_n(t))$ follows  $X_{t + \d t} = X_t \exp(\d Z_t(\lambda))$ where $\d Z_{t}(\lambda) = \sqrt{\alpha} \sum_{ \ell < k} \frac{E_{\ell,k} \d W_{\ell, k}}{\lambda_k(t) - \lambda_\ell(t)}$ with $E_{\ell,k}$ being $1$ at the $(\ell,k)$-entry, $-1$ at $(k,\ell)$, and $0$ otherwise, and i.i.d.\ Wiener increments $ \d W_{\ell, k}$. Note that $\d Z_t(\lambda)^T = -\d Z_t(\lambda)$ is in the Lie algebra $\mathfrak{o}(n)$ of the Lie group $\mathrm{O}(n)$, and hence the matrix exponential implies that $X_{t+ \d t}\in \mathrm{O}(n)$.

The time-reversed diffusion $\bar X_t$ on the Lie group follows $\bar X_{t + \d t} = \bar X_t\exp(\d \bar Z_t(\lambda,\bar X_t))$ with $\d \bar Z_t(\lambda,\bar X_t)=\alpha\nabla \ln p_t(\bar\lambda, \bar X_t) dt+ \sqrt{\alpha}\sum_{\ell < k}  \frac{ E_{(\ell,k)}}{\lambda_k(t) - \lambda_{\ell}(t)}  \d\bar W^{(\ell,k)}_t\in\mathfrak{o}(n)$
  as we show in App.~\ref{Appendix:EigenvectorTimeReversal}.

Phrasing the forward- and backward-dynamics, \emph{including the score}, via Lie \emph{algebra} increments $\d Z_t,\d \bar Z_t$ offers multiple key advantages. First, orthogonality and normalization of $X(t), \bar X(t)$ during simulation follow directly. Crucially, the number of degrees of freedom is $n(n-1)/2$ instead of the naive $n^2$, as is apparent from the summation over $\ell<k$. Strikingly, it enables to implement a permutation-invariant learning architecture. While permutation equivariance of the columns of $X_t$ is already resolved, since they correspond to uniquely ordered eigenvalues, permutation equivariance of the rows of $X_t$ is covered by the following Lemma. This represents two levels of simplification, since permutation equivariance of both, rows and columns, is simplified to fixed columns and permutation-invariance in the rows, as we prove in App.~\ref{sec:app:eigenvecs:lie-algebra-invariance}.

\begin{lemma}[Permutation invariance of score in Lie algebra]\label{lem:permutation-invariance-lie-algebra}
    For any permutation matrix $P$ we have $\nabla \ln p_t(\bar\lambda, P\bar X_t)=\nabla \ln p_t(\bar\lambda, \bar X_t)\in\mathfrak{o}(n)$ for all $t\ge0$, $\bar\lambda(t)\in {\rm Weyl\ Chamber}$ and $\bar X_t\in O(n)$.
\end{lemma}
Finally, we exploit the symmetry under sign flips of eigenvectors that cannot influence the dynamics. This implies that the score obeys $\nabla \ln p_t(\bar \lambda, \bar X_t F)=F\nabla \ln p_t(\bar \lambda, \bar X_t)F$ for any diagonal matrix $F$ with entries $\pm 1$, thus a reduction of the learning target of order $2^d$, as we prove in App.~\ref{sec:app:eigenvecs:sign-symmetry}.

\section{Experiments} \label{sec:exp}

We empirically evaluate DyDM against several state-of-the-art graph diffusion models for learning graph spectra. First, we compare them on a simple bimodal distribution between two WL-equivalent graphs, illustrating the struggle of GNN-based methods (\Cref{sec:exp:dirac-delta}).
Next, we carry out a comparison on a standard graph benchmark dataset and demonstrate scalability on a larger dataset ($15{,}000$ graphs).

\begin{table}[t]%
\caption{Statistical distances of DyDM compared to established models:  DyDM learns the spectrum better in both the $n$-dimensional mean and the marginal Wasserstein sense, without requiring ad~hoc data augmentation. Results are rounded to two decimal places, and results $(*)$ are equal until the fourth decimal, and $(\dagger)$ until the third decimal (see App.~\ref{sec:app:more-digits}).}
\centering
\begin{tabular}{|l|cl|cl|cl|}
\hline
\textbf{Dataset} & \multicolumn{2}{c|}{\makecell{WL-Bimodal}} & \multicolumn{2}{c|}{Comm.\ Small}  & \multicolumn{2}{c|}{Brain} \\
\textbf{Distance}& $\mu$ & $\mathcal{W}_{\mathrm{marg}}$ &  $\mu$ &$\mathcal{W}_{\mathrm{marg}}$ 
&  $\mu$ & $\mathcal{W}_{\mathrm{marg}}$  \\
\hline
DyDM (ours)&
\best{0.02}&\almostbest{0.01^{*}}& %
\best{0.07} & \best{0.02} & %
\best{0.05}& \best{0.03^{\dagger}}\\ %
EDP-GNN & 
0.13&0.08 & %
0.42& 0.14& %
0.07 & $0.03^{\dagger}$ %
\\
GDSS & 
0.24&0.13 & %
0.38 & 0.14 & %
0.33 &  0.12 %
\\
ConGress & 
0.38&0.16& %
0.27&0.11& %
0.13 & $0.03^{\dagger}$  %
\\
DiGress (no trick)&
1.06&0.29 & %
2.51&0.45 & %
0.57 & 0.17 %
\\
DiGress (trick)&
0.03&\best{0.01^{*}}& %
0.09 & 0.03& %
0.12&\almostbest{0.03^{\dagger}}   %
\\
\hline
\end{tabular}
 \label{tab:distances}
\end{table}

We further demonstrate  in App.~\ref{sec:app:DiffusionModelEigenvectorSDE} the accessibility of the eigenvectors by learning the Lie group SDE's time reversal of Section~\ref{sec:dysondiff-eigenvector} on $SO(3)$ and with a numerical adjustment we learn a Lie group diffusion for the WL-Bimodal case.

\subsection{Methodology}\label{sec:exp:methodology} \label{sec:exp:dirac-delta} \label{sec:exp:community}
We compare to the GNN-based models EDP-GNN \citep{niuPermutationInvariantGraph2020} and GDSS \citep{joScorebasedGenerativeModeling2022}, as well as graph-transformer-based ConGress and DiGress \citep{vignacDiGressDiscreteDenoising2022}, where among all the models only DiGress uses a data-augmentation trick: it adds certain graph features, including cycle counts and the first $6$ eigenvalues (for details, see \Cref{sec:related-work}), but this \emph{trick} is supposedly inessential for building a good model \citep{vignacDiGressDiscreteDenoising2022}. The trick, however, improves the learning of certain features, but not necessarily other subgraph structures  \citep{wangGraphDiffusionModels2025}. We thus compare to both the DiGress model ``without the trick'', i.e., just a graph (Markov chain) diffusion model, and with the data-augmentation trick. As we evaluated the full spectrum while the published experimental results only showed partial or no information about the spectrum, we need explicit access to the samples. Since not all models \citep{vignacDiGressDiscreteDenoising2022, niuPermutationInvariantGraph2020} report snapshots, we had to retrain those also on the standard datasets.
We test DyDM on the following datasets, taking recent advice on limitations of certain graph benchmarks into account \cite{bechler-speicherPositionGraphLearning2025}.

\textbf{WL-Bimodal.}~We train the simple bimodal distribution in Fig.~\ref{fig:wl-demonstration}, consisting of $80\%$ graph A and $20\%$ graph B, which are WL equivalent. The dataset has $N=5,000$ random permutations of A and B, and we follow the standard test/train split procedure \citep{joScorebasedGenerativeModeling2022, youGraphRNNGeneratingRealistic2018, niuPermutationInvariantGraph2020} using $80\%$ of the data as train data and the remaining $20\%$ as test data. For hyperparameter tuning of the comparison models, see App.~\ref{sec:app:comparing-to-benchmark-models}.

\textbf{Community.}~Being a standard benchmark \citep{youGraphRNNGeneratingRealistic2018,niuPermutationInvariantGraph2020,joScorebasedGenerativeModeling2022}, we include it for comparison. However, due to heavy undersampling, we can only test for memorization. From our perspective, memorization is the best that can be tested with said benchmark, and we elaborate on this in App.~\ref{sec:app:comparing-to-benchmark-models:undersampling}.

\textbf{Brain.}~From the human connectome graph we drew $15,000$ ego-graphs (i.e.\ the induced subgraph of neighborhoods) of size $5$ to $10$ vertices~\cite{bigbrain,nr-aaai15}. Crucially, this dataset demonstrates scalability to a number of graphs large enough for high statistical fidelity (faithful representation of the underlying distribution) in test and train set. For details, see App.~\ref{sec:app:datasets}.

\subsection{Results}\label{sec:results}

\indent In \Cref{tab:distances} we report mean distances and marginal Wasserstein distances of the spectra. Explicitly, if $\nu_{\mathrm{test}}$ is the distribution of the spectrum of the test dataset and $\nu_{\mathrm{samp}}$ the distribution of our samples, then the distance between the means in $\R^n$ is ${\mu( \nu_{\mathrm{samp}}, \nu_{\mathrm{test}})\!=\!\norm{\mathbb{E}_{\lambda \sim \nu_{\mathrm{samp}}}[\lambda]\!-\!\mathbb{E}_{\lambda  \sim \nu_{\mathrm{test}}}[\lambda]}_2}.$
For statistical feasibility, instead of calculating the full Wasserstein distance we use the averaged marginal Wasserstein distance given by 
$\mathcal{W}_{\mathrm{marg}}(\nu_{\mathrm{samp}}, \nu_{\mathrm{test}}) = \frac{1}{n}\sum_{k= 1}^n \mathcal{W}((\nu_{\mathrm{samp}})_k, (\nu_{\mathrm{test}})_k),$ for $(\nu_{\mathrm{samp}})_k$ the marginal distribution in dimension $k$ and $\mathcal{W}$ the Wasserstein distance between two 1d distributions. With these %
metrics we can evaluate both marginal %
and high-dimensional effects.

These metrics also reveal the limitations of GNN- and Graph-Transformer-based models on the simple WL-Bimodal dataset, as described in \Cref{fig:wl-demonstration}. We also see that this extends to the real-world benchmark dataset Community Small. DyDM, on the other hand, consistently overcomes these issues. Even if we compare to the model with ad~hoc feature augmentation (DiGress with ``trick'', on which we elaborate in \Cref{sec:related-work}), DyDM -- which does not employ feature augmentation -- either improves on or is on par with the feature augmented DiGress.
We demonstrate that this performance still holds when working on large datasets, such as the $15,000$ ego-graphs from the Brain dataset.

With the learned eigenvectors in App.~\ref{sec:app:DiffusionModelEigenvectorSDE},
we demonstrate in Table~\ref{tab:lie-diffusion-best} that the remaining information remains indeed accessible and viable to generate accurate and valid eigenvectors. More precisely, our method performs best in all metrics except the trivial degree distribution in the WL-case -- which aligns with our discussion of the degree distribution being uniform on the WL-equivalence class.

\section{Related Work} \label{sec:related-work}

The spectrum carries key features of graphs \citep{brouwerSpectraGraphs2012}, so that spectral methods have generally proven fruitful in graph research, exploiting information encoded in the spectrum, e.g.\ for  graph comparison \citep{wilson2008study}, or in dominant eigenvectors, such as in clustering, community detection \citep{shi2000normalized,newman2013spectral}, and network embedding \citep{belkin2003laplacian}. For graph diffusion models, however, the spectrum has been used only as either auxiliary features for data augmentation \citep{vignacDiGressDiscreteDenoising2022}, or in a way which makes any remaining $\Theta(n^2)$ degrees of freedom inaccessible, as we outline below.

\textbf{Data augmentation.}~Ref.~\cite{vignacDiGressDiscreteDenoising2022} acknowledges the importance of spectra of graphs and adds several auxiliary features, among them the first $6$ eigenvalues of the Graph Laplacian as graph-level features and the first $2$ non-zero eigenvectors as vertex-level features. While the ``trick'' of adding features is not principled,  we show in \Cref{tab:distances} that these auxiliary features are \emph{necessary} for DiGress to provide a good model. These features are not needed by DyDM, which builds on an \emph{analytical} expression of the evolution of \emph{all} eigenvalues during diffusion of a symmetric matrix.

\textbf{Spectral methods.}~Spectral methods for graph learning have been employed in a variety of contexts, for instance with GANs \citep{martinkusSPECTRESpectralConditioning2022}, which have, however, been outperformed by diffusion-based models \citep{vignacDiGressDiscreteDenoising2022}.
Spectral information for graph \emph{diffusion} models has been explored in recent work, however, none of them considers how a diffusion of the spectrum impacts the remaining information (eigenvectors), leading to either a model that cannot guarantee orthogonality of the eigenvectors \citep{minelloGeneratingGraphsSpectral2025}, or a model
\emph{solely on the spectrum}  as \emph{if the eigenvectors were not impacted by diffusion} \citep{luoFastGraphGeneration2024}, hence making any recovery of remaining information in the form of the eigenvectors impossible. Ref.~\cite{luoFastGraphGeneration2024} argues that the spectrum contains much information, so that upon sampling the spectrum from the OU-based model, the eigenvectors are simply sampled by taking the eigenvectors of a training sample chosen uniformly at random. However, this strategy is very limited, as it allows only for a diffusion in $n$ parameters, losing $\Theta(n^2)$ %
degrees of freedom \emph{irrecoverably}. Since we learn in DyDM the spectrum of an OU diffusion \emph{on the entire graph}, the remaining information remains accessible, see %
\Cref{thm:eigenvector-SDE}. This allows %
learning %
the eigenvectors beyond sampling uniformly %
from %
training data \citep{luoFastGraphGeneration2024}. Our Lie group formulation enforces orthogonality intrinsically, without unnecessarily overparameterizing the degrees of freedom (using $n(n-1)/2 + n$ instead of $n(n+1)$) and without the need for rejection sampling \cite{minelloGeneratingGraphsSpectral2025}.

\section{Limitations and Extensions} \label{sec:limitations-and-extensions}

\noindent \textbf{Beyond graphs.}~Our experiments are limited to graphs. However, since the Dyson SDE is valid for the domain of $\mathrm{Sym}(\R^{n\times n})$, it is applicable beyond graphs. For instance, if correlations between $n$ points are represented as covariance matrices, DyDM could learn the spectrum and thereby quantify the effective dimensionality or concentration of variance along principal components \citep{chenPrincipalcomponentAnalysisParticle2015, estavoyerTheoreticalAnalysisPrincipal2022, hessSimilaritiesPrincipalComponents2000}.

\noindent \textbf{Complex values.}~While our equations and numerics are limited to real-valued data, an interesting generalization of Dyson-BM is to the complex domain. Thus, generalizing to complex-weighted graphs \citep{tianStructuralBalanceRandom2024,amadoComplexWeightedConvolutionalNetworks2025}:~instead of working on $\mathrm{Sym}(\mathbb{R}^{n\times n})$, one works on the space of Hermitian matrices. Similarly, matrices over the algebra of real quaternions may be considered \citep{dysonBrownianMotionModelEigenvalues1962}. 

\noindent \textbf{Emphasis on dominant eigenmodes.}~By changing the loss function, lower eigenvalue-eigenvector pairs could be penalized stronger, thereby learning the graph's main modes, determining key structural properties \cite{brouwerSpectraGraphs2012}.

\section{Conclusion}\label{sec:conclusion}
Leveraging the analytical insights offered by Dyson's Brownian motion from random matrix theory, we introduced DyDM, a diffusion model for spectral learning. 
In the graph domain, we demonstrated the limitations of existing architectures, such as GNNs and graph transformers. 
Building on these insights, we decouple the dynamics so that the spectral component is not constrained by inductive biases, thereby expanding the scope of suitable architectures. 
This allows DyDM to learn the spectrum -- even for challenging graph families -- without being restricted to permutation-equivariant networks. 
This approach eliminates architecture-induced hallucination, enabling DyDM to learn the true distributions accurately and without data augmentation, as demonstrated experimentally.
Furthermore, our framework and the eigenvector SDE lay the foundation for several future extensions with applications also beyond graphs. 
We hope this work opens a new direction for enforcing inductive biases beyond the choice of architecture.

\paragraph{Acknowledgments}
We thank Jakiw Pidstrigach and Marc Roth for helpful discussions. Financial support from the European Research Council (ERC) under the European Union’s Horizon Europe research and innovation program (Grant Agreement No. 101086182 to A.~G.), the Rhodes Trust and EPSRC Centre for Doctoral Training in Mathematics of Random Systems: Analysis, Modelling and Simulation (EPSRC Grant EP/S023925/1) (to T.~S.) and the Swiss National Science Foundation (to C.K. grant number 235409) are acknowledged. This work used the Scientific Compute Cluster at GWDG, the joint data center of Max Planck Society for the Advancement of Science (MPG) and University of Göttingen. In part funded by the Deutsche Forschungsgemeinschaft (DFG, German Research Foundation) – 405797229.
We acknowledge the help of LLMs to aid in implementing standard methods.

\bibliographystyle{unsrtnat} 
\bibliography{paper} 

\begin{thebibliography}{58}
\providecommand{\natexlab}[1]{#1}
\providecommand{\url}[1]{\texttt{#1}}
\expandafter\ifx\csname urlstyle\endcsname\relax
  \providecommand{\doi}[1]{doi: #1}\else
  \providecommand{\doi}{doi: \begingroup \urlstyle{rm}\Url}\fi

\bibitem[{Sohl-Dickstein} et~al.(2015){Sohl-Dickstein}, Weiss, Maheswaranathan,
  and Ganguli]{sohl-dicksteinDeepUnsupervisedLearning2015}
Jascha {Sohl-Dickstein}, Eric Weiss, Niru Maheswaranathan, and Surya Ganguli.
\newblock Deep {{Unsupervised Learning}} using {{Nonequilibrium
  Thermodynamics}}.
\newblock In \emph{International Conference on Machine Learning}, 2015.

\bibitem[Song et~al.(2021)Song, {Sohl-Dickstein}, Kingma, Kumar, Ermon, and
  Poole]{songScoreBasedGenerativeModeling2021}
Yang Song, Jascha {Sohl-Dickstein}, Diederik~P. Kingma, Abhishek Kumar, Stefano
  Ermon, and Ben Poole.
\newblock Score-{{Based Generative Modeling}} through {{Stochastic Differential
  Equations}}.
\newblock In \emph{International Conference on Learning Representations}, 2021.

\bibitem[Ho et~al.(2020)Ho, Jain, and
  Abbeel]{hoDenoisingDiffusionProbabilistic2020}
Jonathan Ho, Ajay Jain, and Pieter Abbeel.
\newblock Denoising diffusion probabilistic models.
\newblock In \emph{Proceedings of the 34th {{International Conference}} on
  {{Neural Information Processing Systems}}}, 2020.

\bibitem[Zhang et~al.(2023)Zhang, Zhang, Zheng, Zhang, Qamar, Bae, and
  Kweon]{zhangSurveyAudioDiffusion2023}
Chenshuang Zhang, Chaoning Zhang, Sheng Zheng, Mengchun Zhang, Maryam Qamar,
  Sung-Ho Bae, and In~So Kweon.
\newblock A {{Survey}} on {{Audio Diffusion Models}}: {{Text To Speech
  Synthesis}} and {{Enhancement}} in {{Generative AI}}.
\newblock \emph{arXiv preprint arXiv:2303.13336}, 2023.

\bibitem[Croitoru et~al.(2023)Croitoru, Hondru, Ionescu, and
  Shah]{croitoruDiffusionModelsVision2023}
Florinel-Alin Croitoru, Vlad Hondru, Radu~Tudor Ionescu, and Mubarak Shah.
\newblock Diffusion {{Models}} in {{Vision}}: {{A Survey}}.
\newblock \emph{IEEE Transactions on Pattern Analysis and Machine
  Intelligence}, 45\penalty0 (9):\penalty0 10850--10869, September 2023.

\bibitem[Watson et~al.(2023)Watson, Juergens, Bennett, Trippe, Yim, Eisenach,
  Ahern, Borst, Ragotte, Milles, Wicky, Hanikel, Pellock, Courbet, Sheffler,
  Wang, Venkatesh, Sappington, Torres, Lauko, De~Bortoli, Mathieu, Ovchinnikov,
  Barzilay, Jaakkola, DiMaio, Baek, and Baker]{watsonNovoDesignProtein2023}
Joseph~L. Watson, David Juergens, Nathaniel~R. Bennett, Brian~L. Trippe, Jason
  Yim, Helen~E. Eisenach, Woody Ahern, Andrew~J. Borst, Robert~J. Ragotte,
  Lukas~F. Milles, Basile I.~M. Wicky, Nikita Hanikel, Samuel~J. Pellock,
  Alexis Courbet, William Sheffler, Jue Wang, Preetham Venkatesh, Isaac
  Sappington, Susana~V{\'a}zquez Torres, Anna Lauko, Valentin De~Bortoli, Emile
  Mathieu, Sergey Ovchinnikov, Regina Barzilay, Tommi~S. Jaakkola, Frank
  DiMaio, Minkyung Baek, and David Baker.
\newblock De novo design of protein structure and function with
  {{RFdiffusion}}.
\newblock \emph{Nature}, 620\penalty0 (7976):\penalty0 1089--1100, August 2023.

\bibitem[Sun and Yang(2023)]{sunDIFUSCOGraphbasedDiffusion2023}
Zhiqing Sun and Yiming Yang.
\newblock {{DIFUSCO}}: {{Graph-based Diffusion Solvers}} for {{Combinatorial
  Optimization}}.
\newblock In \emph{Advances in Neural Information Processing Systems}, 2023.

\bibitem[Niu et~al.(2020)Niu, Song, Song, Zhao, Grover, and
  Ermon]{niuPermutationInvariantGraph2020}
Chenhao Niu, Yang Song, Jiaming Song, Shengjia Zhao, Aditya Grover, and Stefano
  Ermon.
\newblock Permutation {{Invariant Graph Generation}} via {{Score-Based
  Generative Modeling}}.
\newblock In \emph{International Conference on Artificial Intelligence and
  Statistics}, 2020.

\bibitem[Jo et~al.(2022)Jo, Lee, and Hwang]{joScorebasedGenerativeModeling2022}
Jaehyeong Jo, Seul Lee, and Sung~Ju Hwang.
\newblock Score-based {{Generative Modeling}} of {{Graphs}} via the {{System}}
  of {{Stochastic Differential Equations}}.
\newblock In \emph{International Conference on Machine Learning}, 2022.

\bibitem[Vignac et~al.(2022)Vignac, Krawczuk, Siraudin, Wang, Cevher, and
  Frossard]{vignacDiGressDiscreteDenoising2022}
Clement Vignac, Igor Krawczuk, Antoine Siraudin, Bohan Wang, Volkan Cevher, and
  Pascal Frossard.
\newblock {{DiGress}}: {{Discrete Denoising}} diffusion for graph generation.
\newblock In \emph{International Conference on Learning Representations}, 2022.

\bibitem[Babai(2016)]{babaiGraphIsomorphismQuasipolynomial2016}
L{\'a}szl{\'o} Babai.
\newblock Graph isomorphism in quasipolynomial time [extended abstract].
\newblock In \emph{Proceedings of the Forty-Eighth Annual {{ACM}} Symposium on
  {{Theory}} of {{Computing}}}, 2016.

\bibitem[Huang et~al.(2022)Huang, Sun, Du, Fu, and
  Lv]{huangGraphGDPGenerativeDiffusion2022}
Han Huang, Leilei Sun, Bowen Du, Yanjie Fu, and Weifeng Lv.
\newblock {{GraphGDP}}: {{Generative Diffusion Processes}} for {{Permutation
  Invariant Graph Generation}}.
\newblock In \emph{{{IEEE International Conference}} on {{Data Mining}}}, 2022.

\bibitem[Xu et~al.(2024)Xu, Qiu, Chen, Chen, Fan, Pan, Zeng, Das, and
  Tong]{xuDiscretestateContinuoustimeDiffusion2024}
Zhe Xu, Ruizhong Qiu, Yuzhong Chen, Huiyuan Chen, Xiran Fan, Menghai Pan,
  Zhichen Zeng, Mahashweta Das, and Hanghang Tong.
\newblock Discrete-state {{Continuous-time Diffusion}} for {{Graph
  Generation}}.
\newblock In \emph{Advances in Neural Information Processing Systems}, 2024.

\bibitem[Morris et~al.(2019)Morris, Ritzert, Fey, Hamilton, Lenssen, Rattan,
  and Grohe]{morrisWeisfeilerLemanGo2019}
Christopher Morris, Martin Ritzert, Matthias Fey, William~L. Hamilton, Jan~Eric
  Lenssen, Gaurav Rattan, and Martin Grohe.
\newblock Weisfeiler and {{Leman Go Neural}}: {{Higher-Order Graph Neural
  Networks}}.
\newblock In \emph{AAAI Conference on Artificial Intelligence}, 2019.

\bibitem[Xu et~al.(2018)Xu, Hu, Leskovec, and Jegelka]{xuHowPowerfulAre2018}
Keyulu Xu, Weihua Hu, Jure Leskovec, and Stefanie Jegelka.
\newblock How {{Powerful}} are {{Graph Neural Networks}}?
\newblock In \emph{International {{Conference}} on {{Learning
  Representations}}}, 2018.

\bibitem[Huang and Yau(2024)]{huangSpectrumRandomDregular2024}
Jiaoyang Huang and Horng-Tzer Yau.
\newblock Spectrum of random d-regular graphs up to the edge.
\newblock \emph{Communications on Pure and Applied Mathematics}, 77\penalty0
  (3):\penalty0 1635--1723, 2024.
\newblock ISSN 1097-0312.

\bibitem[Luo et~al.(2024)Luo, Mo, and Pan]{luoFastGraphGeneration2024}
Tianze Luo, Zhanfeng Mo, and Sinno~Jialin Pan.
\newblock Fast {{Graph Generation}} via {{Spectral Diffusion}}.
\newblock \emph{IEEE Transactions on Pattern Analysis and Machine
  Intelligence}, 46:\penalty0 3496--3508, 2024.

\bibitem[James et~al.(2023)James, Witten, Hastie, Tibshirani, and
  Taylor]{jamesIntroductionStatisticalLearning2023}
Gareth James, Daniela Witten, Trevor Hastie, Robert Tibshirani, and Jonathan
  Taylor.
\newblock \emph{An {{Introduction}} to {{Statistical Learning}}: With
  {{Applications}} in {{Python}}}.
\newblock Springer {{Texts}} in {{Statistics}}. Springer, 2023.

\bibitem[Pavliotis(2014)]{pavliotisStochasticProcessesApplications2014}
Grigorios~A. Pavliotis.
\newblock \emph{Stochastic {{Processes}} and {{Applications}}: {{Diffusion
  Processes}}, the {{Fokker-Planck}} and {{Langevin Equations}}}.
\newblock Texts in {{Applied Mathematics}}. Springer New York, 2014.

\bibitem[Anderson(1982)]{andersonReversetimeDiffusionEquation1982}
Brian D.~O. Anderson.
\newblock Reverse-time diffusion equation models.
\newblock \emph{Stochastic Processes and their Applications}, 12\penalty0
  (3):\penalty0 313--326, 1982.

\bibitem[Meringer(1999)]{meringerFastGenerationRegular1999}
Markus Meringer.
\newblock Fast generation of regular graphs and construction of cages.
\newblock \emph{Journal of Graph Theory}, 30\penalty0 (2):\penalty0 137--146,
  1999.

\bibitem[Tirion(1996)]{tirionLargeAmplitudeElastic1996}
Monique~M. Tirion.
\newblock Large {{Amplitude Elastic Motions}} in {{Proteins}} from a
  {{Single-Parameter}}, {{Atomic Analysis}}.
\newblock \emph{Physical Review Letters}, 77\penalty0 (9):\penalty0 1905--1908,
  1996.

\bibitem[Haliloglu et~al.(1997)Haliloglu, Bahar, and
  Erman]{halilogluGaussianDynamicsFolded1997}
Turkan Haliloglu, Ivet Bahar, and Burak Erman.
\newblock Gaussian {{Dynamics}} of {{Folded Proteins}}.
\newblock \emph{Physical Review Letters}, 79\penalty0 (16):\penalty0
  3090--3093, October 1997.

\bibitem[Dyson(1962)]{dysonBrownianMotionModelEigenvalues1962}
Freeman~J. Dyson.
\newblock A {{Brownian-Motion Model}} for the {{Eigenvalues}} of a {{Random
  Matrix}}.
\newblock \emph{Journal of Mathematical Physics}, 3\penalty0 (6):\penalty0
  1191--1198, 1962.

\bibitem[Allez et~al.(2014)Allez, Bun, and
  Bouchaud]{allezEigenvectorsGaussianMatrices2015}
Romain Allez, Jo{\"e}l Bun, and Jean-Philippe Bouchaud.
\newblock The eigenvectors of {{Gaussian}} matrices with an external source.
\newblock \emph{arXiv preprint arXiv:1412.7108}, 2014.

\bibitem[De~Bortoli et~al.(2022)De~Bortoli, Mathieu, Hutchinson, Thornton, Teh,
  and Doucet]{debortoliRiemannianScoreBasedGenerative2022}
Valentin De~Bortoli, Emile Mathieu, Michael Hutchinson, James Thornton,
  Yee~Whye Teh, and Arnaud Doucet.
\newblock Riemannian {{Score-Based Generative Modelling}}.
\newblock \emph{Advances in Neural Information Processing Systems},
  35:\penalty0 2406--2422, December 2022.

\bibitem[Yang(2022)]{yangTopicsRandomMatrix}
Fang Yang.
\newblock Topics in random matrix theory.
\newblock https://yangf75.github.io/RMT(2022Fall,49-61).pdf, 2022.

\bibitem[Bertolini et~al.(2025)Bertolini, Le, and
  Clevert]{bertoliniDiffusionGenerativeModeling2025}
Marco Bertolini, Tuan Le, and Djork-Arn{\'e} Clevert.
\newblock Diffusion {{Generative Modeling}} on {{Lie Group Representations}}.
\newblock In \emph{Advances in Neural Information Processing Systems}, October
  2025.

\bibitem[Wang et~al.(2025)Wang, Liu, Pang, Chen, and
  Zhang]{wangGraphDiffusionModels2025}
Xiyuan Wang, Yewei Liu, Lexi Pang, Siwei Chen, and Muhan Zhang.
\newblock Do {{Graph Diffusion Models Accurately Capture}} and {{Generate
  Substructure Distributions}}?
\newblock \emph{arXiv preprint arXiv:2502.02488}, 2025.

\bibitem[{Bechler-Speicher} et~al.(2025){Bechler-Speicher}, Finkelshtein,
  Frasca, M{\"u}ller, T{\"o}nshoff, Siraudin, Zaverkin, Bronstein, Niepert,
  Perozzi, Galkin, and Morris]{bechler-speicherPositionGraphLearning2025}
Maya {Bechler-Speicher}, Ben Finkelshtein, Fabrizio Frasca, Luis M{\"u}ller,
  Jan T{\"o}nshoff, Antoine Siraudin, Viktor Zaverkin, Michael~M. Bronstein,
  Mathias Niepert, Bryan Perozzi, Mikhail Galkin, and Christopher Morris.
\newblock Position: {{Graph Learning Will Lose Relevance Due To Poor
  Benchmarks}}.
\newblock In \emph{International Conference on {{Machine Learning Position
  Paper Track}}}, 2025.

\bibitem[You et~al.(2018)You, Ying, Ren, Hamilton, and
  Leskovec]{youGraphRNNGeneratingRealistic2018}
Jiaxuan You, Rex Ying, Xiang Ren, William Hamilton, and Jure Leskovec.
\newblock {{GraphRNN}}: {{Generating Realistic Graphs}} with {{Deep
  Auto-regressive Models}}.
\newblock In \emph{International Conference on Machine Learning}, 2018.

\bibitem[Amunts et~al.(2013)Amunts, Lepage, Borgeat, Mohlberg, Dickscheid,
  Rousseau, Bludau, Bazin, Lewis, Oros-Peusquens, Shah, Lippert, Zilles, and
  Evans]{bigbrain}
Katrin Amunts, Claude Lepage, Louis Borgeat, Hartmut Mohlberg, Timo Dickscheid,
  Marc-{\'E}tienne Rousseau, Sebastian Bludau, Pierre-Louis Bazin, Lindsay~B.
  Lewis, Ana-Maria Oros-Peusquens, Nadim~J. Shah, Thomas Lippert, Karl Zilles,
  and Alan~C. Evans.
\newblock Bigbrain: An ultrahigh-resolution 3d human brain model.
\newblock \emph{Science}, 340\penalty0 (6139):\penalty0 1472--1475, 2013.

\bibitem[Rossi and Ahmed(2015)]{nr-aaai15}
Ryan~A. Rossi and Nesreen~K. Ahmed.
\newblock The network data repository with interactive graph analytics and
  visualization.
\newblock In \emph{AAAI Conference on Artificial Intelligence}, 2015.
\newblock URL \url{http://networkrepository.com}.

\bibitem[Brouwer and Haemers(2012)]{brouwerSpectraGraphs2012}
Andries~E. Brouwer and Willem~H. Haemers.
\newblock \emph{Spectra of {{Graphs}}}.
\newblock Universitext. Springer New York, New York, NY, 2012.

\bibitem[Wilson and Zhu(2008)]{wilson2008study}
Richard~C Wilson and Ping Zhu.
\newblock A study of graph spectra for comparing graphs and trees.
\newblock \emph{Pattern Recognition}, 41\penalty0 (9):\penalty0 2833--2841,
  2008.

\bibitem[Shi and Malik(2000)]{shi2000normalized}
Jianbo Shi and Jitendra Malik.
\newblock Normalized cuts and image segmentation.
\newblock \emph{IEEE Transactions on Pattern Analysis and Machine
  Intelligence}, 22\penalty0 (8):\penalty0 888--905, 2000.

\bibitem[Newman(2013)]{newman2013spectral}
Mark~EJ Newman.
\newblock Spectral methods for community detection and graph partitioning.
\newblock \emph{Physical Review E—Statistical, Nonlinear, and Soft Matter
  Physics}, 88\penalty0 (4):\penalty0 042822, 2013.

\bibitem[Belkin and Niyogi(2003)]{belkin2003laplacian}
Mikhail Belkin and Partha Niyogi.
\newblock Laplacian eigenmaps for dimensionality reduction and data
  representation.
\newblock \emph{Neural Computation}, 15\penalty0 (6):\penalty0 1373--1396,
  2003.

\bibitem[Martinkus et~al.(2022)Martinkus, Loukas, Perraudin, and
  Wattenhofer]{martinkusSPECTRESpectralConditioning2022}
Karolis Martinkus, Andreas Loukas, Nathana{\"e}l Perraudin, and Roger
  Wattenhofer.
\newblock {{SPECTRE}}: {{Spectral Conditioning Helps}} to {{Overcome}} the
  {{Expressivity Limits}} of {{One-shot Graph Generators}}.
\newblock In \emph{International {{Conference}} on {{Machine Learning}}}, 2022.

\bibitem[Minello et~al.(2025)Minello, Bicciato, Rossi, Torsello, and
  Cosmo]{minelloGeneratingGraphsSpectral2025}
Giorgia Minello, Alessandro Bicciato, Luca Rossi, Andrea Torsello, and Luca
  Cosmo.
\newblock Generating graphs via spectral diffusion.
\newblock In \emph{International Conference on Learning Representations}, 2025.

\bibitem[Chen et~al.(2015)Chen, Li{\'e}geois, {de Bruyn}, and
  Soddu]{chenPrincipalcomponentAnalysisParticle2015}
H.~Y. Chen, Rapha{\"e}l Li{\'e}geois, John~R. {de Bruyn}, and Andrea Soddu.
\newblock Principal-component analysis of particle motion.
\newblock \emph{Physical Review E}, 91\penalty0 (4):\penalty0 042308, 2015.

\bibitem[Estavoyer and Fran{\c
  c}ois(2022)]{estavoyerTheoreticalAnalysisPrincipal2022}
Maxime Estavoyer and Olivier Fran{\c c}ois.
\newblock Theoretical analysis of principal components in an umbrella model of
  intraspecific evolution.
\newblock \emph{Theoretical Population Biology}, 148:\penalty0 11--21, December
  2022.

\bibitem[Hess(2000)]{hessSimilaritiesPrincipalComponents2000}
Berk Hess.
\newblock Similarities between principal components of protein dynamics and
  random diffusion.
\newblock \emph{Physical Review E}, 62\penalty0 (6):\penalty0 8438--8448,
  December 2000.

\bibitem[Tian and Lambiotte(2024)]{tianStructuralBalanceRandom2024}
Yu~Tian and Renaud Lambiotte.
\newblock Structural {{Balance}} and {{Random Walks}} on {{Complex Networks}}
  with {{Complex Weights}}.
\newblock \emph{SIAM Journal on Mathematics of Data Science}, 6\penalty0
  (2):\penalty0 372--399, June 2024.

\bibitem[Amado et~al.(2025)Amado, Schwarz, Tian, and
  Lambiotte]{amadoComplexWeightedConvolutionalNetworks2025}
Cristina~L{\'o}pez Amado, Tassilo Schwarz, Yu~Tian, and Renaud Lambiotte.
\newblock Complex-{{Weighted Convolutional Networks}}: {{Provable
  Expressiveness}} via {{Complex Diffusion}}.
\newblock In \emph{The {{Fourth Learning}} on {{Graphs Conference}}}, October
  2025.

\bibitem[Keating(2023)]{keatingRandomMatrixTheory2023}
John~P. Keating.
\newblock Random {M}atrix {T}heory.
\newblock Lecture Notes, Oxford University, 2023.

\bibitem[Anderson et~al.(2009)Anderson, Guionnet, and
  Zeitouni]{andersonIntroductionRandomMatrices2009}
Greg~W. Anderson, Alice Guionnet, and Ofer Zeitouni.
\newblock \emph{An {{Introduction}} to {{Random Matrices}}}.
\newblock Cambridge University Press, 2009.

\bibitem[Katori and Tanemura(2003)]{katoriNoncollidingBrownianMotions2003}
Makoto Katori and Hideki Tanemura.
\newblock Noncolliding {{Brownian}} motions and {{Harish-Chandra}} formula.
\newblock \emph{Electronic Communications in Probability}, 8:\penalty0
  112--121, 2003.

\bibitem[{\O}ksendal(2003)]{oksendalStochasticDifferentialEquations2003}
Bernt {\O}ksendal.
\newblock \emph{Stochastic {{Differential Equations}}}.
\newblock Universitext. Springer Berlin Heidelberg, 2003.

\bibitem[Karras et~al.(2024)Karras, Aittala, Lehtinen, Hellsten, Aila, and
  Laine]{karrasAnalyzingImprovingTraining2024}
Tero Karras, Miika Aittala, Jaakko Lehtinen, Janne Hellsten, Timo Aila, and
  Samuli Laine.
\newblock Analyzing and {{Improving}} the {{Training Dynamics}} of {{Diffusion
  Models}}.
\newblock In \emph{Proceedings of the IEEE/CVF Conference on Computer Vision
  and Pattern Recognition}, 2024.

\bibitem[Song and Ermon(2020)]{songImprovedTechniquesTraining2020}
Yang Song and Stefano Ermon.
\newblock Improved techniques for training score-based generative models.
\newblock In \emph{Advances in Neural Information Processing Systems}, 2020.

\bibitem[Loshchilov and Hutter(2019)]{LoshchilovH19}
Ilya Loshchilov and Frank Hutter.
\newblock Decoupled weight decay regularization.
\newblock In \emph{International Conference on Learning Representations
  (ICLR)}, 2019.

\bibitem[Bradbury et~al.(2018)Bradbury, Frostig, Hawkins, Johnson, Katariya,
  Leary, Maclaurin, Necula, Paszke, Vander{P}las, Wanderman-{M}ilne, and
  Zhang]{jax2018github}
James Bradbury, Roy Frostig, Peter Hawkins, Matthew~James Johnson, Yash
  Katariya, Chris Leary, Dougal Maclaurin, George Necula, Adam Paszke, Jake
  Vander{P}las, Skye Wanderman-{M}ilne, and Qiao Zhang.
\newblock {JAX}: composable transformations of {P}ython+{N}um{P}y programs,
  2018.
\newblock URL \url{http://github.com/jax-ml/jax}.

\bibitem[Kidger and Garcia(2021)]{kidger2021equinox}
Patrick Kidger and Cristian Garcia.
\newblock {E}quinox: neural networks in {JAX} via callable {P}y{T}rees and
  filtered transformations.
\newblock \emph{Differentiable Programming workshop at Neural Information
  Processing Systems}, 2021.

\bibitem[Marjanovic et~al.(2016)Marjanovic, Piggott, and
  Solo]{marjanovicNumericalMethodsStochastic2016}
Goran Marjanovic, Marc~J. Piggott, and Victor Solo.
\newblock Numerical methods for stochastic differential equations in the
  {Stiefel} manifold made simple.
\newblock In \emph{{IEEE} 55th {Conference} on {Decision} and {Control}
  ({CDC})}, 2016.

\bibitem[Hsu(2002)]{hsu2002manifolds}
Elton~P. Hsu.
\newblock \emph{Stochastic Analysis on Nanifolds}, volume~38 of \emph{Graduate
  Studies in Mathematics}.
\newblock American Mathematical Society, Providence, RI, 2002.

\bibitem[Lee et~al.(2019)Lee, Lee, Kim, Kosiorek, Choi, and Teh]{lee2019set}
Juho Lee, Yoonho Lee, Jungtaek Kim, Adam~R. Kosiorek, Seungjin Choi, and
  Yee~Whye Teh.
\newblock Set transformer: {A} framework for attention-based
  permutation-invariant neural networks.
\newblock In \emph{International Conference on Machine Learning}, 2019.

\bibitem[Perez et~al.(2018)Perez, Strub, de~Vries, Dumoulin, and
  Courville]{perezFilm18}
Ethan Perez, Florian Strub, Harm de~Vries, Vincent Dumoulin, and Aaron~C.
  Courville.
\newblock {FiLM}: Visual reasoning with a general conditioning layer.
\newblock In \emph{Proceedings of the Thirty-Second AAAI Conference on
  Artificial Intelligence and Thirtieth Innovative Applications of Artificial
  Intelligence Conference and Eighth AAAI Symposium on Educational Advances in
  Artificial Intelligence}, 2018.

\end{thebibliography}

\medskip

\newpage 
\appendix

\section*{Appendix
}

The Appendix is structured as follows. We first derive the tractable loss $\tilde L(\theta)$ in \Cref{sec:app:loss}. We then provide proofs of the \eqref{eq:SDE-spectrum} SDE in \Cref{sec:app:eigenvalue-SDE-proof} as well as for the \eqref{eq:SDE-eigenvectors} in \Cref{sec:app:eigenvector-SDE-proof}. We then explain the applicability of Anderson's time reversal (\Cref{sec:app:time-reversal}), and provide a time-rescaling of \eqref{eq:SDE-spectrum} in \Cref{sec:app:time-rescaling}. In \Cref{sec:app:stepsize-controller}, we provide the adaptive step size controller, followed by an explanation of the shooting mechanism (\Cref{sec:app:shooting}). We provide a theoretical analysis of the complexity of the numerical update steps for \eqref{eq:SDE-spectrum} and \eqref{eq:SDE-eigenvectors} in dependence of the graph size $n$ in \Cref{sec:app:complexity-update-step}, followed by a discussion of empirical resource considerations in \Cref{sec:app:empirical-considerations}. We then explain the 
sampling procedure (\Cref{sec:app:sampling}) and give  engineering details (\Cref{sec:app:engineering}). We explain the challenges of Dyson's Brownian motion which we overcame with DyDM in \Cref{sec:app:challenges-which-we-all-solved}. We present our theoretical argument for using inductive bias in \Cref{sec:argument-against-learning-all-permutations} and prove the Lemma on WL equivalence of $k$-regular graphs in \Cref{sec:app:wl-k-reg-proof}. In \Cref{sec:app:laplacian}, we demonstrate DyDM on the Graph Laplacian, showing its applicability beyond adjacency spectra and show how it can learn properties such as the algebraic connectivity (Fiedler value). For the experiments, we first elaborate on the datasets (\Cref{sec:app:datasets}) followed by an explanation of our extensive benchmarking in \Cref{sec:app:comparing-to-benchmark-models}, including a comment about undersampling in some benchmark datasets in \Cref{sec:app:comparing-to-benchmark-models:undersampling}.
We show the learning dynamics of $4$ different runs of EDP-GNN in \Cref{sec:app:learning-dynamics}. We give mathematical details for the extension to \eqref{eq:SDE-eigenvectors} and its time-reversal in \Cref{sec:app:DiffusionModelEigenvectorSDE}, and demonstrate the recoverability of the remaining information in Table~\ref{tab:lie-diffusion-best} and  by plotting the generated graphs in Fig.~\ref{fig:graphs-we-generated}.

\section{Making the loss tractable} \label{sec:app:loss}

In this section, we deduce a general loss formula for an SDE on $\R$ that will be applied to \eqref{eq:SDE-spectrum}. We consider the process $X = (X(t))_{t \geq 0}$ determined by the SDE $$\d X(t) = a(t,X(t)) \d t + b(t,X(t)) \d W_t$$ with initial condition $X(0)$. We assume throughout this section that $X(t)$ is absolutely continuous for all $t \geq 0$ and denote by $p_t$ the density of $X(t)$. We furthermore assume that the joint densities of $X(t)$ and $X(s)$ also have density for all $t,s \geq 0$ that we write as $p_{t,s}$. Observe that by Bayes formula for $t \geq s \geq 0$ we have for $x,y\in \R^d$ that 
\begin{equation}\label{eq:pt_Bayes}
    p_{t,s}(y,x) = p_{t|s}(y|x) p_s(x),
\end{equation}
where $p_{t|s}(y|x)$ is the conditional density of $y$ given $x$.

The canonical loss arising from $s(y,t) = \grad_{M(t)} \ln p_t(M)$ is
\begin{align}
	L'(\theta) = \E{t \sim \mathcal{U}[0,T],M(0) \sim p_0, M(t) \sim p_{t \mid 0}(\cdot|M(0))}{\norm{ s_\theta(M(t),t) - \grad_{M(t)} \ln p_{t}(M(t))}^2_2} \label{eq:OU-loss-unconditional}.
\end{align}  Note that the difference between $L'(\theta)$ and $L(\theta)$ is that we use the gradient of the \emph{conditional} density $\grad_{M(t)}\ln p_{t \mid 0}(M(t)|M(0))$ in $L(\theta)$. It is a well-known fact that $L'(\theta) = L(\theta) + \mathrm{const}(\theta)$.  We will first explain how the loss $L'(\theta)$ from \eqref{eq:OU-loss-unconditional} and $\tilde{L}(\theta)$ from \eqref{eqn:loss} are the same up to a constant, that is $L'(\theta) = \tilde{L}(\theta) + \mathrm{const}(\theta)$, which therefore results in the same gradient descent as with $L(\theta)$. 

We generalize the loss from \eqref{eq:OU-loss-unconditional} to weighting time by a function $\eta$. So we consider the following generalized loss 
\begin{align}
L'(\theta) 
\coloneqq& \frac{1}{T}\int_0^T \eta(t) \int_{\R^n}  \int_{\R^n}  \norm {s_\theta (y, t) - \grad_y\ln(p_{t}(y )) }_2^2 \; p_{t ,0}(y, x) \,\d y \d x \d t \nonumber\\
	=& \frac{1}{T}\int_0^T \eta(t) \int_{\R^n}  \int_{\R^n}  \norm {s_\theta (y, t) - \grad_y\ln(p_{t}(y )) }_2^2 \; p_{t \mid 0}(y \mid x)\d y p_0(x)\d x \d t,
\end{align}
where $\eta(t)$ is some weighting function with $\int_{0}^T \eta(t) \d t= 1$. We wrote it in the above form with $p_{t \mid 0}$ denoting the conditional density, since the three integrals can be replaced by $\E{t \sim U{[0,T]}}{\cdots \E{x \sim p_0}{\E{y \sim p_{t \mid 0}(\cdot \mid x)}{\norm{\cdots}_2^2}}}$, which we can sample from if we assume (1) sample access to $p_0$, (2) known density at any $t$ given a dirac-delta $p_0$, (3) the term in the norm is tractable. (1) is assumed by the problem definition, (2) is known for an Ornstein-Uhlenbeck forward SDE, (3) can be solved by realizing that there is an equivalent loss function $\hat L(\theta)= L(\theta) + \rm const(\theta)$ where $p_t$ inside the norm is converted to a $p_{t \mid 0}$, which is known (Gaussian density) in an OU setting.

Here, steps (2) and (3) fail. Note that by the polarization identity $$\norm {s_\theta (y, t) - \grad_y\ln(p_{t}(y )) }_2^2 =\norm {s_\theta (y, t) }_2^2 + \norm { \grad_y\ln(p_{t}(y )) }_2^2 -2 s_\theta (y, t)^T\grad_y\ln(p_{t}(y ))$$ We will rewrite the mixed term $ s_\theta(y,t)^T \grad_{y}\ln(p_t(y))$ of the $L^2$ norm in such a way that the first quadratic term $\norm {s_\theta (y, t) }_2^2$ remains unchanged and the second quadratic term $\norm { \grad_y\ln(p_{t}(y )) }_2^2$ is constant in $\theta$. First, we rewrite
\begin{align}
	& \frac{1}{T} \int_0^T \eta(t)\int_{\R^n\times \R^n}  s_\theta(y,t)^T \grad_{y}\ln(p_t(y)) p_{t,0}(y,x) \diff x \diff y \diff t \label{mixedterm}\\
	=& \frac{1}{T} \int_0^T \eta(t)\int_{\R^n} s_\theta(y,t)^T \grad_{y}\ln(p_t(y)) p_{t}(y)\diff y \diff t \nonumber\\
	=& \frac{1}{T} \int_0^T \eta(t)\int_{\R^n} s_\theta(y,t)^T \grad_{y}\left[p_t(y)\right]\diff y \diff t. \label{MixedTermSimplerForm}
\end{align} 
We next observe that for $0 \leq s' \leq s$ we have the following:
\begin{align}
    \grad_{y}\left[p_s(y)\right] &= \grad_{y}\left[\int_{\R^n} p_{s | s'}(y | z) p_{s'}(z) \diff z\right]\nonumber \\
    &= \int_{\R^n} \grad_{y} p_{s | s'}(y | z) p_{s'}(z) \diff z \nonumber\\
    &= \int_{\R^n} \grad_{y} p_{s | s'}(y | z) \frac{p_{s,s'}(y,z)}{p_{s|s'}(y|z)} \diff z \nonumber\\
    &= \int_{\R^n} \grad_{y} \ln  p_{s | s'}(y | z)p_{s,s'}(y,z)\diff z.\nonumber
\end{align}
So we now perform in \eqref{MixedTermSimplerForm} for a small $h > 0$ a change of variables to $t + h$ and apply the latter equality with $s = t + h$ and $s'= t$. Then up to ignoring the boundary at $0$ and $T$, and using by \eqref{eq:pt_Bayes} that $p_{t+h,t}(y,z) = p_{t + h|t}(y|z)p_t(z)$ it follows that the loss from \eqref{eqn:loss} \begin{align}
    \eqref{MixedTermSimplerForm} = \frac{1}{T} \int_0^T \eta(t + h)\int_{\R^n} \int_{\R^n} s_\theta(y,t + h)^T \grad_{y} \ln p_{t + h|t}(y|z) p_{t + h|t}(y|z) \diff y \, p_t(z) \diff z \diff t. 
\end{align} So it follows that 
\begin{align}
    \tilde{L}(\theta) = \frac{1}{T} \int_0^T \eta(t + h)\int_{\R^n} \int_{\R^n} \norm {s_\theta (y, t+h) - \grad_y\ln(p_{t+h|t}(y|z)) }_2^2 p_{t+h|t}(y|z) \diff y \, p_t(z) \diff z \diff t\end{align} is equal to $L(\theta)$ up to a constant term in $\theta$ (ignoring the boundary terms at $0$ and $T$). 

We will now make a series of approximations to calculate the loss $\tilde{L}(\theta)$. The first one is to approximate the integral $\int_0^T$ over $t$ by a sum $\sum_{i = 1}^k$ over the time points $t_0 < t_1 < \ldots < t_k$ with $t_0 = 0$ and $t_k = T$.
The second approximation we make is that we approximate the latter integral $\int_{\R^n}\int_{\R^n}$ by sampling a path from $p_t$ at the time steps $t_i$. Indeed, we denote for each integer $1 \leq r \leq N$ by $x^{(r)}_{t_i}$ the sample path of $p_t$. Moreover, we actually make the time grid also dependent on our sample path. So for each $1 \leq r \leq N$, let $t_0^{(r)} < t_1^{(r)} < \ldots < t_{k^{(r)}}^{(r)}$ with $t_0^{(r)} = 0$ and $t_{k^{(r)}}^{(r)} = T$ be the discretization of $[0,T]$.  Thus, the overall loss can be approximated as
\begin{align}\label{IntialApproximation}
	\tilde{L}(\theta) \approx &\frac{1}{N}\sum_{r = 1}^N \sum_{i = 1}^{k^{(r)}} \frac{t_i^{(r)} - t_{i-1}^{(r)}}{T} \eta(t_i^{(r)})  \norm{  s_\theta(x_{i}^{(r)},t_i^{(r)}) - \grad_{x_i^{(r)}}\left[\ln(p_{t_i^{(r)} \mid t_{i-1}^{(r)}}(x_{i}^{(r)} \mid x_{i-1}^{(r)})) \right]}_2^2\nonumber  \\&+ \rm const(\theta).
\end{align}

We finally approximate the incremental score function $\grad_{x_i^{(r)}}\left[\ln(p_{t_i^{(r)} \mid t_{i-1}^{(r)}}(x_{i}^{(r)} \mid x_{i-1}^{(r)})) \right]$ as follows. If $h>0$ is a small time step, we can approximate the conditional random variable $X_{t+h} \mid  \mathcal  F_t$ with $x:= X_t$ by
\begin{align}
	X_{t+h} \mid  \mathcal  F_t 
	&\approx x + a(t,x)h + b(t,x) N(0,h) \tag{where $N(0,h)$ is a centered Gaussian RV with variance $h$}\\
	&\sim \mathcal \Normal{x+a(t,x)h}{b^2(t,x)h},
\end{align}
so that we have for the density
\begin{align}
	p_{t+h|t}(y|x) 
	&\approx \frac{1}{\sqrt{2 \pi b^2(t,x) h}} \exp(-\frac{(y-x-a(t,x)h)^2}{2 b^2(t,x)h}),
\end{align}
which means for the scores
\begin{align}\label{ConditionalDistributionApprox}
	\grad_y \ln \left(p_{t+h|t}(y|x)  \right)
	&\approx - \frac{y-x-a(t,x)h}{b^2(t,x) h}.
\end{align}

Thus, combining \eqref{IntialApproximation} and \eqref{ConditionalDistributionApprox}, the loss $\tilde{L}(\theta)$ can be approximated by the following as we use in our model:
\begin{align}
	\frac{1}{N}\sum_{r = 1}^N \sum_{i = 1}^{k^{(r)}} \frac{t_i^{(r)} - t_{i-1}^{(r)}}{T} \eta(t_i^{(r)}) \norm{  s_\theta(x_i^{(r)},t_i^{(r)}) 
	- \frac{
			 a(t_{i-1}^{(r)},x_{i-1}^{(r)})\cdot (t_{i}^{(r)}-t_{i-1}^{(r)}) - (x_i^{(r)} - x_{i-1}^{(r)})
		}{
			b^2(t_{i-1}^{(r)},x_{i-1}^{(r)})\cdot (t_{i}^{(r)}-t_{i-1}^{(r)}) 
		}
	}_2^2
    \label{eq:pathwise-loss} %
\end{align}

\section{Spectral Dyson SDE}\label{sec:app:eigenvalue-SDE-proof}

\begin{reptheorem}{thm:eigenvalue-SDE}
   \stateThmEigenvaluesNoTag
\end{reptheorem}

\begin{proof}
We prove the theorem in two parts.

\paragraph{Dyson SDE.}
    We first prove \eqref{eq:SDE-spectrum}. This part of the proof is based on \cite{keatingRandomMatrixTheory2023} but generalizes it to arbitrary coefficients. For mathematical details on the $\alpha = \frac{1}{n}$, $\beta = 0$ case, see also \cite{andersonIntroductionRandomMatrices2009}. Suppose $M(t)$ satisfies the SDE  \cref{eq:SDE-mtx}. That is, with $M(0) \in \mathrm{Sym}(\R^{n\times n})$ we have
\begin{align*}
    \d M_{ij}(t) = - \beta M_{ij} \d t + D_{ij} \d B_{ij}
\end{align*} 
with $D_{ij} \coloneqq \sqrt{(1+\delta_{ij}) \alpha}$ for any constants $\alpha, \beta \in \R^+$ with $B_{ij}(t) = B_{ji}(t) \; \forall t>0$.
Let  $\lambda_1 \geq \ldots \geq \lambda_n$ be the eigenvalues. We choose as $v_1, \ldots, v_n$ an orthonormal basis of eigenvectors ($Mv_k = \lambda_k v_k$), which exists by the spectral theorem for symmetric matrices.

Due to symmetry, we may constrain the set of indices $(i,j)$ to $\mc I \coloneqq \{(i,j)\mid 1 \leq i \leq j \leq n\}$. For any $k\in [n]$, the eigenvalue $\lambda_k$ can thus be seen as a function of the set of Itô processes $\{M_\eta \mid \eta \in \mc I\}$. Hence, we have by Itô's Lemma,
\begin{align}
    \d \lambda_k 
    &= \underbrace{\frac{\partial \lambda_k}{\partial t} dt}_{\equiv 0} + \sum_{\eta \in \mathcal I} \frac{\partial \lambda_k}{\partial M_\eta}\d M_\eta + \frac{1}{2} \sum_{\eta,\xi \in \mc I}\frac{(\partial \lambda_k)^2}{\partial M_\eta \partial M_\xi} \d M_\eta \d M_\xi,\label{eq:eigenvalue-SDE-proof:ito-vanilla}
    \intertext{where the first part is $0$ since $\lambda_k(t)$ is only a function of the $M_{\eta}(t)$, not of time. By \cref{eq:SDE-mtx}, we get}
    &= \sum_{\eta \in \mc I}  \left( - \beta M_{\eta} \frac{\partial \lambda_k}{\partial M_\eta} +  \frac{1}{2}  D_\eta^2 \frac{\partial^2 \lambda_k}{(\partial M_\eta)^2} \right) \d t + D_\eta  \frac{\partial \lambda_k}{\partial M_\eta} \d B_{\eta}. \label{eq:eigenvalue-SDE-proof:ito-2}
\end{align}

It remains to calculate the partial derivatives. In what follows, we successively apply properties of the spectrum in order to obtain equations for the partial derivatives. We have
\begin{align}
     M v_k &= \lambda_k v_k  \\
     \intertext{taking the partial derivative with respect to $M_{ij}$ for $(i,j) \in \mc I$ on both sides and applying the product rule yields}
     \frac{\partial M}{\partial M_{ij}}  v_k + M \frac{\partial  v_k}{\partial M_{ij}} &= \frac{\partial \lambda_k}{\partial M_{ij}}  v_k + \lambda_k \frac{\partial  v_k}{\partial M_{ij}}. \label{eq:eigenvalue-SDE-proof:eig-prop-1}
\end{align}
Further, by orthogonality of the eigenvectors we have for any $k,l \in [n]$
\begin{align}
    v_k^T v_l &= \delta_{kl}\\
    \frac{\partial v_k^T}{\partial M_{ij}}  v_l + v_k^T \frac{\partial  v_l}{\partial M_{ij}} &=0. \label{eq:eigenvalue-SDE-proof:eig-prop-2}
    \intertext{which, taking the transpose, implies for $l=k$}
    \frac{\partial v_k^T}{\partial M_{ij}}  v_k &= v_k^T \frac{\partial  v_k}{\partial M_{ij}} =0. \label{eq:eigenvalue-SDE-proof:eig-prop-2:l-eq-k}
\end{align}

Multiplying \cref{eq:eigenvalue-SDE-proof:eig-prop-1} by $v_l^T$ from the left yields
\begin{align}
v_l^T \frac{\partial M}{\partial M_{ij}}  v_k + \underbrace{v_l^T M \frac{\partial  v_k}{\partial M_{ij}}}_{(I)} &= \underbrace{\frac{\partial \lambda_k}{\partial M_{ij}} v_l^T v_k}_{(II)} + \underbrace{\lambda_k v_l^T \frac{\partial  v_k}{\partial M_{ij}}}_{(III)}
\end{align}
where the terms simplify as follows.
\begin{align}
    (I) &= (1-\delta_{kl}) \lambda_l v_l^T \frac{\partial  v_k}{\partial M_{ij}} \tag{by \cref{eq:eigenvalue-SDE-proof:eig-prop-2:l-eq-k} } \\
    (II) &= \delta_{lk}  \frac{\partial \lambda_k}{\partial M_{ij}} \\
    (III) &=  (1-\delta_{kl}) \lambda_k v_l^T \frac{\partial  v_k}{\partial M_{ij}} \tag{by \cref{eq:eigenvalue-SDE-proof:eig-prop-2:l-eq-k} },
\end{align}
resulting in the following equation for $l=k$
\begin{align}
    v_k^T \frac{\partial M}{\partial M_{ij}}  v_k &= \frac{\partial \lambda_k}{\partial M_{ij}} \label{eq:eigenvalue-SDE-proof:ref-only:dlfkjsalk}
\end{align}
and for $l\neq k$:
\begin{align}
    v_l^T \frac{\partial M}{\partial M_{ij}}  v_k + \lambda_l v_l^T  \frac{\partial  v_k}{\partial M_{ij}} &= \lambda_k v_l^T \frac{\partial  v_k}{\partial M_{ij}} \\
    v_l^T \frac{\partial M}{\partial M_{ij}}  v_k &=  \left(\lambda_k - \lambda_l  \right) v_l^T \frac{\partial  v_k}{\partial M_{ij}}.  \label{eq:eigenvalue-SDE-proof:vl-transpose-partial-vk-partial-mij}
\end{align}

The matrix derivative is, trivially,
\begin{align}
	\left(\frac{\partial M}{\partial M_{ij}} \right)_{kl} = \begin{cases}
		1 & \text{for }(k,l) = (i,j) \text{ or }(l,k) = (i,j)\\
		0 & \text{else}
	\end{cases}, \label{eq:eigenvalue-SDE-proof:partial-M-partial-Mij}
\end{align}  
so that we can simplify \cref{eq:eigenvalue-SDE-proof:ref-only:dlfkjsalk} to
\begin{align}
   \frac{\partial \lambda_k}{\partial M_{ij}}  &= (v_k)_i (v_k)_j (2-\delta_{ij}) \label{eq:eigenvalue-SDE-proof:lambda-k-deriv}
\end{align}
and \cref{eq:eigenvalue-SDE-proof:vl-transpose-partial-vk-partial-mij} to
\begin{align}
    \left(\lambda_k - \lambda_l  \right) v_l^T \frac{\partial  v_k}{\partial M_{ij}} &=  (v_k)_i (v_l)_j + (1-\delta_{ij}) (v_k)_j (v_l)_i \label{eq:eigenvalue-SDE-proof:ref-only:qeqwweq}\\
    &= (v_l)_i (v_k)_j + (1-\delta_{ij}) (v_l)_j (v_k)_i.
\end{align}

Using that the $v_k$ are an ONB of eigenvectors, we can conclude with \cref{eq:eigenvalue-SDE-proof:lambda-k-deriv} that the first summand of \cref{eq:eigenvalue-SDE-proof:ito-2}  is
\begin{align}
    \sum_{\eta \in \mc I} - \beta M_\eta \frac{\partial \lambda_k }{\partial M_\eta }
    &= -\beta \lambda_k. \label{eq:eigenvalue-SDE-proof:for-ito-2:first-summand}
\end{align}

Since $\{v_1, \ldots, v_n\}$ is an orthonormal basis, we may project on the basis vectors as follows
\begin{align}
	\frac{\partial  v_k}{\partial M_{ij}} 
	&= \sum_{l \in [n]} v_l^T \frac{\partial  v_k}{\partial M_{ij}}  v_l \tag{projection into ONB basis}
    \intertext{applying \cref{eq:eigenvalue-SDE-proof:eig-prop-2:l-eq-k} gives}
	&=\sum_{l \in [n]\setminus \{k\}}  v_l^T \frac{\partial  v_k}{\partial M_{ij}}  v_l  \label{eq:eigenvalue-SDE-proof:partial-vk-partial-mij-in-vl-onb}\\
    \intertext{which yields with \cref{eq:eigenvalue-SDE-proof:ref-only:qeqwweq}}
	&=\sum_{l \in [n]\setminus \{k\}} \frac{1}{\lambda_k - \lambda_l} \left(( v_k)_i ( v_l)_j+ (1-\delta_{ij}) ( v_k)_j ( v_l)_i  \right)  v_l. \label{eq:eigenvalue-SDE-proof:eig-vec-prop-1-and-2:v-k-deriv}
\end{align}

To obtain the second order partial derivative of $\lambda_k$, we observe
\begin{align}
	\frac{\partial^2 \lambda_k}{\partial M_{ij} \partial M_{ij}} 
    =&\frac{\partial }{\partial M_{ij}} \left( ( v_k)_i ( v_k)_j (2- \delta_{ij}) \right) \tag{by \cref{eq:eigenvalue-SDE-proof:lambda-k-deriv}}\\
	=& (2- \delta_{ij}) \left(\frac{\partial ( v_k)_i}{\partial M_{ij}} ( v_k)_j + ( v_k)_i \frac{\partial ( v_k)_j }{\partial M_{ij}} \right) \nonumber\\
    =& (2-\delta_{ij}) \sum_{l \in [n]\setminus \{k\}} \frac{1}{\lambda_k - \lambda_l}
 \bigg( ( v_l)_i ( v_k)_j ( v_l)_i ( v_k)_j+ ( v_l)_j ( v_k)_i (1-\delta_{ij})( v_l)_i ( v_k)_j \notag \\
	&+ ( v_l)_i ( v_k)_j ( v_l)_j ( v_k)_i+ ( v_l)_j ( v_k)_i (1-\delta_{ij}) ( v_l)_j ( v_k)_i \bigg) 
	 \tag{by \cref{eq:eigenvalue-SDE-proof:eig-vec-prop-1-and-2:v-k-deriv}}\\
	 =& (2-\delta_{ij}) \sum_{l \in [n]\setminus \{k\}} \frac{1}{\lambda_k - \lambda_l} 
	 \bigg(( v_l)_i^2 ( v_k)_j^2 + ( v_l)_j^2 ( v_k)_i^2 (1- \delta_{ij}) \notag \\
     & + ( v_l)_i ( v_l)_j ( v_k)_i ( v_k)_j (2-\delta_{ij}) \bigg) .\label{eq:eigenvalue-SDE-proof:lambda-k-second-deriv}
\end{align}

These second order partial derivative are summed over $\mc I$ in \cref{eq:eigenvalue-SDE-proof:ito-2}, so that combining \cref{eq:eigenvalue-SDE-proof:lambda-k-second-deriv} with the definition of $D_{ij}$ yields
\begin{align}
    \sum_{(i,j) \in \mc I} D_{ij}^2  \frac{\partial^2 \lambda_k}{(\partial M_{ij})^2} 
    &= \sum_{j=1}^n \sum_{i=1}^j  \alpha (1+\delta_{ij})  (2-\delta_{ij}) \sum_{l \in [n]\setminus \{k\}} \frac{1}{\lambda_k - \lambda_l} 
	 \bigg(( v_l)_i^2 ( v_k)_j^2 \notag \\
     & + ( v_l)_j^2 ( v_k)_i^2 (1- \delta_{ij}) + ( v_l)_i ( v_l)_j ( v_k)_i ( v_k)_j (2-\delta_{ij}) \bigg)
     \intertext{we reorder and note that since the summand is symmetric in $i,j$, we may change the range of summation and absorb the coefficient $2-\delta_{ij}$,}
    &=  \alpha \sum_{l \in [n]\setminus \{k\}} \frac{1}{\lambda_k - \lambda_l}  \sum_{j=1}^n \sum_{i=1}^n  (1+\delta_{ij})
	 \bigg(( v_l)_i^2 ( v_k)_j^2 \notag \\
     & + ( v_l)_j^2 ( v_k)_i^2 (1- \delta_{ij}) + ( v_l)_i ( v_l)_j ( v_k)_i ( v_k)_j (2-\delta_{ij}) \bigg)
     \intertext{where we realize that upon accounting for all $\delta_{ij}$, the first two summands are simply $ \norm{v_l}_2^2  \norm{v_k}_2^2$, while the third summand may be written an inner product}
&=  \alpha \sum_{l \in [n]\setminus \{k\}} \frac{1}{\lambda_k - \lambda_l}\left(2 \norm{v_l}_2^2  \norm{v_k}_2^2  +  2(\underbrace{v_l^T v_k}_{0})^2\right) 
\intertext{since we chose an ONB, this simplifies to}
&= 2  \alpha \sum_{l \in [n]\setminus \{k\}} \frac{1}{\lambda_k - \lambda_l}. \label{eq:eigenvalue-SDE-proof:for-ito-2:second-summand}
\end{align}

It remains to determine an explicit expression of the Brownian motion in \cref{eq:eigenvalue-SDE-proof:ito-2}. We have

\begin{align}
    \sum_{\eta \in \mc I} D_\eta \frac{\partial \lambda_k}{\partial M_\eta} \d B_\eta 
    &= \sum_{(i,j) \in \mc I} \sqrt{(1+\delta_{ij}) \alpha } (2-\delta_{ij}) (v_k)_i (v_k)_j  \d B_{ij} \nonumber
    \intertext{using that $B_{ij}(t) = B_{ji}(t)$, we obtain}
    &=  \sqrt{2\alpha }\sum_{i=1}^n \sum_{j=1}^n \sqrt{\frac{1+\delta_{ij}}{2}} (v_k)_i (v_k)_j  \d B_{ij} \nonumber
    \intertext{we may now define $\d \tilde B_{k}$ as follows}
    &= \sqrt{2\alpha } \d \tilde B_{k}.
\end{align}

Indeed, the set $\{\tilde B_{k} \mid k \in [n]\}$ is in distribution equal $n$ independent standard Brownian motions, since  $\E{\d \tilde B_k} = 0$ and for $k, l \in [n]$
\begin{align}
    \E{ \d \tilde B_k \d \tilde B_l}
    &=  \E{ \frac{1}{2}\sum_{ij} \sum_{st} \sqrt{1+\delta_{ij}}\sqrt{1+\delta_{st}} (v_k)_i (v_k)_j (v_l)_s (v_l)_t  \d B_{ij}  \d B_{st} } \nonumber
    \intertext{where we realize that the product of the differentials is $0$ except for $(i,j) = (s,t)$ and $(i,j) = (t,s)$}
    &= \E{v_k^T v_l v_k^T v_l }\d t\nonumber \\
    &= \delta_{kl} \d t.
\end{align}

We can thus conclude that by \cref{eq:eigenvalue-SDE-proof:ito-2} we have,

\begin{align}
    \d \lambda_k = \left( - \beta \lambda_k +  \alpha \sum_{l\in [n] \setminus \{k\}} \frac{1}{\lambda_k - \lambda_l} \right) \d t + \sqrt{2 \alpha} \d W_k
\end{align}

where $\{W_k \mid k \in [n]\}$ are $n$ independent Brownian motions.

\paragraph{Invariant Distribution.}

We now show that \begin{align}
     	p_{\mathrm{inv}}( \lambda)  = \frac{1}{Z} \exp(-U( \lambda)) \quad\quad \text{ for } \quad\quad U( \lambda) = \frac{\beta}{2 \alpha } \sum_k \lambda_k^2 - \sum_{k<\ell } \ln|\lambda_k-\lambda_\ell|,
     \end{align} for $ \lambda \in C_n$ and $Z$ a normalizing constant so that $p_{\mathrm{inv}}$ is a probability measure, is the invariant distribution. We use the Fokker-Planck equation. Indeed, recall that if we have an SDE $$\d \lambda_t =  f(\lambda_t) \d t +  L(\lambda_t) \d B_t$$ in dimension $n$, where $f(\lambda) = (f_1(\lambda), \ldots, f_n(\lambda))$ is a $C^2$-function, $L(\lambda)$ is matrix-valued $C^2$-function and $B_t$ is a $n$-dimensional Brownian motion, then an invariant distribution $p(x)$ satisfies 
     \begin{equation}\label{FokkerPlank}
         \sum_{i = 1}^n \frac{\partial}{\partial \lambda_i} [f_i(\lambda)p(\lambda)] = \frac{1}{2} \sum_{i,j = 1}^n \frac{\partial^2}{\partial \lambda_i \partial \lambda_j} [ L(\lambda) L(\lambda)^T]_{ij} p(\lambda).
     \end{equation}

In our case, for the $\lambda(\alpha,\beta)$-SDE we have that $L(\lambda,t) = \sqrt{2\alpha} 1_n$ and therefore $[L(\lambda,t) L(\lambda,t)^T]_{ij} = 2\alpha \delta_{ij}$. So the right hand side of \eqref{FokkerPlank} equals $$\alpha \sum_{i = 1}^n \frac{\partial^2}{(\partial \lambda_i)^2} p(\lambda).$$ 
Assume for now that $\lambda = (\lambda_1, \ldots , \lambda_n)$ satisfies $\lambda_1 > \lambda_2 > \ldots > \lambda_n$ and $$U(\lambda) = c\sum_i \lambda_i^2 - \sum_{i < j} \ln (\lambda_i - \lambda_j)$$ for some constant $c > 0$ to be determined. We first calculate for a fixed $i$, 
\begin{align*}
    \frac{\partial}{\partial \lambda_{i}} p(\lambda) &= -\frac{1}{Z}\exp(-U(\lambda)) \frac{\partial}{\partial \lambda_{i}} U(\lambda) \\
    &= -\frac{1}{Z}\exp(-U(\lambda))\left( 2c\lambda_{i} - 
    \sum_{i < j} \frac{1}{\lambda_{i} - \lambda_j} + \sum_{j < i} \frac{1}{\lambda_j - \lambda_{i}} \right) \\
    &= -\frac{1}{Z}\exp(-U(\lambda))\left( 2c\lambda_{i} -  \sum_{j \neq i} \frac{1}{\lambda_{i} - \lambda_j} 
    \right).
\end{align*} Therefore, 
\begin{align*}
    \frac{\partial^2}{(\partial \lambda_{i})^2} p(\lambda)  
    &= \frac{1}{Z}\exp(-U(\lambda))\left( 2c\lambda_{i} -  \sum_{j \neq i} \frac{1}{\lambda_{i} - \lambda_j} 
    \right)^2 -\frac{1}{Z}\exp(-U(\lambda)) \left( 2c + \sum_{j \neq i} \frac{1}{(\lambda_{i} - \lambda_j)^2} \right) \\
    &= \left(  \sum_{j \neq i} \frac{1}{\lambda_{i} - \lambda_j} 
     -  2c\lambda_{i} \right)\frac{\partial}{\partial \lambda_{i}} p(\lambda) - p(\lambda)\left( 2c + \sum_{j \neq i} \frac{1}{(\lambda_{i} - \lambda_j)^2} \right)
\end{align*} and so the right hand side of \eqref{FokkerPlank} is equal to $$\alpha \sum_{i = 1}^n \frac{\partial^2}{(\partial \lambda_i)^2} p(\lambda) = \sum_{i = 1}^n\left(  \alpha\sum_{j \neq i} \frac{1}{\lambda_{i} - \lambda_j} 
     -  2\alpha c\lambda_{i} \right)\frac{\partial}{\partial \lambda_{i}} p(\lambda) - \sum_{i = 1}^np(\lambda)\left( 2\alpha c + \alpha\sum_{j \neq i} \frac{1}{(\lambda_{i} - \lambda_j)^2} \right)$$

Now the left hand side of \eqref{FokkerPlank} is equal to
\begin{align*}
    & \sum_{i = 1}^n \frac{\partial}{\partial \lambda_{i}} \left[ \left(\alpha \sum_{i\neq j} \frac{1}{\lambda_{i} - \lambda_j} - \beta \lambda_{i} \right)p(\lambda)  \right] \\
    = &\sum_{i = 1}^n \left(\alpha \sum_{i\neq j} \frac{1}{\lambda_{i} - \lambda_j} - \beta \lambda_{i} \right)\frac{\partial}{\partial \lambda_{i}} p(\lambda) + \sum_{i = 1}^n \left(-\alpha \sum_{i\neq j} \frac{1}{(\lambda_{i} - \lambda_j)^2} - \beta  \right)p(\lambda)
\end{align*}

So it follows that in \eqref{FokkerPlank} the left-hand side is equal to the right-hand side if and only if $\beta = 2\alpha c$ or equivalently $c = \beta/2\alpha$, concluding the proof. \end{proof}

\begin{figure}[H]
\centering
    \includegraphics[width = 0.3\textwidth]{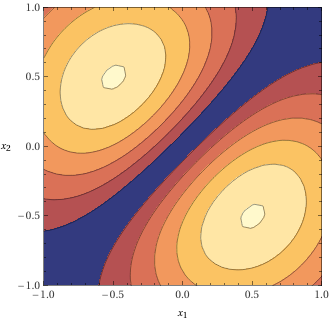}

    \caption{\centering Plot of the invariant density of \eqref{eq:SDE-spectrum} for $d = 2$ and $\alpha = \beta = 1$}
\end{figure}

\section{Inferring the Eigenvector Dynamics} \label{sec:app:eigenvector-SDE-proof}

\begin{reptheorem}{thm:eigenvector-SDE}
   \stateThmEigenvectorsNoLabel
\end{reptheorem}

\begin{proof}
    Analogously to the proof of \Cref{thm:eigenvalue-SDE}, we may view the $v_k$ for $k\in [n]$ as a function of the matrix components $M_{ij}$ for $(i,j) \in \mc I\coloneqq \{(i,j): 1 \leq i \leq j \leq n\}$. We thus have by Itô's lemma 
    \begin{align}
        \d v_k 
        &= \sum_{\eta \in \mc I} \frac{\partial v_k}{\partial M_\eta}\d M_\eta + \frac{1}{2} \sum_{\eta, \xi \in \mc I} \frac{\partial^2 v_k}{\partial M_\eta \partial M_\xi} \d M_\eta \d M_\xi\nonumber\\
        &= \sum_{\eta \in \mc I} \left(-\beta M_\eta \frac{\partial v_k}{\partial M_\eta} + \frac{1}{2} D_\eta^2  \frac{\partial^2 v_k}{(\partial M_\eta)^2}  \right) dt + D_\eta \frac{\partial v_k}{\partial M_\eta} \d B_\eta. \label{eq:app:eigenvector-derivation:post-ito-3-terms}
    \end{align}

    For the first summand of \cref{eq:app:eigenvector-derivation:post-ito-3-terms}, we observe, using \cref{eq:eigenvalue-SDE-proof:partial-vk-partial-mij-in-vl-onb}, that
    \begin{align}
         &\sum_{(i,j) \in \mc I} M_{ij} \frac{\partial v_k}{\partial M_{ij}} \d t=\sum_{l \neq k}  \sum_{(i,j) \in \mc I} v_l^T \frac{\partial v_k}{\partial M_{ij}} v_l M_{ij} \d t 
        \intertext{and further use \cref{eq:eigenvalue-SDE-proof:vl-transpose-partial-vk-partial-mij}}
        =& \sum_{l \neq k}  \frac{1}{\lambda_k - \lambda_l}  \Big( v_l^T \underbrace{\sum_{(i,j) \in \mc I}  \frac{\partial M}{\partial M_{ij}} M_{ij}}_{= M \text{ by \cref{eq:eigenvalue-SDE-proof:partial-M-partial-Mij}}} v_k \Big) v_l \d t \nonumber \\
        =& \sum_{l \neq k}  \frac{\lambda_k}{\lambda_k - \lambda_l}  (v_l^T v_k) v_l \d t.\\
        \intertext{Since $l\neq k$ and the $\{v_l \mid l \in [n]\}$ are orthogonal, we get}
        =& 0.
    \end{align}

    For the second summand of \cref{eq:app:eigenvector-derivation:post-ito-3-terms}, we observe
    \begin{align}
        &\frac{\partial^2 v_k }{(\partial M_\eta)^2 }  = \frac{\partial}{\partial M_\eta } \frac{\partial v_k }{\partial M_\eta } \\
        \intertext{and using
        \Cref{eq:eigenvalue-SDE-proof:vl-transpose-partial-vk-partial-mij} and
\Cref{eq:eigenvalue-SDE-proof:partial-vk-partial-mij-in-vl-onb} we get        
        }
        =& \sum_{m \neq k} \frac{\partial}{\partial M_\eta } \left[ \frac{1}{\lambda_k - \lambda_m} v_m^T \frac{\partial M}{\partial M_\eta} v_k v_m  \right] \\
\intertext{by noting $\frac{\partial}{\partial M_\eta} \frac{\partial M}{\partial M_\eta} = 0$, we apply the chain rule to obtain}
        =& \sum_{m \neq k}  \Bigg\{ 
        - \underbrace{\left(\frac{1}{\lambda_k - \lambda_m}\right)^2 \left(\frac{\partial \lambda_k}{\partial M_\eta}  - \frac{\partial \lambda_m}{\partial M_\eta}   \right) v_m^T \frac{\partial M}{\partial M_\eta} v_k v_m}_{(i)}
        +  \underbrace{\frac{1}{\lambda_k - \lambda_m} \frac{\partial v_m^T}{\partial M_\eta} \frac{\partial M}{\partial M_\eta} v_k v_m}_{(ii)} 
         \\
        & \quad  
        +  \underbrace{\frac{1}{\lambda_k - \lambda_m} v_m^T \frac{\partial M}{\partial M_\eta} \frac{\partial v_k}{\partial M_\eta} v_m }_{(iii)}
        +  \underbrace{\frac{1}{\lambda_k - \lambda_m} v_m^T \frac{\partial M}{\partial M_\eta} v_k \frac{\partial{v_m}}{\partial M_\eta}}_{(iv)} 
        \Bigg\}.
    \end{align}
We now analyze each term.

\paragraph{Term (i)}

We recall \Cref{eq:eigenvalue-SDE-proof:lambda-k-deriv}
\begin{align}
   \frac{\partial \lambda_k}{\partial M_{ij}}  &= (v_k)_i (v_k)_j (2-\delta_{ij}),\nonumber
\end{align}
as well as that the partial derivative $\frac{\partial M}{\partial M_{ij}} $ is $0$ except at $i,j$ and $j,i$, where it is $1$ (using symmetry).

Pulling the summation over $\eta$ and all $\eta$-dependent terms in, we have 
\begin{align}
    &\sum_{\eta\in \mc I} (1+\delta_\eta) \left(\frac{\partial \lambda_k}{\partial M_\eta}  - \frac{\partial \lambda_m}{\partial M_\eta}   \right) v_m^T \frac{\partial M}{\partial M_\eta} v_k \nonumber\\
    =& \sum_{ij \in \mc I} (1+\delta_{ij}) 
    \Bigg\{\left[(v_k)_i (v_k)_j (2-\delta_{ij}) - (v_m)_i (v_m)_j (2-\delta_{ij}) \right] \nonumber\\
    & \qquad \left[ (v_m)_i (v_k)_j + (v_m)_j (v_k)_i (1-\delta_{ij}) \right]
    \Bigg\}\nonumber \\
    =& 2 \left[v_k v_k^T v_k^T v_m - v_m^T v_m  v_m^T v_k \right] \nonumber\\
    =& 2 (\delta_{km} - \delta_{km})\nonumber\\
    =& 0\nonumber.
\end{align}
Thus, the overall contribution of term $(i)$ is $0$.

\paragraph{Term (ii)}
For term (ii) we have by
        \Cref{eq:eigenvalue-SDE-proof:vl-transpose-partial-vk-partial-mij} and
\Cref{eq:eigenvalue-SDE-proof:partial-vk-partial-mij-in-vl-onb} 
\begin{align}
    (ii) 
    =& \frac{1}{\lambda_k - \lambda_m} 
    \sum_{s \neq m} \frac{1}{\lambda_m - \lambda_s}   v_m^T  \frac{\partial M}{\partial M_\eta} v_s v_s^T
    \frac{\partial M}{\partial M_\eta} v_k v_m.
\end{align}

Using
\begin{align}
    v_m^T  \frac{\partial M}{\partial M_{ij}}v_s = (v_m)_i (v_s)_j + (v_m)_j (v_s)_i (1-\delta_{ij}),
\end{align}

we note that 
\begin{align}
   & \sum_{\eta \in \mc I} D_\eta^2  v_m^T  \frac{\partial M}{\partial M_\eta} v_s v_s^T
    \frac{\partial M}{\partial M_\eta} v_k \nonumber\\
    =& \alpha \sum_{i,j \in \mc I}  (1+ \delta_{ij}) \left[(v_m)_i (v_s)_j + (v_m)_j (v_s)_i (1-\delta_{ij}) \right] \left[ (v_k)_i (v_s)_j + (v_k)_j (v_s)_i (1-\delta_{ij})\right]\nonumber \\
    =& \alpha \left\{ \left[ \sum_i (v_m)_i (v_k)i \right] \left[\sum_i (v_s)_i^2 \right] + v_m^Tv_s v_s^T v_k \right\}\nonumber \\
    =&\alpha  \delta_{km} + \alpha \delta_{ms} \delta_{sk} \label{eq:app:eigenvector-derivation:dira-delta-simplification} \\
    =& 0.\nonumber
\end{align}
Thus, the contribution of term $(ii)$ vanishes.

\paragraph{Term (iii)}
Analogous to term $(ii)$. 

\paragraph{Term (iv)}
By using \Cref{eq:eigenvalue-SDE-proof:vl-transpose-partial-vk-partial-mij} and
\Cref{eq:eigenvalue-SDE-proof:partial-vk-partial-mij-in-vl-onb}, we get  
\begin{align}
    (iv) =&\frac{1}{\lambda_k - \lambda_m} v_m^T \frac{\partial M}{\partial M_\eta} v_k \frac{\partial{v_m}}{\partial M_\eta} \nonumber\\
    =& \frac{1}{\lambda_k - \lambda_m} v_m^T \frac{\partial M}{\partial M_\eta} v_k  \sum_{p \neq m} \frac{1}{\lambda_m - \lambda_p} v_p^T \frac{\partial M}{\partial M_\eta} v_m v_p \nonumber\\
    =& \frac{1}{\lambda_k - \lambda_m} 
    \sum_{p \neq m}  \frac{1}{\lambda_m - \lambda_p}  \Big( v_m^T \frac{\partial M}{\partial M_\eta} v_k \Big) \Big( v_p^T \frac{\partial M}{\partial M_\eta} v_m \Big) v_p.\nonumber
\end{align}

By pulling in the $\eta$-dependent terms from \Cref{eq:app:eigenvector-derivation:post-ito-3-terms} which affect term $(iv)$, and then by subsequently pulling the sum over $\eta$ in, we get 
\begin{align}
    &\sum_{\eta \in \mc I} \frac{1}{2} D_\eta^2 v_m^T \frac{\partial M}{\partial M_\eta} v_k  v_p^T \frac{\partial M}{\partial M_\eta} v_m \\
    \intertext{which, using \cref{eq:app:eigenvector-derivation:dira-delta-simplification}, becomes }
    =&\frac{\alpha}{2} \left(\delta_{kp} + \delta_{km} \delta_{mp} \right)\nonumber\\
    =& \frac{\alpha}{2} \delta_{kp}.\nonumber
\end{align}

Moreover, using that $p \neq m$, we can conclude that the total contribution of term $(iv)$ is
\begin{align}
    &\sum_{\eta \in \mc I} \frac{1}{2} D_\eta^2 \sum_{m \neq k} (iv) \nonumber\\
    =& -\frac{\alpha}{2} \sum_{m \neq k} \frac{1}{(\lambda_k - \lambda_m)^2} v_k. \nonumber
\end{align}

Before proceeding, we prove the following lemma.

\begin{lemma}\label{lem:BM-orthonormal-trafo}
	Let $ v^{(k)}$ be a set of orthonormal vectors and $B_{ij}$ a set of independent standard Brownian motions with $B_{ij} = B_{ji}$, for $i,j,k \in [n]$. We have that 
	$$\tilde B_{lk}:= ( v^{(l)})^T \begin{pmatrix}
		\sqrt{2}B_{11} & B_{12} & B_{13} & \ldots & B_{1n} \\
		B_{21} & \sqrt{2} B_{22} & B_{23} & \ldots & B_{2n} \\
		\vdots &  & \vdots & & \vdots \\
		B_{n1} & B_{n2} & B_{n3 }& \ldots & \sqrt{2} B_{nn} \\
	\end{pmatrix}  v^{(k)} $$
	is a Brownian motion with variance $1+\delta_{lk}$, that is $\tilde B_{lk}  \distreq \sqrt{1 + \delta_{lk}} B$ for $B$ a standard Brownian motion. $\tilde B_{lk}$ is independent of $\tilde B_{ab}$ for $(l,k) \neq (a,b)$ and $(l,k) \neq (b,a)$.
\end{lemma}
\begin{proof}
	We have for any $a,b \in [n]$ that 
	$$\d \tilde B_{ab}=\sum_{i,j} ( v^{(a)})_i ( v^{(b)})_j \sqrt{1+\delta_{ij}} \d B_{ij}.$$ 
	Thus, $\E{\d \tilde B_{ab}}=0$. We further have for $c,d \in [n]$:
	\begin{align}
		\E{\d \tilde B_{ab} \d \tilde B_{cd}} 
		&= \sum_{ijkl} 
		( v^{(a)})_i ( v^{(b)})_j  
		( v^{(c)})_k ( v^{(d)})_l
		\sqrt{1+\delta_{ij}} \sqrt{1+\delta_{kl}} 
		\E{\d B_{ij} \d B_{kl}} \nonumber\\
		&=  \sum_{ijkl} 
		( v^{(a)})_i ( v^{(b)})_j  
		( v^{(c)})_k ( v^{(d)})_l
		\sqrt{1+\delta_{ij}} \sqrt{1+\delta_{kl}} 
		\1_{(i,j) = (k,l) \lor (i,j) = (l,k)} \d t \nonumber\\
		&= \sum_{ij} 
		( v^{(a)})_i ( v^{(b)})_j  
		\left\{
			\1_{i \neq j}
			\left[ 
				( v^{(c)})_i ( v^{(d)})_j
				+ ( v^{(c)})_j ( v^{(d)})_i
			\right]
			+ \1_{i = j} ( v^{(c)})_i ( v^{(d)})_i
		\right\}
		(1+\delta_{ij}) \d t \nonumber\\
		&= \sum_{ij} 
		( v^{(a)})_i ( v^{(b)})_j  
		\left[
				( v^{(c)})_i ( v^{(d)})_j
				+ ( v^{(c)})_j ( v^{(d)})_i
		\right] \d t \nonumber\\
		&= \sum_{ij} 
		( v^{(a)})_i ( v^{(b)})_j 
				( v^{(c)})_i ( v^{(d)})_j \d t 
		+\sum_{ij} 
		( v^{(a)})_i ( v^{(b)})_j
		 ( v^{(c)})_j ( v^{(d)})_i \d t \nonumber\\
		 &= ( v^{(a)})^T  v^{(c)} ( v^{(b)})^T  v^{(d)}\d t + ( v^{(a)})^T  v^{(d)} ( v^{(b)})^T  v^{(c)} \d t\nonumber\\
		 &= \1_{(a,b) = (c,d)}\d t + \1_{(a,b) = (d,c)}\d t
         \label{eq:app:brownian-motion-trafo:cov-expr}
	\end{align}
	
	Thus, in particular the process $\tilde B_{ab}$ has variance $t$ for $a \neq b$ and variance $2t$ for $a=b$.  

    Since the linear combination of independent Brownian motions is jointly normal, we see from the Covariance property in \cref{eq:app:brownian-motion-trafo:cov-expr} that $\tilde B_{lk}$ is independent of $\tilde B_{ab}$ for $(l,k) \neq (a,b)$ and $(l,k) \neq (b,a)$.
\end{proof}

We can now turn to the third summand of \cref{eq:app:eigenvector-derivation:post-ito-3-terms}.

We have 
\begin{align}
   &\sum_{\eta \in \mc I} D_\eta \frac{\partial v_k}{\partial M_\eta} \d B_\eta \nonumber\\
   =& \sqrt{\alpha} \sum_{l \neq k} \frac{1}{\lambda_k - \lambda_l} v_l^T \sum_{\eta \in \mc I} \frac{\partial M}{\partial M_{\eta}} v_k  \sqrt{1+\delta_{\eta}} \d B_\eta v_l \nonumber\\
   =&  \sqrt{\alpha} \sum_{l \neq k} \frac{1}{\lambda_k - \lambda_l} v_l^T \begin{pmatrix}
		\sqrt{2}\d B_{11} & \d B_{12} &  \d B_{13} & \ldots & \d B_{1n} \\
		\d B_{21} & \sqrt{2} \d B_{22} &  \d B_{23} & \ldots &\d  B_{2n} \\
		\vdots &  & \vdots & & \vdots \\
		\d B_{n1} & \d B_{n2} & \d B_{n3 }& \ldots & \sqrt{2} \d B_{nn} \\
	\end{pmatrix}  v_k v_l  \nonumber\\
    =&  \sqrt{\alpha} \sum_{l \neq k} \frac{1}{\lambda_k - \lambda_l} v_l \d \tilde B_{lk}
\end{align}
where we used \Cref{lem:BM-orthonormal-trafo} in the last step so that $\d \tilde B_{lk}$ are symmetric Brownian motions with variance $1+\delta_{lk}$.

Having established all the terms in the SDE in Theorem~\ref{thm:eigenvector-SDE}, we check that this dynamics of the eigenvectors gives rise to normalized vectors, assuming that the initial vectors $(v_1(0),\dots,v_n(0))$ are normalized. To this end, it suffices to note that $v_k(t+\d t)=v_k(t)+dv_k(t)$ is normalized given that $v_k(t)$ is normalized, that is $(v_k(t))^2=1$ for any $t$. By continuity of $t\mapsto v_k(t)$, it suffices to show that  $v_k(\d t)=v_k(0)+dv_k(0)$ remains normalized. We use the SDE from Theorem \ref{thm:eigenvector-SDE} to compute that up to terms of order $\d t^{3/2}$ or higher we have (omitting ``(0)'' in the notation)
\begin{align}
&    (v_k(0)+\d v_k(0))^2 = 1 + 2v_k^T\d v_k+(\d v_k)^2\nonumber\\
    =& 1+2\left(\frac{-\alpha}2\right)\sum_{l\ne k}\frac{\d t}{(\lambda_k-\lambda_l)^2}+2\sqrt{\alpha}\sum_{l\ne k}\frac1{\lambda_l-\lambda_k}\underbrace{v_k^Tv_l}_{=0}\d w_{lk} \nonumber\\
    &+\alpha\sum_{l\ne k}\sum_{j\ne k}\frac{\d w_{lk}\d w_{jk}}{(\lambda_l-\lambda_k)(\lambda_j-\lambda_k)}+\mathcal O(\d t^{3/2})\nonumber\\
    =&1-\alpha\sum_{l\ne k}\frac{\d t}{(\lambda_k-\lambda_l)^2}+\alpha\sum_{l\ne k}\sum_{j\ne k}\frac{\delta_{jl}\d t}{(\lambda_k-\lambda_l)(\lambda_k-\lambda_j)} + \mathcal O(\d t^{3/2})\nonumber\\
    =& 1 + \mathcal O(\d t^{3/2}).
\end{align}
Since contributions of $\mathcal O(\d t^{3/2})$ do not contribute to the dynamics for $\d t\to 0$, this implies that the $v_k(\d t)$ remain normalized. The argument may be iterated, i.e., $(v_k(2\d t))^2=(v_k(\d t)+\d v_k(\d t))^2=1 + \mathcal O(\d t^{3/2})$ and so on, and thus the $v_k(t)$ remain normalized for all $t>0$.

Finally, as an instructive consistency check, we also show that the eigenvectors remain orthogonal to each other (as they must, since $M$ is symmetric for all $t\ge 0$). This can be checked from the SDE in an analogous manner using again the continuity of $t\mapsto v_k(t)$. 
 Note that for $k\ne l$ and given $v_k^T(0)v_l(0)=0$, it suffices to show that $v_k^T(\d t)v_l(\d t)=0$. Indeed, we have (omitting ``(0)'' in the notation) 
\begin{align}
    & (v_k(0) +\d v_k(0))^T(v_l(0)+\d v_l(0)) = 0 + v_k^Td v_l+ v_l^Td v_k+ \d v_k^Td v_l\nonumber\\
    =&\sqrt{\alpha}\sum_{i\ne l}\frac1{\lambda_l-\lambda_i}v_k^Tv_i\d w_{il}+\sqrt{\alpha}\sum_{j\ne k}\frac1{\lambda_k-\lambda_j}v_l^Tv_j\d w_{jk} \nonumber\\
    &+ \alpha\sum_{j\ne k}\sum_{i\ne l}\frac{v_j^Tv_i \d w_{il}\d w_{jk}}{(\lambda_k-\lambda_j)(\lambda_l-\lambda_i)} + \mathcal O(\d t^{3/2}) \nonumber\\
    =&\frac{\sqrt{\alpha}}{\lambda_l-\lambda_k}\underbrace{(\d w_{kl}-\d w_{lk})}_{=0} + \alpha\sum_{j\ne k}\sum_{i\ne l}\frac{\delta_{ij}\delta_{ik}\delta_{jl}\d t}{(\lambda_k-\lambda_j)(\lambda_l-\lambda_i)} + \mathcal O(\d t^{3/2}) \nonumber\\
    =&0+\mathcal O(\d t^{3/2}).
\end{align}
The argument may be iterated and thus the claim holds for all $t>0$.

\end{proof}

\section{Time reversal: Existence and Uniqueness} \label{sec:app:time-reversal}
The time-reversal of the \eqref{eq:SDE-spectrum} in the sense of Ref.~\cite{andersonReversetimeDiffusionEquation1982} is given in Eq.~\eqref{eq:dyson-timereversal}. Since the drift coefficient in the \eqref{eq:SDE-spectrum} is not locally Lipschitz continuous, the existence and uniqueness of strong solutions to \eqref{eq:SDE-spectrum} and Eq.~\eqref{eq:dyson-timereversal}, and the applicability of Ref.~\cite{andersonReversetimeDiffusionEquation1982} are not obvious. While the existence and uniqueness of a strong solution to the \eqref{eq:SDE-spectrum} is well established (see, e.g., Lemma 4.3.3 in Ref.~\cite{andersonIntroductionRandomMatrices2009}), we here sketch how to ensure the other two points.

To ensure existence and uniqueness of a strong solution to Eq.~\eqref{eq:dyson-timereversal}, repeat the arguments in Lemma 4.3.3 in Ref.~\cite{andersonIntroductionRandomMatrices2009}, where the divergent $x^{-1}$-term in the drift is replaced by the locally Lipschitz continuous approximation $\phi(x)=x^{-1}$ for $|x|\ge R^{-1}$ and $\phi(x)=xR^2$ otherwise. For any $R>0$, the desired statements follow from the local Lipschitz continuity of the drift, and details on the limit $R\to 0$ can be found in Ref.~\cite{andersonIntroductionRandomMatrices2009}. The only difference to the forward motion is the additional drift term containing the score function. However, since the dynamics is almost surely contained in the interior of the Weyl~chamber~\citep{katoriNoncollidingBrownianMotions2003}, the propagator in the score-contribution is, as usual, dominated by white noise as $\d t\to 0$. Therefore, this term will not cause complications as $R\to 0$ and the arguments from Ref.~\cite{andersonIntroductionRandomMatrices2009} imply uniqueness and existence of a strong solution to Eq.~\eqref{eq:dyson-timereversal}.

Using the same regularization of the divergent term, for any $R>0$ the statements of Ref.~\cite{andersonReversetimeDiffusionEquation1982} are directly applicable. Since this regularization is piecewise continuous, we can take the limit $R\to 0$ under time reversal. Since we just established the existence of the solution to Eq.~\eqref{eq:dyson-timereversal}, the limit $R\to 0$ converges.
Furthermore, the arguments are expected to generalize (regularize $\phi(x) =x^{-2}$) to the eigenvector equations and their time reversal.

\section{Time rescaling} \label{sec:app:time-rescaling}

We rescale time for \eqref{eq:SDE-spectrum}.  Let $T(s)$ be a continuous, differentiable rescaling of time, monotonically increasing, with $T(0) = 0$ (for convenience).

Let $\gamma_k(t) := \lambda_k(T(t))$. With this definition, the Ito SDE \cref{eq:SDE-spectrum} corresponds to the Ito integral
\begin{align}
    \gamma_k(t) 
    :=& \lambda_k(T(t)) = \lambda_k(0) + \int_0^{T(t)} \left( \alpha \sum_{i \neq k} \frac{1}{\lambda_k(s) - \lambda_i(s)}-\beta \lambda_k(s) \right) \d s  + \int_0^{T(t)} \sqrt{2 \alpha } \d B_k(s)\nonumber\\
    =& \int_0^t  \left( \alpha \sum_{i \neq k} \frac{1}{\lambda_k(T(s)) - \lambda_i(T(s))}-\beta \lambda_k(T(s)) \right) T'(s) \d s + \int_0^t \sqrt{2 \alpha T'(s) } \d B_k(s) \tag{using Thm.\ 8.5.7 in \cite{oksendalStochasticDifferentialEquations2003}} \\
    =&  \int_0^t  \left( \alpha \sum_{i \neq k} \frac{1}{\gamma_k(s) - \gamma_i(s)}-\beta \gamma_k(s) \right) T'(s) \d s + \int_0^t \sqrt{2 \alpha T'(s) } \d B_k(s)
\end{align}

which we can write in SDE notation
\begin{align}
    \d \gamma_k(t) = \left( \alpha \sum_{i \neq k} \frac{1}{\gamma_k(t) - \gamma_i(t)}-\beta \gamma_k(t) \right) T'(t) \d t+ \sqrt{2 \alpha T'(t)} \d B_k(t).
 \end{align}

By using $T(t):= \frac{1}{\alpha} t$, we obtain 
\begin{align}
    \d \gamma_k(t) = \left(\sum_{i \neq k} \frac{1}{\gamma_k(t) - \gamma_i(t)}- \frac{\beta}{\alpha} \gamma_k(t) \right) \d t+ \sqrt{2} \d B_k(t),
 \end{align}
so that we can summarize the two parameters to $\eta := \frac{\beta}{\alpha}$:
\begin{align}
    \d \gamma_k(t) = \left(\sum_{i \neq k} \frac{1}{\gamma_k(t) - \gamma_i(t)}- \eta \gamma_k(t) \right) \d t+ \sqrt{2} \d B_k(t) 
 \end{align}
 which we call  ``$\gamma(\eta)$-SDE''.

 Dyson's conjecture says that the $\lambda(\frac{1}{N}, \frac{1}{2})$-SDE converges to global equilibrium in time $\Theta(1)$ (see \cite{yangTopicsRandomMatrix}). Running $\lambda(\frac{1}{N}, \frac{1}{2})$ until time 1 is the same as running the $\gamma(\frac{N}{2})$-SDE until time $T(1) = \frac{1}{N}$.

\section{Stepsize Controller} \label{sec:app:stepsize-controller}

Dyson Brownian motion almost surely never crosses the singularities. Hence, conditioning on non-crossing corresponds to conditioning on an event of probability one, which does not change the dynamics. Given the noise, we can thus calculate the maximal step size, beyond which we would cross the singularity. This is a very useful upper bound, which we employ in practice to get the numerical scheme working. It has two effects: (1) close to the boundary of the Weyl Chamber, it avoids numerically stepping over the singularities and (2) far from the boundary, it allows for larger step size, increasing efficiency.

\subsection{Forward in Time} \label{sec:app:stepsize-controller:forward}

\newcommand{\stepsizeFwdAlgo}{
  \State \textbf{Input:} position $\lambda \in \mathbb{R}^n$, time $t \in \mathbb{R}^+$, independent normal samples $u \sim \mathcal{N}^n$.
  \For{$k \in [n-1]$}
    \State \parbox[t]{\dimexpr\linewidth-\algorithmicindent}{%
        $\delta t_k \gets $ maximal step size based on $\lambda_{k+1}-\lambda_k$ and samples $u_k$, $u_{k+1}$ as described in \Cref{sec:app:stepsize-controller:forward}
        }
  \EndFor
  \State \textbf{Output:} stepsize $\min_i \delta t_i$.
}

\begin{figure}[t]
    \begin{algorithm}[H]
  \caption{Forward stepsize controller for Dyson SDE}
      \label{alg:stepsize_fwd}
      \begin{algorithmic}[1]
        \stepsizeFwdAlgo
      \end{algorithmic}
    \end{algorithm}
  \caption{Forward step size controller which exploits that non-crossing of paths happens with probability $1$. The exact calculations are carried out in \Cref{sec:app:stepsize-controller:forward}.}
\end{figure}

The difference between components $k$ and $k+1$ of $ \lambda$ after a first-order discretization step of size $\d t$ reads for any time $t \in \R^+$
\begin{align}
	\Delta_k(t+\d t) &\coloneqq \lambda_{k}(t+\d t) - \lambda_{k+1}(t+\d t) \\
	&= \lambda_{k}(t) - \lambda_{k+1}(t)  + \d t f_k(t)  +\sqrt{\d t} g_k
	\intertext{with functions $f_k(t) \coloneqq  \alpha \left( \sum_{i \neq k} \frac{1}{\lambda_k(t) - \lambda_i(t)} -  \sum_{i \neq k+1} \frac{1}{\lambda_{k+1}(t) - \lambda_i(t)} \right) - \beta \left( \lambda_k(t) - \lambda_{k+1}(t) \right)$, $g_k \coloneqq \sqrt{2 \alpha} \left( X_k - X_{k+1} \right)$ where the $X_j\stackrel{iid}{\sim}  \mc N(0,1)$ are the independent standard Gaussian random variables taken in the update step of the numerical SDE scheme. Note that $\Delta_k$ is a random variable, but if the step size is sufficiently small we must have almost surely}
	\Delta_k(t + \d t) &>  0.\nonumber%
\end{align}
To find the maximal step size, we observe that the equation above is a quadratic function in the substituted $\tau := \sqrt{\d t}$ yielding the inequality that
\begin{align}
 \tau^2 + \frac{g_k}{f_k(t)}\tau + \frac{\Delta_k(t)}{f_k(t)}
	\qquad \text{is }\qquad \begin{cases}
		> 0 & \text{if $f_k(t)>0$}, \\
		<0 & \text{if $f_k(t)<0$.}
	\end{cases}
\end{align}

We treat first the case $f_k(t) > 0$. The inequality is equivalent to
\begin{align}
	\left(\tau + \frac{g_k}{2 f_k(t)} \right)^2 > \left( \frac{g_k}{2 f_k(t)}\right)^2 - \frac{\Delta_k(t)}{f_k(t)}, \label{eq:app:stepsize:dkfjasdlk}
	\intertext{where we know that $\frac{\Delta_k(t)}{f_k(t)} > 0$ must hold. Hence, if $g_k>0$, any stepsize $\d t>0$ works. 
	Otherwise, if $g_k <0$, the quadratic formula shows immediately that both roots will be in the $\tau>0$ regime. Since the parabola is convex in $\tau$, for inequality (\ref{eq:app:stepsize:dkfjasdlk}) to be fulfilled, we take $\tau$ in the range from $0$ to the smallest root. For $\d t$, that means}
	\d t \in \left( 0,  \frac{1}{4} \left(-\frac{g_k}{f_k(t)} - \sqrt{\frac{g_k^2}{f_k(t)^2} -4 \frac{\Delta_k(t)}{f_k(t)}} \right)^2\right). \label{eq:app:stepsize:some-dt-range}
\end{align}

If $f_k(t) <0$, we have
\begin{align}
	\left(\tau + \frac{g_k}{2 f_k(t)} \right)^2 < \left( \frac{g_k}{2 f_k(t)}\right)^2 - \frac{\Delta_k(t)}{f_k(t)}.
\end{align} 
Since $\frac{\Delta_k(t)}{f_k(t)} < 0$, we know that the inequality is satisfied at $\tau = 0$, and that the largest root $\tau_2$ will be positive:$$\tau_{1,2} = \frac{1}{2} \left(-\frac{g_k}{f_k(t)} \mp \sqrt{\frac{g_k^2}{f_k(t)^2}-4 \frac{\Delta_k(t)}{f_k(t)}}\right).$$
Thus, any stepsize in the following interval is valid
$$\d t \in \left( 0,  \frac{1}{4} \left(-\frac{g_k}{f_k(t)} + \sqrt{\frac{g_k^2}{f_k(t)^2} -4 \frac{\Delta_k(t)}{f_k(t)}} \right)^2\right)$$

Note that $f_k(t) \neq 0$ and $g_k \neq 0$ almost surely.

\subsection{Backward in Time}\label{sec:app:stepsize-controller:backwards}

Backwards in time, we carry out the analogous computation for the more involved backwards SDE in \cref{eq:dyson-timereversal}. In essence, this boils again down to solving a quadratic equation and considering all edge cases.

\section{Shooting mechanism} \label{sec:app:shooting}

During inference, we require access to the learned score. In the forward diffusion, solely numerical errors due to the time discretisation may lead to crossings of %
singularities. On the one hand, when going backwards in time, due to inconsistencies in the learned score, the repulsion in \cref{eq:dyson-timereversal} might be too weak and the sample path may leave the Weyl Chamber for any sensible step size. Since the score is not defined outside the Weyl Chamber, this is problematic. On the other hand, the step size obtained by conditioning on not leaving 
the Weyl Chamber in this ill-trained point would be so small that the numerical solver would get stuck. 
To overcome this, we incorporate a shooting mechanism: If the repulsion force %
is too weak to prevent crossing of the singularity, resulting in a microscopic step size upon conditioning 
to remain in
the Weyl Chamber, we repel with the invariant-state drift, %
i.e., we replace the learned score with the score in the invariant state, as we explain below.
This mechanism ensures that we stay in the Weyl Chamber, while minimizing its impact on the distribution. 
With well-tuned parameters, the shooting %
gets rarely triggered (less than $0.5\%$ of steps) but is essential, as already 
a single event would %
cause getting stuck (upon conditioning) or leaving the Weyl Chamber.

In the forward dynamics, if a certain step size would lead to the probability-$0$ event of leaving the Weyl-Chamber, we know that the source of the error is the finite-time-step discrete approximation, so that decreasing the step size will always provide a fix. In the backward dynamics, however, this is not guaranteed: As discussed, a possible reason for this probability-$0$ event in the backwards dynamics is that the score $s_\theta$ might be not perfectly learned: $s_\theta(\lambda, t) \neq s(\lambda,t)$ for some $\lambda, t$. To avoid this probability-$0$ event, we use the following ``shooting mechanism'': If we were to leave the Weyl-Chamber, we replace the learned score with the analytically known score in the invariant state, eliminating the use of the neural network at that point and leading effectively to a repulsion with the negative forward drift. This mechanism is not expected to change the dynamics in any unfavorable way since it is only applied very rarely, and only in cases where the actual learned dynamics is a much worse approximation (since it would give rise to measure zero  events).

\subsection{Comparison to direct sampling with eigendecompositions} \label{app:comparision:why-direct-simulation-is-unwise}

One may wonder why we do not directly sample from the matrix-valued process  \eqref{eq:SDE-mtx} at time $t$, perform an eigendecomposition to obtain $\lambda(t)$, and then use the increment $\lambda(t+\d t)$ to learn the score via Eq.~\ref{eqn:loss}. While this would render learning simulation-free—relegating the Dyson SDE to the loss derivation and backward dynamics—we avoid this approach due to computational efficiency and precision. This strategy is approximately $100$ times slower for graphs of size $n=10$, as accurate eigendecomposition is computationally expensive; performing it at every step becomes prohibitively costly. Moreover, because we train on increments along the simulated trajectory, a single simulation run allows us to learn from the entire path up to time $t$. Our SDE implementation renders the training runtime on the order of a simulation-free diffusion model.

\section{Complexity dependence on graph size \texorpdfstring{$n$}{n} of numerical step for \eqref{eq:SDE-spectrum} and \eqref{eq:SDE-eigenvectors}} \label{sec:app:complexity-update-step}

When  simulating the \eqref{eq:SDE-spectrum} numerically, for a fixed eigenvalue $\lambda_k$, we need to calculate its distance to any $\lambda_\ell$ for $\ell \in [n] \setminus \{k\}$ in order to obtain the repulsion force entering the drift of the \eqref{eq:SDE-spectrum}. Since we need to do this for each component, we have for a fixed time step a complexity of $\Oh{n^2}$. 

When simulating \eqref{eq:SDE-eigenvectors} numerically, for one time step, we update the entire set of $n$ eigenvectors together. Each of the $n (n-1)/2$ entries of the Lie algebra can be computed in time $\Th{1}$ (note that $E_{(l,k)}$ is a basis element of the Lie algebra). This update requires thus $\Oh{n^2}$. Accounting for the projection from the Lie algebra in the Lie group, this yields a complexity of $\Oh{n^3}$ for an eigenvector update.

Since \eqref{eq:SDE-spectrum} decouples from \eqref{eq:SDE-eigenvectors}, we thus have a complexity for an update step of $\Oh{n^2}$ if one is interested in only the spectrum, and a complexity of $\Oh{n^3}$ if one is interested in the spectrum and eigenvectors. Note that this modularity allows the practitioner to first learn the spectrum in isolation and do any hyperparameter tuning steps on the (smaller) spectral model, and then use this already tuned model to train and optimize for the remaining eigenvector dynamics.

In comparison, an OU-based approach on the entire graph with a GNN-based algorithm requires for a fixed time point an update of the SDE in $n^2$ entries in time $\Oh{n^2}$ followed by $C$  message passing steps (a GNN is trained using message passing, where $C$ is a constant defining how many hops a message is being passed), with each message passing taking time $\Oh{n^3}$, leading to an overall complexity of $\Oh{n^3}$ per time step, assuming that $C$ is constant.

\section{Empirical Considerations} \label{sec:app:empirical-considerations}

\begin{figure}[ht!]
    \centering
    \includegraphics[width=0.5\linewidth]{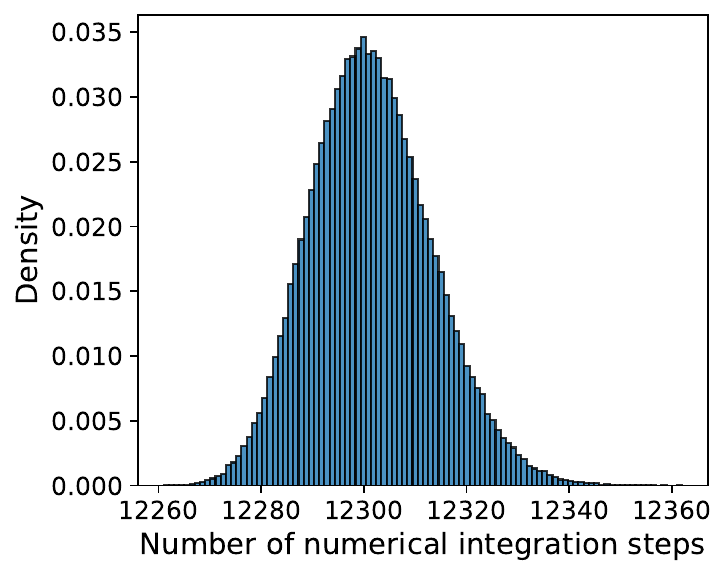}
    \caption{Concentration of time steps of the numerical integrator: Within $5$ second, we generated $105,000$ forward paths using our numerical scheme, where the total number of steps per path are distributed as depicted in this histogram. As can be seen, the distribution concentrates well.}
    \label{fig:time-steps-concentrate-well}
\end{figure}

While we give a thorough analysis of the computational complexity of each update step, the number of time steps is determined by the adaptive step size algorithm. For practical reasons, our method can be configured with a minimal (and maximal) step size. We illustrate the distribution of the number of time steps on the Brain dataset in \Cref{fig:time-steps-concentrate-well}, where it can be observed that the number of time steps taken concentrates very well. Note that this number of time steps perfectly suffices even for a dataset as large as Brain, which contains $15,000$ graphs. Note further that while these are the number of steps the numerical integrator takes forward, learning has to happen only on much fewer steps, through the described pre-defined schedule. In the case of Brain, learning happens only on $651$ time points. The numerical generation of $105,000$ paths with step number depicted in \Cref{fig:time-steps-concentrate-well} took in total $5$ seconds. For memory considerations, note that we require for training only a single A100 GPU ($80$ GB Video Random Access Memory).

We further note that the skip event (see \Cref{alg:dyson-training-fwd}) is indeed a rare event: In the Brain setting, the empirical probability of a step being a skip is $10^{-5}$. Similarly, shootings in the analogous backwards runs happen rarely. Note that skips and shoots are needed for correctness and stability, but no negative impact on runtime due to their rarity.

\section{Inference} \label{sec:app:sampling}

\newcommand{\dysonAlgoBackwards}{
\State{\textbf{Input:} dimension $n$, schedule $\mc T = \{t_j\}$, trained score network $s_\theta$}
\State{$\lambda(T) \gets $ Spectrum of a random GOE matrix.} \Comment{Sampling from invariant distribution.}
  	\State{$t\gets T$}
  	\While{$t>0$} 
  	\State{Let $ u \sim \mc N(0,I_n)$} 
  	\State{$\delta t \gets\text{BackwardStepsizeController}( \lambda(t), u)$} \Comment{{Step size as described in ~\ref{sec:app:stepsize-controller:backwards}.}}
    \State{$\hat s_\theta \gets$ Interpolation of $s_\theta(\lambda,t_j)$ and $s_\theta(\lambda,t_{j+1})$ for $t_j, t_{j+1} \in \mc T$ closest points in schedule}
    \State{${\rm BackwardsDrift} \gets$ drift of \cref{eq:dyson-timereversal} at time $t$ and point $\lambda(t)$ with score $\hat s_\theta(\lambda,t_{j+1})$}
  	\If{$\lambda_t - {\rm BackwardsDrift} \cdot \delta t -u \sqrt{2 \alpha \delta t}  \in {\rm Weyl\ Chamber}$}
    \State{$\lambda_{t-\delta t} \gets \lambda_t - {\rm BackwardsDrift} \cdot \delta t - u \sqrt{2 \alpha \delta t}$}
    \Else \Comment{Otherwise -- in a rare event -- shooting mechanism triggers (see \Cref{sec:app:shooting})}
    \State{$\lambda_{t-\delta t} \gets \lambda_t - (-{\rm ForwardDrift}) \cdot \delta t - u \sqrt{2 \alpha \delta t}$} \Comment{${\rm ForwardDrift}$ as in \ref{sec:dyson-BM}} 
  	\EndIf
  	\State{$t \gets t - \delta t$}
  	\EndWhile

 \State{\textbf{Output:} spectral sample $\lambda(0) \in \R^n$}
}

\begin{figure}[t]
    \begin{algorithm}[H]
  \caption{DyDM sampling}
      \label{alg:dyson-algo-backwards}
      \begin{algorithmic}[1]
        \dysonAlgoBackwards
      \end{algorithmic}
    \end{algorithm}
  \caption{
  Sampling from the Dyson Diffusion Model. The \cref{eq:dyson-timereversal} is evolved backward in time with the shooting mechanism and an adaptive step size, ensuring that the paths remain in the Weyl Chamber. 
  }
\end{figure}

We describe in \Cref{alg:dyson-algo-backwards} the sampling procedure with the shooting mechanism (see \Cref{sec:app:shooting} for details) incorporated, ensuring that we remain in the Weyl Chamber.

\section{Engineering}\label{sec:app:engineering}
We follow \cite{karrasAnalyzingImprovingTraining2024} by using SiLU activations. We further use EMA to average over multiple runs \citep{songImprovedTechniquesTraining2020}.

A key strength of DyDM is that the score $s_\theta(\lambda,t)$ can be parameterized with any learning architecture, without being constrained to GNNs or graph transformers. We demonstrate this by parameterizing the score with a simple MLP. The size of layers varies by application, and we document it for each dataset in the uploaded Supplementary Material. For instance, for the large dataset of $15,000$ brain ego graphs, our model consists of a hidden MLP of depth $4$, where the input and output layer have width $64$ and the hidden layers have width $256$. The space + time data is first scaled up with a linear layer from size $n+1=11$ followed by a batch norm to feed into the hidden MLP, and upon processing through the hidden MLP, it  gets scaled down through a simple linear layer to size $n=10$. In the hidden MLP, we employ as nonlinearities scaled SiLU functions, as argued by \cite{karrasAnalyzingImprovingTraining2024}. We use AdamW for optimization \cite{LoshchilovH19}.

To make full use of the GPU memory, in one epoch, we sample $N_{paths}$ paths in parallel. If the dataset is too small to fill the GPU memory, we sample multiple, independent, paths of the same data points in parallel. From these samples, we update the score network $s_\theta$ in smaller batches. Proceeding in this way ensures that JAX uses the full potential of the GPU. These parameters ($N_{paths}$ and batch size) can be specified in the configuration file.

Our runtimes are as follows: For WL-Bimodal, sampling $1000$ paths takes $16$ seconds and training takes $19$ hours.
For Community Small, sampling $1000$ paths takes $60$ seconds and training takes $10$ hours.
For Brain, sampling of $1000$ paths is very fast at $11$ seconds. Training takes $16$ hours. Runtime-wise, our strength therefore lies in fast generation. 

We sample $1000$ paths to make the reported distances statistically robust, and report in addition to the mean also the Wasserstein of the marginals of these $1000$ samples to the ground truth, accounting both for high-dimensional and marginal effects (see \Cref{sec:results}). 

We implement the model in JAX \cite{jax2018github} (version 0.9.0, licensed under  Apache license, version 2.0) and Equinox \url{https://github.com/patrick-kidger/equinox} \cite{kidger2021equinox} (version 0.13.6, licensed under Apache license, version 2.0).

\subsection{Choice of Time Grid} \label{sec:app:time-grid}
As outlined in \Cref{sec:dyson-sde-to-diffusion-model}, we choose an exponential time grid on which the objective is learned. This is due to the mixing behavior of Dyson's Brownian motion. For instance, for the Brain dataset, we use an exponential time grid as detailed in \Cref{tab:app:exp-timegrid}.
\begin{table}[H]
\caption{Example of exponential time grid, here for the brain dataset which contains in total $15,000$ graphs. $\d t$ is $0.05$, and the final time is $T=12.0$.} \label{tab:app:exp-timegrid}
\centering
\begin{tabular}{ccc}
\textbf{from} & \textbf{to} & \textbf{stepsize} \\
\hline
$0$ & $1/8$ & $1/64 \cdot \d t$ \\
$1/8$ & $1/4$ & $1/32 \cdot \d t$ \\
$1/4$ & $1/2$ & $1/16 \cdot \d t$ \\
$1/2$ & $1$ & $1/8 \cdot \d t$ \\
$1$ & $2$ & $1/4 \cdot \d t$ \\
$2$ & $3$ & $1/2 \cdot \d t$ \\
$3$ & $7$ & $1 \cdot \d t$ \\
$7$ & T & $2 \cdot \d t$ \\
\end{tabular}
\end{table}

\subsection{Preprocessing and dealing with different dimensions}
As is the case for an Ornstein-Uhlenbeck diffusion, the speed of convergence depends for \eqref{eq:SDE-spectrum} on $(1)$ the coefficients and $(2)$ the initial condition. We can thus choose to significantly vary the final time $T$ or rescale the initial condition. We choose the latter (although both options are feasible). To that end, we rescale the spectra with an affine transformation in a preprocessing step. In more detail: For a given set of graphs, we perform an eigendecomposition, and rescale so that among the entire dataset the largest eigenvalue is at most $\lambda_{\rm max}$ and at least $\lambda_{\rm min}$. For instance, on the benchmark models we chose $\lambda_{\rm max} =5$,  $\lambda_{\rm min} =-5$. If a spectrum has eigenvalues of multiplicities greater than $1$, we perform an $\epsilon$-perturbation, where the $\epsilon$ depends on the distance of the closest eigenvalues in the dataset. In postprocessing, the spectra are scaled back and the perturbation is undone for eigenvalues that are at most $\epsilon$ apart after generation.
Note that with this preprocessing, only one eigendecomposition per training graph is necessary.

\section{Challenges of Dyson's Brownian Motion} \label{sec:app:challenges-which-we-all-solved}
Using Dyson's Brownian motion for a diffusion presents several challenges, all of which we overcame in this paper. First, the Dyson SDE is not an OU process, but instead an SDE with singularities of order $\mathcal{O}(1/(\lambda_k - \lambda_l))$ in the drift, posing both theoretical and numerical challenges.
Second, the conditional density $p(x \mid x_0)$ is non-Gaussian and challenging to obtain, as with any non-OU diffusion process. Therefore, we do not have access to a canonical loss function. Finally, the non-availability of conditional distributions means that training is not simulation-free.

We overcome the obstacles mentioned above and provide a diffusion model for the spectra of graphs based on the Dyson SDE (Fig.~\ref{fig:mainfig}). The model is not only efficient but is also able to distinguish between spectra of graphs that GNNs are blind to (Fig.~\ref{fig:wl-demonstration}). In addition, with DyDM, no ad~hoc data augmentation is necessary. 
Further, through \eqref{eq:SDE-eigenvectors}, we give the dynamics of the remaining information in form of conditioned eigenvector dynamics, hence making them accessible for future work devoted to eigenvector diffusion.

\subsection{Why not log-transform the SDE?}
One idea could be to transform the spectral SDE into terms $\lambda_1, \lambda_2-\lambda_3 ,\ldots , \lambda_{n-1}-\lambda_n$ and take logarithms, to avoid singularities. However, this is not desirable for multiple reasons: First, upon applying Ito to this SDE, we (i) lose the $\ln$  and obtain singularities again and (ii) get higher order singularities  
 $\d \ln(x_t) = \frac{1}{x_t} \d x_t -\frac1{2x_t^2}(\d x_t)^2$.
 Second, the space that would need to be sampled would certainly not decrease, since now the transformed domain reaches from $-\infty$ to $+\infty$.
Hence, we choose the method described in the main part.

\section{Why not learn on all  \texorpdfstring{$n!$}{n!}  many graph representations?} \label{sec:argument-against-learning-all-permutations}
In short, learning $n!$ more data is much harder. This point has been mentioned by previous literature, e.g.\ \cite{niuPermutationInvariantGraph2020}. However, if we go deeper, the interested reader might wonder why exactly.

\subsection{Rigorous argument}

We give here a toy example, where the challenge can be phrased rigorously. Suppose we have a binary matrix $X \in \{0,1\}^{m,k }$ with independent entries, which are  for $i \in [m], j \in [k ] $   distributed as $X_{ij} \sim \Bernoulli{p_j}$ for some unknown $p_j \in [0,1]$. The task is to estimate the $p_j$. The motivation for this example stems from the following setting: We want to learn the probability of $k$ objects (for instance, graphs), each having $m$ representations (for instance, representations of graphs such as adjacency matrices). Each column of the matrix $X$ thus consists  of all representations of the same object. We define two estimators, with ${\rm est_A}$ using the inductive bias and ${\rm est_B}$ not using it. 

To that end, suppose we have $N$ uniformly at random obtained samples $Z_1, \ldots, Z_N$. In more detail, that means that we sample for each $\ell \in [N]$ a pair of indices $(i_\ell, j_\ell) \in [m] \times [k]$ uniformly at random, and describe the obtained sample by $Z_\ell \sim X_{i_\ell, j_\ell} $. Estimator ${\rm est_A}$ makes use of the inductive bias. That is, for $u \in [m], v\in [k]$ we define
\begin{align}
    {\rm est_A^{(u,v)}} \coloneq \frac{k}{N} \sum_{\ell \in [N]} Z_\ell \1_{j_\ell = v}. \nonumber
\end{align}
Estimator ${\rm est_B}$ does not make use of the inductive bias. That is, for $u\in [m], v \in [k]$, we define 
\begin{align}
     {\rm est_B^{(u,v)}} \coloneq \frac{k \cdot m}{N} \sum_{\ell \in [N]} Z_\ell \1_{i_\ell = u, j_\ell = v}. \nonumber
\end{align}
Clearly, both estimators are unbiased: $\E{{\rm est_A^{(u,v)}}} = \E{ {\rm est_B^{(u,v)}}} = p_v$. However, their mean squared error, defined for $X\in \{A,B\}$ as
$$ {\rm MSE}({\rm est_X}) \coloneqq \frac{1}{m \cdot k} \sum_{u \in [m], v \in [k]} \E{\left({\rm est_X^{(u,v)}} - p_v \right)^2},$$
 varies significantly between ${\rm est_A}$ and ${\rm est_B}$.
 
\begin{theorem}[Mean square error] \label{cor:avg-mse:orders}
    In dependence of the problem size $m$ and number of samples $N$, the mean square error of ${\rm est_A}$ is of order $\Theta(1/N)$, while the mean square error of ${\rm est_B}$ is of order $\Theta(m/N)$.
\end{theorem}
To prove \Cref{cor:avg-mse:orders}, we derive the mean squared error for both estimators.

\begin{lemma}[MSE of ${\rm est_A}$]\label{lem:avg-mse:estA}
    The mean squared error of estimator ${\rm est_A}$ is
    $$ {\rm MSE}\left({\rm est_A} \right) =  \frac{1}{N} \sum_{v \in [k]} \left( p_v - \frac{1}{k} p_v^2 \right). $$
\end{lemma}
\begin{proof}
   We have for $u \in [m], v \in [k]$
   \begin{align*}
    &\E{\left({\rm est_A^{(u,v)}}-p_v \right)^2 } \\
    =& \E{ \left( \frac{k}{N} \sum_{\ell \in [N]} Z_\ell \1_{j_\ell = v} -p_v \right)^2} \\
    =& \frac{k^2}{N^2} \sum_{\ell \in [N] } \E{ \sum_{o \in [N]}Z_\ell \1_{j_\ell = v} X_o \1_{j_o = v}} - 2  \frac{k}{N} \E{ \sum_{\ell \in [N]} Z_\ell \1_{j_\ell = v}} p_v + p_v^2  \\
    =& \left(1 - \frac{1}{N}\right) p_v^2 + \frac{k}{N} p_v - 2 p_v^2 + p_v ^2 \\
    =&  \frac{k}{N} p_v - \frac{1}{N} p_v^2.
   \end{align*}

   Averaging over all $u \in [m], v \in [k]$, we obtain the desired result.
\end{proof}

\begin{lemma}[MSE of ${\rm est_B}$]\label{lem:avg-mse:estB}
    The mean squared error of estimators ${\rm est_B^{(u,v)}}$ for  $u \in [m]$, $v \in [k]$ is
    $$ {\rm MSE}\left({\rm est_B} \right) =  \frac{1}{N} \sum_{v \in [k]} \left( m \cdot p_v - \frac{1}{k} p_v^2 \right). $$
\end{lemma}
\begin{proof}
    We have for $u\in [m], v \in [k]$
    \begin{align*}
      &\E{\left({\rm est_B^{(u,v)}}-p_v \right)^2 } \\
    =& \E{ \left( \frac{k \cdot m}{N} \sum_{\ell \in [N]} Z_\ell \1_{i_\ell = u, j_\ell = v} -p_v \right)^2} \\
    =&  \frac{k^2 m^2}{N^2} \sum_{\ell \in [N] } \E{ \sum_{o \in [N]}Z_\ell \1_{i_\ell = u, j_\ell = v} Z_o \1_{i_o = u, j_o = v}} - 2  \frac{k\: m}{N} \E{ \sum_{\ell \in [N]} Z_\ell \1_{i_\ell = u, j_\ell = v}} p_v + p_v^2 \\
    =& \left(1 - \frac{1}{N} \right) p_v^2 + \frac{k \: m}{N} p_v -2 p_v^2 + p_v^2 \\
    =& \frac{k \: m}{N} p_v - \frac{1}{N}p_v^2
    \end{align*}
    Summing over all estimators for $u \in[m], v \in [k]$ gives the desired result.
\end{proof}
By considering the estimation problem as a problem of  parameters $m$, $N$, \Cref{cor:avg-mse:orders} follows directly from \Cref{lem:avg-mse:estA}, \Cref{lem:avg-mse:estB}.

The same argument may be carried out with Normal instead of Bernoulli random variables.

\section{WL-equivalence of regular graphs}\label{sec:app:wl-k-reg-proof}
We now prove \Cref{lem:wl-equivalence-regular}.
\begin{replemma}{lem:wl-equivalence-regular}
    \stateLemmaKReg
\end{replemma}
\begin{proof}
	Recall that the 1-Weisfeiler-Leman (WL) algorithm tests graph equivalence by iteratively updating vertex \textit{colors}. Initially, all vertices share one color. In each step, a vertex's new color is determined by the multiset (a set allowing for duplicates) of its own color and its neighbors' current colors. This continues until the coloring stabilizes. Two graphs are WL-equivalent if this process generates identical color counts (histograms) at every step. 

    For a $k$-regular graph, since every vertex has exactly $k$ neighbors, in each iteration of the WL-algorithm all vertices have the same color. Thus in particular the histograms are always the same and therefore any two $k$-regular graphs are WL-equivalent. 
    
    A $k$-regular graph is not WL-equivalent to a $\ell$-regular graph for $k\neq \ell$ since the colors after the first iteration are distinct as the number of neighbors is different. Moreover, if a graph is not regular, then the first iteration must assign a different new color to at least two vertices in the first iteration. Therefore, the color histogram is not the same as the color histogram of a regular graph and thus they are not WL-equivalent. This shows that the WL-equivalence class of a $k$-regular graph is exactly the set of $k$-regular graphs. 
\end{proof}

In particular, it follows from Morris \citep{morrisWeisfeilerLemanGo2019} that GNNs cannot distinguish $k$-regular graphs.

\section{Laplacian Spectrum}\label{sec:app:laplacian}
The Laplacian spectrum captures key properties of a graph, such as the Fiedler Value encoding the algebraic connectivity of the graph. Our model works perfectly well on the Laplacian spectrum, for which we plot the marginal densities on the datasets ``Brain'' and ``Community Small''. Note that the model perfectly learns key properties such as the fact that the smallest eigenvalue is $0$ and learns the distributions well.

\begin{figure}[ht!]
    \centering
    \includegraphics[width=\linewidth]{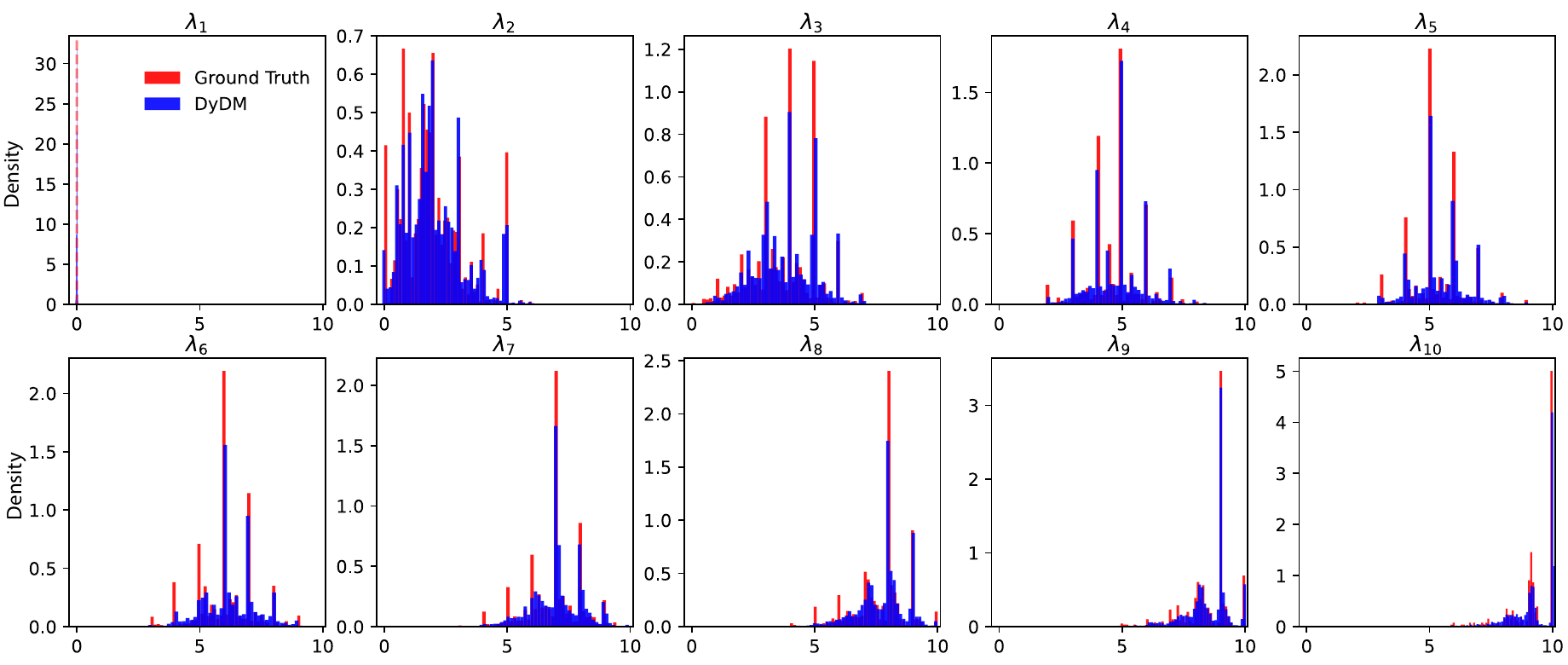}
    \caption{Performance of DyDM on the Laplacian of the Brain dataset compared to the underlying distribution: DyDM perfectly learns the (rough) distribution and key properties such as the smallest eigenvalue always being $0$ (faint line, since it is very concentrated) or the distribution of the Fiedler Value ($\lambda_2$).}
    \label{fig:laplacian-brain-DyDM-very-good}
\end{figure}

\begin{figure}[ht!]
    \centering
    \includegraphics[width=\linewidth]{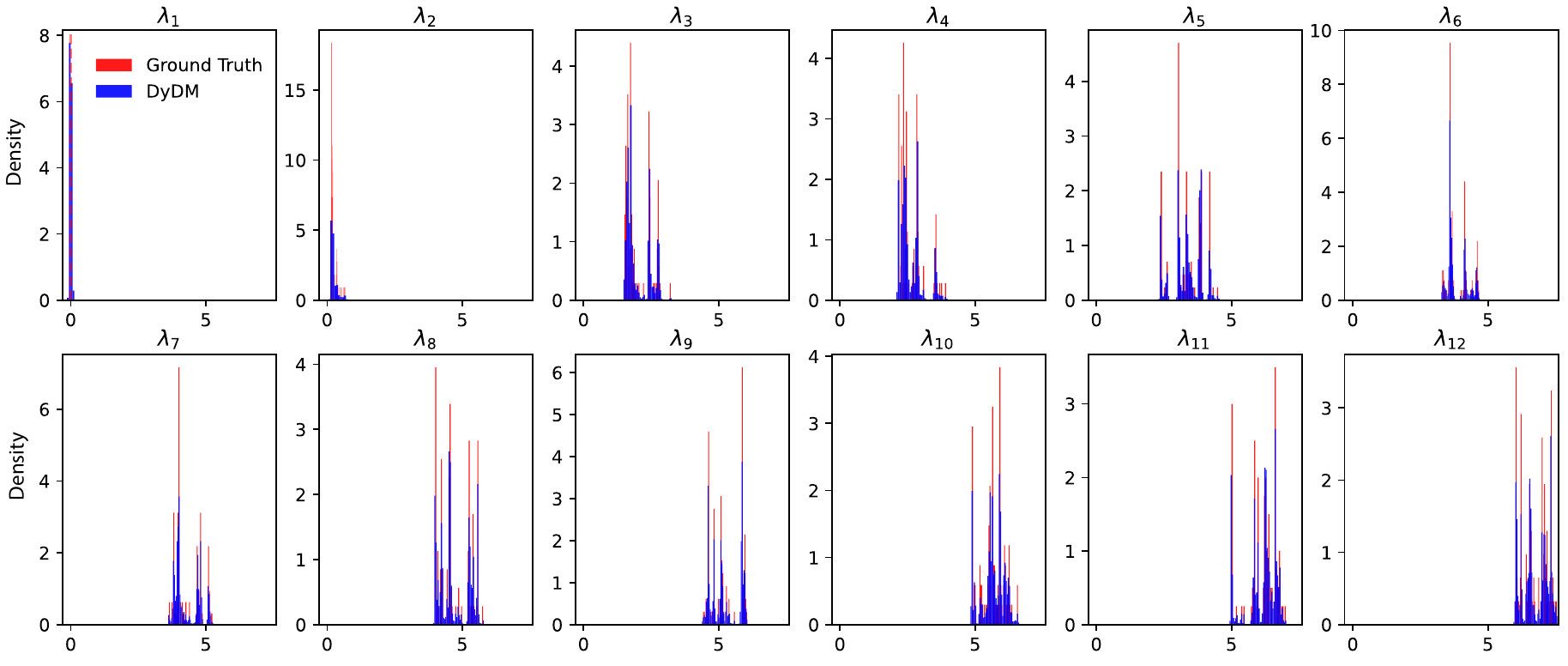}
    \caption{Performance of DyDM on the Laplacian of the Community Small  dataset: DyDM learns the distribution and key properties very well.}
    \label{fig:laplacian-community-DyDM-very-good}
\end{figure}

For Brain (\Cref{fig:laplacian-brain-DyDM-very-good}), we plot for each eigenvalue $\lambda_i$ a histogram of the DyDM-generated marginal (blue) compared to a histogram of the Ground Truth (note that the Brain dataset test for generalization).  For the commonly used ``Community Small'' dataset, we show in \Cref{fig:laplacian-community-DyDM-very-good} the first $12$ eigenvalues, since only $73$ graphs have more than $12$ nodes.

Note that in this Section, we use the convention $\lambda_1 \leq \cdots \leq \lambda_n$ to stay consistent with Graph Laplacian literature.


\section{Datasets}\label{sec:app:datasets}
\paragraph{WL-Bimodal} The WL-Bimodal dataset consists of $80\%$ graph $A$ and $20\%$ graph B (see Fig.~\ref{fig:wl-demonstration}) adjacency matrices. We drew among all permutations $5,000$ permutations uniformly at random and shuffled the graphs. Hence, we have $5,000$ matrices, of which $4,000$ represent graph $A$ and $1,000$ represent graph $B$. Adjacency matrices representing the same graph are not identical matrices, but rather permutations of each other, making  this dataset relevant to \emph{graph} generative models.
The first $80\%$ of this dataset are used for training, the remaining $20\%$ are used for testing. 

\paragraph{Community-small} This standard benchmark dataset \citep{niuPermutationInvariantGraph2020,joScorebasedGenerativeModeling2022,youGraphRNNGeneratingRealistic2018} consists of $100$ graphs of size up to $20$  vertices. We comment in \Cref{sec:app:comparing-to-benchmark-models:undersampling} on the small dataset ($100$ graphs) compared to the big dimension (up to 20 vertices) and the thereby induced effect of undersampling.

\paragraph{Brain}
We report this dataset in our repository. In detail, we construct from the brain graph \cite{bigbrain,nr-aaai15} so-called ego-graphs. That is, we take the (distance 1) neighborhoods of vertices, and consider the induced subgraph. From those, we generate $15,000$ graphs of size $n=5$ to $n=10$ vertices with eigenvalue multiplicity up to $3$, with the closest eigenvalues -- which are not multiplicities -- having distance $0.036$. We take $70\%$ as train graphs, $15\%$ as validation, and the remaining $15\%$ as test graphs.

\section{Comparing to Benchmark Models} \label{sec:app:comparing-to-benchmark-models}

\subsection{On undersampling}
 \label{sec:app:comparing-to-benchmark-models:undersampling}
For the ``bimodal'' case, we have sufficient statistics for the $10$-dimensional space $C_{10}$ ($N=5,000$ graph samples, each isomorphic to one of two graphs) and know in addition the underlying distribution; hence, an extensive interpretation of this result is appropriate. For a fair comparison, we thus follow the standard test/train split procedure as reported in \citep{joScorebasedGenerativeModeling2022, youGraphRNNGeneratingRealistic2018, niuPermutationInvariantGraph2020} using $80\%$ of the data as train data and the remaining $20\%$ as test data.

Conversely, the standard benchmark set ``community small'' \citep{niuPermutationInvariantGraph2020,joScorebasedGenerativeModeling2022,youGraphRNNGeneratingRealistic2018} contains \emph{only} 100 graphs, and each has a size of up to $n=20$ vertices. Thus, a comparison from the learned distribution based on (few) training samples to (very few) test samples suffers from undersampling. This becomes very stark if one considers the following issue: If one would extract $80\%$ of the dataset for training, \emph{where} they are taken from already matters a lot: Whether they are taken from the front or back changes the maximal graph size in the training set. Depending on whether the training data is taken from the front or back of the dataset, a model trained on the training set might thus have no possibility to learn the correct maximal graph size.
More generally, poorness of benchmarks in graph generative learning has been recently addressed by \cite{bechler-speicherPositionGraphLearning2025}. To offer some consistent comparison, we do include the standard benchmark ``community small'', but focus on \emph{memorization} rather than the (on those benchmarks untestable) generalization. 

To overcome the issue of undersampling, we construct a set of $15,000$ ego-graphs from the brain dataset \cite{bigbrain} as described in \Cref{sec:app:datasets}, which is sufficiently large to \emph{not} suffer from undersampling. We train both our model and DiGress on $70\%$ (= $10,500$ graphs), perform hyperparameter tuning (see below for details on the DiGress hyperparameter tuning) on a validation set of $15\%$, and test on the remaining $15\%$.

\subsection{GDSS}\label{sec:app:gdss}
To compare to the GDSS model \cite{joScorebasedGenerativeModeling2022}, we take the following approach to ensure a fair comparison: The GDSS model has been trained and optimised on the ego small and community small dataset, so that we take for these datasets the snapshots and hyperparameters given by the original paper \cite{joScorebasedGenerativeModeling2022}. For the remaining datasets, we start with the settings from the community-small dataset since that has similar size, adapt the maximal number of vertices to the dataset, and then perform hyperparameter tuning as described in Appendix C of the GDSS paper: We form a grid search on the model's following parameters: The scale coefficient in $\{0.1, 0.2, 0.3, 0.4, 0.5, 0.6, 0.7, 0.8, 0.9, 1.0 \}$, the signal-to-noise ratio in $\{0.05,0.1,0.15,0.2\}$, and in addition to the GDSS paper, we also try different $\beta_{max}$ in $\{1,10,20\}$, and try different batch sizes $\{128,4096\}$. We test with and without EMA. The motivation for the additional parameters we hyper tune is that we observed that they further improve the GDSS model. This gives full fairness to the model. The training time on one A100 GPU per job was for the brain dataset about 24 hours for each of the configurations with batch size $128$ and approximately 1 hour and 45 minutes for each of the configurations with batch size $4096$. 

For the community small dataset, training took $15$ minutes, sampling $450$ seconds.

For the WL-Bimodal case, again using an $A100$ GPU, one training run of GDSS took $5$ hours, and one sampling run of $1{,}000$ graphs took $450$ seconds.

On the Two-WL graph case, our hyperparameter tuning resulted in $280$ combinations that we trained and sampled from. From each of those 280 models, we generated $1,000$ graphs. Most of those models worked fine, that is, $269$ models did not contain NaNs in their output. From those, we select the best model based on the following relative error: From the generated samples, we calculate the share of spectra $\epsilon$-close to the spectrum of graph $A$, say $\hat p_A$ and the share of spectra $\epsilon$-close to graph $\hat p_B$ (we choose the $l_2$ distance with $\epsilon = 0.2$). Recall that in the ground truth, we have $p_A = 0.8$, $p_B = 0.2$. We then selected the best model based on the  relative error
$$ \frac{\abs{\hat p_A - p_A}}{p_A} +  \frac{\abs{\hat p_B - p_B}}{p_B}.$$
We then trained the best hyperparameter choice with $5$ different seed values and report their mean and standard deviation.

In summary, we invested a lot of resources in following both the hyperparameter tuning given in the GDSS paper and, in addition, tried new hyperparameters, leading to the $280$ models that we trained and sampled from. This ensures maximal fairness.

\subsection{EDP-GNN}
We proceeded analogously to GDSS: We used the given configurations for community small. We observe that the model already performs hyperparameter tuning of the noise scales during sampling. For the other datasets, we have used the given configurations for community small and in addition tried the learning rates $\{0.001, 0.0002 \}$, number of diffusion steps $\{1,000,2,000 \}$ and number of layers $\{4,6 \}$. In the two graph case, for example, the optimal configuration was with learning rate $0.0002$, number of diffusion steps being equal to $2,000$ and $6$ layers. The training time for $5,000$ epochs on a H100 GPU was approximately 10 hours. 

\subsection{DiGress}
As with the previous models, we used the given configurations for community small. For the other datasets, we have used the given configurations for community small and in addition tried the learning rates $\{0.001, 0.0002 \}$, weight decay parameters $\{ 10^{-2}, 10^{-12} \}$, number of diffusion steps $\{500,1,000 \}$ and number of layers $\{5,8 \}$. The model was quite robust to hyperparameter tuning and in the two graph example, for all parameters the model sampled the $80\%$-graph with likelihood between $75\%$ and $82\%$. The training time for $1,000$ epochs on a H100 GPU was approximately 3 hours. 

\subsection{ConGress}
Since ConGress and DiGress come from the same paper, we have proceeded almost exactly as in DiGress. We tried the learning rates $\{0.001, 0.0002 \}$, weight decay parameters $\{ 10^{-2}, 10^{-12} \}$, number of diffusion steps $\{500,1,000 \}$ and number of layers $\{6,8 \}$. The model was again rather robust to hyperparameter tuning, yet not as much as DiGress. In the two graph example, for all parameters the model sampled the $80\%$-graph with likelihood between $10\%$ and $25\%$. The default learning rate from the community-small configuration was $0.0002$, yet we have observed that the results were significantly better with learning rate $0.001$. The training time for $1,000$ epochs on a H100 GPU was approximately 3 hours. 

\subsection{Comparison table with more digits}\label{sec:app:more-digits}
We provide here the table shown in the main part \Cref{tab:distances} but with more digits (\emph{not} implying that all are statistically significant): This rationalizes which entries in \Cref{tab:distances} are dark green and which ones are light green. Note that this result is included only for transparency, since the reported digits here go beyond the significant digits.
\begin{table}[H]
\caption{Statistical distances of DyDM compared to standard models, as in \Cref{tab:distances} but with more digits.}
\centering
\begin{tabular}{|l|cc|cc|cc|}
\hline
\textbf{Dataset} & \multicolumn{2}{c|}{\makecell{WL-Bimodal}} & \multicolumn{2}{c|}{Community Small}  & \multicolumn{2}{c|}{Brain} \\
\textbf{Distance}& $\mu$ & $\mathcal{W}_{\mathrm{marg}}$ &  $\mu$ &$\mathcal{W}_{\mathrm{marg}}$ 
&  $\mu$ & $\mathcal{W}_{\mathrm{marg}}$  \\
\hline
DyDM (ours)&
\best{0.0166}&\almostbest{0.0076}& %
\best{0.0671} & \best{0.0172} & %
\best{0.0455}& \best{0.0275}\\ %
EDP-GNN & 
0.1342&0.0750 & %
0.4164& 0.1356& %
0.0723 & 0.0321 %
\\
GDSS & 
0.2416&0.1339 & %
0.3812 & 0.1444 & %
0.3276 &  0.1191 %
\\
ConGress & 
0.3802&0.1590& %
0.2741&0.1138& %
0.1315 & 0.0296  %
\\
DiGress (no trick) &
1.0568&0.2852 & %
2.5088&0.4481 & %
0.5686 & 0.1749  %
\\
DiGress (trick) &
0.0302&\best{0.0073}& %
0.0934 & 0.0254& %
0.1208&\almostbest{0.0285}   %
\\
\hline
\end{tabular}
 \label{tab:distances-more-digits}
\end{table}

\begin{table}[H]
\caption{Empirical standard deviation of runs.}
\centering
\begin{tabular}{|l|cc|cc|cc|}
\hline
\textbf{Dataset} & \multicolumn{2}{c|}{\makecell{WL-Bimodal}} & \multicolumn{2}{c|}{Community Small}  & \multicolumn{2}{c|}{Brain} \\
\textbf{Distance}& $\mu$ & $\mathcal{W}_{\mathrm{marg}}$ &  $\mu$ &$\mathcal{W}_{\mathrm{marg}}$ 
&  $\mu$ & $\mathcal{W}_{\mathrm{marg}}$  \\
\hline
DyDM (ours)&
0.0057& 0.0011& %
0.0341 & 0.0044 & %
-&-\\ %
EDP-GNN & 
0.0251&0.0226& %
- & - & %
0.0838 & 0.0265 %
\\
GDSS & 
0.0125&0.0067& %
0.0617 & 0.0055 & %
0.0951 &  0.0099 %
\\
ConGress & 
0.3143& 0.0736 & 
0.0171&0.0033& %
0.0943 & 0.0131 %
\\
DiGress (no trick) &
0.0349 &0.0060 & %
0.0169&0.0057& %
0.4252 & 0.1293 %
\\
DiGress (trick) &
0.0150&0.0033& %
0.0129 & 0.0025& %
0.0458 & 0.0065 %
\\
\hline
\end{tabular}
 \label{tab:distances-emprical-std}
\end{table}

\section{Learning dynamics of EDP-GNN} \label{sec:app:learning-dynamics}

\begin{figure}[t!!]
    \centering
    \includegraphics[width=\linewidth]{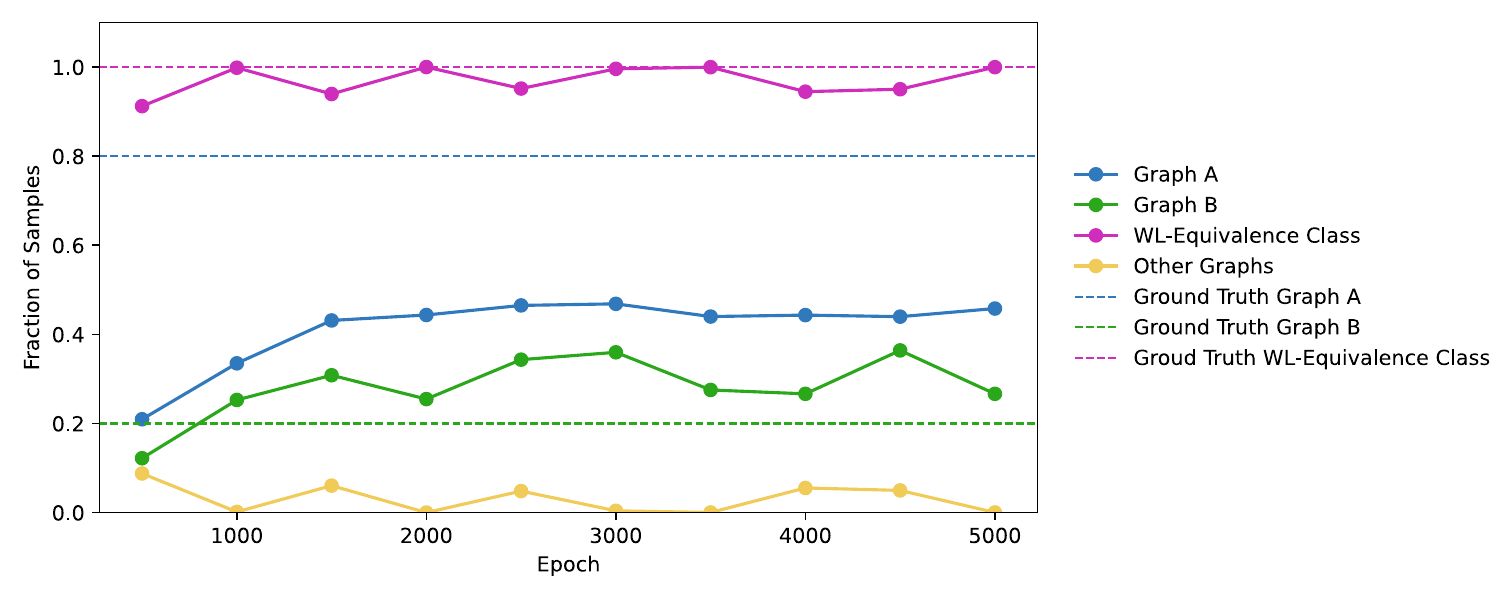}
    \caption{Learning dynamics of EDP-GNN on the WL-Bimodal dataset (ground truth = $80\%$ graph A, $20\%$ graph B): The model learns very quickly (after less than 500 epochs) the WL-equivalence class, but struggles to learn graphs $A$ and $B$.}
    \label{fig:wl-edp-gnn-learning-dynamics}
\end{figure}

We report in \Cref{fig:wl-edp-gnn-learning-dynamics} the learning progress of EDP-GNN on the WL-Bimodal dataset. We average over $4$ different training and sampling runs of EDP-GNN. After $5,000$ epochs, we observe the result of EDP-GNN reported in \Cref{fig:wl-demonstration}.

We observe that the model quickly learns the WL equivalence class (the pink line is from epoch $500$ onward close to $1$). The share of graph $A$ and graph $B$ samples initially increase until epoch $1,500$, but then remain low and significantly different from the ground truth (blue and green dashed lines). Importantly, a significant share of the samples are WL-equivalent but neither isomorphic to graph $A$ nor to graph $B$.

\section{Towards a Diffusion Model for the Eigenvector SDE}\label{sec:app:DiffusionModelEigenvectorSDE}

In this appendix, we discuss a diffusion model for the Eigenvector SDE. First, in section~\ref{Appendix:SDEO(n)} we present some preliminaries on SDEs on $\mathrm{O}(n)$. In section~\ref{Appendix:EigenvectorStratonovich} we show that the Stratonovich SDE in \eqref{EigenvectorStratonovichSDE} agrees with the \eqref{eq:SDE-eigenvectors}. In section~\ref{Appendix:EigenvectorTimeReversal} we establish the time-reversal formula for the eigenvector SDE, and in section~\ref{Appendix:EigenvectorLossfunction} we explain the corresponding loss function.

Foundational results on diffusion models on Riemannian manifolds were established by \cite{debortoliRiemannianScoreBasedGenerative2022}, treating the case of Brownian motions with a drift. Moreover, recently \cite{bertoliniDiffusionGenerativeModeling2025} discussed diffusion models for Lie groups acting on $\mathbb{R}^n$. However, a diffusion model for SDEs of the type of the \eqref{eq:SDE-eigenvectors} has not been studied in the literature so far, and in this appendix we develop initial results towards establishing diffusion models for more general SDEs on Lie groups.

We denote by $\mathrm{O}(n)$ the orthogonal group $$\mathrm{O}(n) = \{ X \in \mathrm{GL}_n(\R) \,:\, X^T = X^{-1}  \},$$ where $\mathrm{GL}_n(\R) = \{ X \in \mathbb{R}^{n \times n} \,:\, \det(X) \neq 0 \}$ is the group of invertible $n \times n$ matrices. The Lie algebra of $\mathrm{O}(n)$ is the tangent space at the identity and can be calculated to be $$\mathfrak{o}(n) = T_e\mathrm{O}(n) = \{ Z \in \R^{n \times n} \,:\, Z^T = -Z \}.$$ Moreover, it is then easily seen that the tangent space at $X \in \mathrm{O}(n)$ is given by $$T_X\mathrm{O}(n) = X\mathfrak{o}(n) = \{XZ \,: \, Z \in \mathfrak{o}(n) \}.$$

The key insight to implement an efficient diffusion model for \eqref{eq:SDE-eigenvectors} is to rewrite the SDE as 
an exponential update rule that makes use of the Lie algebra. To that end, we first rewrite the SDE as a
Stratonovich SDE on the Lie group $\mathrm{O}(n)$. Denote by $\mathfrak{o}(n) =T_{\mathrm{Id}}\mathrm{O}(n) = \{ A \in \mathbb{R}^{n\times n} \,:\, A^T = -A \}$ the Lie algebra of $\mathrm{O}(n)$, which is the tangent space of $\mathrm{O}(n)$ at the identity matrix. Let $E_{(\ell,k)} \in \mathfrak{o}(n)$ for $1\leq \ell < k\leq d$ be the matrix that is $1$ at the $(\ell,k)$-entry, $-1$ at $(k,\ell)$, and $0$ otherwise. We prove in Proposition~\ref{Prop:EigenvectorStratonovich} that the \eqref{eq:SDE-eigenvectors} $X(t) = (v_1(t), \dots, v_n(t)) \in \mathrm{O}(n)$ yields on $\mathrm{O}(n)$ the Stratonovich SDE
\begin{equation}\label{EigenvectorStratonovichSDE}
    dX(t) = \sqrt{\alpha}\sum_{1\leq\ell < k\leq n} \left(\frac{X(t)E_{(\ell,k)}}{\lambda_{k}(t)-\lambda_{\ell}(t)} \right) \circ \d W^{(\ell, k)}(t),
\end{equation}
which allows to derive the numerical update step for small $h > 0$, $X(t + h) \approx X(t)\exp(Z)$ \cite{marjanovicNumericalMethodsStochastic2016}, where $Z =\sqrt{\alpha h}\sum_{\ell < k} \frac{E_{(\ell,k)} \mathcal{N}^{(\ell,k)}(0,1)}{\lambda_{k}(t)-\lambda_{\ell}(t)} \in \mathfrak{o}(n)$, $\mathcal{N}^{(\ell,k)}(0,1)$ are independent samples of standard 1-dimensional Gaussians, and $\exp(Z) = \sum_{i = 0}^{\infty}\frac{A^i}{i!}$ is the matrix exponential.
For $h=\d t$, the normal samples $\sqrt{h}\mathcal{N}^{(\ell,k)}(0,1)$ become the Wiener increments $\d W_{\ell,k}$ in Sec.~\ref{sec:dysondiff-eigenvector}.

While SDEs of the form \eqref{EigenvectorStratonovichSDE} on $\mathrm{O}(n)$ have not been studied for diffusion models so far, we observe that the latter SDE is,
as explained in App.~\ref{Appendix:EigenvectorTimeReversal}, similar to Brownian motion on the Lie group $\mathrm{O}(n)$ as studied in \cite{debortoliRiemannianScoreBasedGenerative2022}. Indeed the Brownian motion $B^{\mathrm{O}(n)}(t)$ on $\mathrm{O}(n)$ written as a Stratonovich SDE is of the form $dB^{\mathrm{O}(n)}(t) = \sum_{\ell < k} (B^{\mathrm{O}(n)}(t) E_{(\ell,k)}) \circ \d W^{(\ell, k)}(t)$.
This similarity with Brownian motion on $\mathrm{O}(n)$ allows us to deduce a time reversal formula analogously to  \cite{debortoliRiemannianScoreBasedGenerative2022}. Indeed, denote by $\mathcal{X}(\mathrm{O}(n))$ the vector fields on $\mathrm{O}(n)$ and by $\nabla \ln p_{t} \in \mathcal{X}(\mathrm{O}(n))$ the gradient vector field of $\ln p_{t}$. We then need to transform the gradient vector field by the map $Q(s):\mathcal{X}(\mathrm{O}(n)) \to \mathcal{X}(\mathrm{O}(n))$ given for $V\in \mathcal{X}(\mathrm{O}(n))$ and $X \in \mathcal{O}(n)$ as $$(Q(s)V)(X) = \sum_{\ell < k} \frac{\langle V(X), X E_{(\ell,k)}\rangle}{(\lambda_{k}(t) - \lambda_{\ell}(t))^2}  X E_{(\ell,k)}.$$ The resulting time reversal formula is for $Y(s) = X(T-s)$ given by 
\begin{align}\label{EigenvectorInversionFormula}
    dY\!(s) =& \alpha(Q(T-s)\nabla \ln p_{T-s})(\bar\lambda, Y(s)) \d t  + \sqrt{\alpha}\sum_{\ell < k}\! \left(\!\frac{Y(s) E_{(\ell,k)}}{\lambda_k(T\!\!-\!\!s) \!-\! \lambda_{\ell}(T\!\!-\!\!s)} \!\right)\!\circ\! \d W^{(\ell,k)}(s).\!
\end{align}
In analogy to the forward dynamics above, this translates into an exponential update rule that in the notation of Eq.~\eqref{eq:dyson-timereversal} (i.e., $T-s\to t$ and time-reversal denoted by an overline, $Y(s)\to\bar X(t)$) reads $\bar X_{t + \d t} = \bar X_t\exp(\d \bar Z_t(\lambda,\bar X_t))$ with $\d \bar Z_t(\lambda,\bar X_t)=\alpha\nabla \ln p_t(\bar\lambda, \bar X_t) dt+ \sqrt{\alpha}\sum_{\ell < k}  \frac{ E_{(\ell,k)}}{\lambda_k(t) - \lambda_{\ell}(t)}  \d\bar W^{(\ell,k)}_t\in\mathfrak{o}(n)$ (see also algorithm 1 in  \citep{debortoliRiemannianScoreBasedGenerative2022} for related results for Brownian dynamics on a manifold).

\subsection{Preliminaries on SDEs on  \texorpdfstring{$\mathrm{O}(n)$}{O(n)}}\label{Appendix:SDEO(n)}

We first briefly recall some preliminary material on SDEs, particularly Stratonovich SDEs, on Lie groups. An excellent reference for this topic is \cite{hsu2002manifolds}, as well as the appendix in \cite{debortoliRiemannianScoreBasedGenerative2022}.

\paragraph{Vector Fields and Stratonovich SDEs on $\mathbb{R}^n$.} Recall that a smooth vector field $V$ on $\R^d$ is a map that smoothly assigns to each point a vector. Formally, we can therefore view $V$ as a smooth map $$V: \R^n \to \R^n, \quad \quad x \mapsto (V_1(x), \ldots , V_n(x)).$$ Given two smooth vector fields $V$ and $U$ on $\R^d$, we define the covariant derivative $\nabla_U V$ of $V$ with respect to $U$ to be the vector field 
\begin{align*}
    (\nabla_U V)(x) &= \frac{d}{dt}\bigg|_{t = 0} V(x + tU(x)) \\ &= \left(  \sum_{j = 1}^n U_j(x)\frac{\partial V_1}{\partial x_j}(x), \sum_{j = 1}^n U_j(x)\frac{\partial V_2}{\partial x_j}(x), \ldots ,  \sum_{j = 1}^n U_j(x)\frac{\partial V_n}{\partial x_j}(x) \right).
\end{align*}

Given $d \geq 1$, denote by $W^{1}, \ldots, W^{d}$ independent one-dimensional Brownian motions and let $V_1, \ldots , V_d$ be vector fields on $\R^d$. Then the Stratonovich SDE $$dX(t) = \sum_{i} V_{i}(X_t) \circ dW^i(t)$$ denotes the Ito SDE $$dX(t) = \frac{1}{2}\left(\sum_{i} (\nabla_{V_i} V_i)(X(t)) \right)dt + \sum_{i} V_i(X(t)) dW^i(t).$$ Just to have a concrete example in mind: The $n$-dimensional Brownian motion can be written by taking $d = n$ and considering the constant vector fields $V_i(x) = e_i$ for $e_i$ the standard basis vectors.

\paragraph{Vector Fields and Stratonovich SDEs on $\mathrm{O}(n)$.}

In this subsection, we consider the concrete case of a Stratonovich SDE driven by left invariant vector fields on $\mathrm{O}(n)$. Indeed, a left invariant vector field $V$ is determined by a Lie algebra element $z \in \mathfrak{o}(n)$ and given  by $$V(X) = Xz \in T_X\mathrm{O}(n).$$ We now exploit that $\mathrm{O}(n)$ is embedded in $\R^{n\times n}$ so we can write a Lie group SDE as an SDE on $\R^{n\times n}$. Indeed, let $V_1, \ldots , V_d$ be left invariant vector fields $V_i(X) = Xz_i$ and consider the Lie group SDE $$dX(t) = \sum_{i} (X(t) z_i) \circ dW^i(t).$$ Based on Section 1.2 in \citet{hsu2002manifolds}, we can write the above SDE as an Ito SDE on $\mathbb{R}^{n \times n}$ with the terms 
\begin{equation}\label{O(n)StratonovicasR^n}
    dX(t) = \frac{1}{2}\sum_{i} (X(t)z_i^2) dt + \sum_i (X(t)z_i) dW^i(t).
\end{equation}

\subsection{Writing the Eigenvector SDE as a  \texorpdfstring{$\mathrm{O}(n)$}{O(n)} Stratonovich SDE}\label{Appendix:EigenvectorStratonovich}

\begin{proposition}\label{Prop:EigenvectorStratonovich}
The $\mathrm{O}(n)$ Stratonovich SDE from \eqref{EigenvectorStratonovichSDE} is the same as \eqref{eq:SDE-eigenvectors} viewed as an SDE on $\R^{n \times n}$.
\end{proposition}

\begin{proof}
    This follows by a direct calculation using equation \eqref{O(n)StratonovicasR^n} for time-dependent vector fields. Recall that $E_{(\ell,k)} = e_{\ell}e_k^T - e_ke_{\ell}^T$ for $\ell < k$, where $e_i$ is the standard basis vector of $\R^n$ viewed as a row vector. Then denoting by $e_{ii}$ the diagonal matrix that is $1$ at the $(i,i)$ entry and zero everywhere else, it holds that $$E_{(\ell,k)}^2 = - (e_{\ell\ell} + e_{kk}).$$
   Therefore the $\mathrm{O}(n)$ Stratonovich SDE from \eqref{EigenvectorStratonovichSDE} is the Ito SDE on $\R^{n \times n}$ given by $$dX(t) = -\frac{\alpha}{2} \sum_{\ell < k} \frac{X(t)(e_{\ell\ell} + e_{kk})}{(\lambda_k(t) - \lambda_{\ell}(t))^2} + \sqrt{\alpha}\sum_{\ell < k} \frac{X(t)E_{(\ell,k)}}{\lambda_k(t) - \lambda_{\ell}(t)} dW_{(\ell,k)}(t).$$ By writing $X(t) = (v_1(t) \dots v_n(t))$ as row vectors, the last equation is easily seen to be exactly the \eqref{eq:SDE-eigenvectors}.
\end{proof}

\subsection{Deduction of Time Reversal Formula}\label{Appendix:EigenvectorTimeReversal}

Recall that the Brownian motion on $\mathrm{O}(n)$ is given by the $\mathrm{O}(n)$ Stratonovich SDE $$dB^{\mathrm{O}(n)}(t) = \sum_{\ell < k} (B^{\mathrm{O}(n)}(t)E_{(\ell,k)}) \circ dW^{(l,k)}(t).$$ We next consider the time-scaled Brownian motion on $\mathrm{O}(n)$ given by $$X(t) = g(t)B^{\mathrm{O}(n)}(t)$$ for some scalar function $g(t)$.

As in \cite{debortoliRiemannianScoreBasedGenerative2022} we denote by $\nabla \ln p_{T-s}$ the gradient vector field of $\ln p_{T-s}$. It is shown (or rather stated that it can be shown) in Theorem 1 in \citet{debortoliRiemannianScoreBasedGenerative2022} that the time reversal of the Brownian motion $Y(s) = X(T-s)$ satisfies the SDE $$dY(s) = g(T-s)^2\nabla \ln p_{T-s}(Y(s)) \, dt + g(T-s)dB^{\mathrm{O}(n)}(s).$$ 
The time reversal formula \eqref{EigenvectorInversionFormula} is obtained by adapting the above result to our SDE \eqref{EigenvectorStratonovichSDE}, where the vector fields are scaled by the factors $\frac{\sqrt{\alpha}}{\lambda_k(t) - \lambda_{\ell}(t)}$. The final result is stated in  \eqref{EigenvectorInversionFormula}.

\subsection{Loss Function}\label{Appendix:EigenvectorLossfunction}

As for the eigenvalue SDE, we have no direct access to the probability measures $p_t(x_t|x_0)$. We therefore employ a loss analogous to the one used to train the eigenvalues and explained in App.~\ref{sec:app:loss}.

To approximate the score, we observe that similarly to Equation~(8) in \citet{debortoliRiemannianScoreBasedGenerative2022}, $$\lim_{t \to s} (t-s)(\nabla \ln p_{t|s})(X_t|X_s) \approx -\Sigma(t,s)\cdot \ln_{\mathrm{O}(n)}(X_t^TX_s) \quad \text{ for } \quad \Sigma(t,s) = \mathrm{diag}\left(\frac{\alpha(t-s)}{(\lambda_{k}(s) - \lambda_{\ell}(s))^2}\right),$$ where $\ln_{\mathrm{O}(n)}:\mathrm{O}(n) \to \mathfrak{o}(n)$ is the matrix logarithm given for $X \in \mathrm{O}(n)$ by $\ln_G(X) = \sum_{k = 1}^{\infty} \frac{(-1)^{k + 1}}{k} (X-\mathrm{Id})^k$ and  we view $\Sigma(t,s)$ as a diagonal matrix with respect to the standard basis $E_{(\ell,k)}$ of $\mathfrak{o}(n)$. 

As discussed in the main part, for a small $h>0$, we have the approximation  $$X(t + h) \approx X(t)\exp(Z) \quad\quad \text{ with } \quad\quad Z =\sqrt{\alpha h}\sum_{\ell < k} \frac{E_{(\ell,k)} \mathcal{N}^{(\ell,k)}(0,1)}{\lambda_{k}(t)-\lambda_{\ell}(t)} \in \mathfrak{o}(n),$$ where $\mathcal{N}^{(\ell,k)}(0,1)$ are independent samples of standard 1-dimensional Gaussians and $\exp(Z) = \sum_{i = 0}^{\infty}\frac{Z^i}{i!}$ is the matrix exponential.

Assume now that we have $N$ discretizations of the interval $[0,T]$ for some $T > 0$ that we denote for $1 \leq r \leq N$ as $$0 = t_0^{(r)} < t_1^{(r)} < t_2^{(r)} < \ldots < t_{k^{(r) -1 }}^{(r)} < t_{k^{(r)}}^{(r)} = T.$$ For those we have $N$ sampled paths of our SDE from \eqref{EigenvectorStratonovichSDE} denoted for $1 \leq r \leq N$ as $$X_0^{(r)}, X_{t_1^{(r)}}^{(r)}, X_{t_2^{(r)}}^{(r)}, \ldots X_{t_{k^{(r)}-1}^{(r)}}^{(r)}, X_T^{(r)}  \quad\quad \text{with Lie algebra increments } \quad\quad Z_0^{(r)}, Z_{t_1^{(r)}}^{(r)}, Z_{t_2^{(r)}}^{(r)} \ldots Z_{t_{k^{(r)}-1}^{(r)}}^{(r)}$$ so that for all $0 \leq i \leq k^{(r)}-1$ it holds that $$X_{t^{(r)}_{i + 1}}^{(r)} = X_{t^{(r)}_{i}}^{(r)}\exp(Z^{(r)}_{t^{(r)}_{i}}) \quad\quad \text{ and therefore } \quad\quad  -Z^{(r)}_{t^{(r)}_{i}} = \ln_{\mathrm{O}(n)}((X_{t^{(r)}_{i + 1}}^{(r)})^T X_{t^{(r)}_{i}}^{(r)}).$$ 

By the last equation, it therefore follows that $$\nabla \ln p_{t|s}(X_{t^{(r)}_{i + 1}}^{(r)}|X_{t^{(r)}_{i}}^{(r)}) \approx \frac{\Sigma(t_{i + 1}^{(r)}, t_{i}^{(r)})Z^{(r)}_{t^{(r)}_{i}}}{t_{i + 1}^{(r)}- t_{i}^{(r)}}.$$

Analogously to App.~\ref{sec:app:loss}, we therefore want to optimize the loss 
 $$\frac{1}{N}\sum_{r = 1}^N \sum_{i = 1}^{k^{(r)}-1} \frac{t_{i + 1}^{(r)}- t_{i}^{(r)}}{T} \bigg|\bigg|s(X_{t_{i + 1}^{(r)}}^{(r)}, t_{i + 1}^{(r)}) -  \frac{\Sigma(t_{i + 1}^{(r)}, t_{i}^{(r)})Z^{(r)}_{t^{(r)}_{i}}}{t_{i + 1}^{(r)}- t_{i }^{(r)}}\bigg|\bigg|,$$ where the norm is the $L^2$-norm on the Lie algebra $\mathfrak{o}(n)$.

\subsection{Invariance on the Lie algebra}\label{sec:app:eigenvecs:lie-algebra-invariance}

Since permutations of adjacency matrices have the same spectrum, the permutation invariance of graphs did not have to be considered when learning purely on spectral information, see, e.g., Fig.~\ref{fig:mainfig}. 
This changes for the eigenvectors in $X(t) = (v_1(t), \dots, v_n(t)) \in \mathrm{O}(n)$. While the column indices are fixed since each eigenvector $v_i(t)$ corresponds to a fixed eigenvalue $\lambda_i(t)$, the row indices change upon permutation of the adjacency matrix. Therefore, the desired learning algorithm must be invariant under permutations $X\mapsto PX$, $Y\mapsto PY$, where $P$ is a permutation matrix swapping the rows of $X$ and $Y$, respectively. The backwards dynamics is mathematically described and numerically implemented as 
$Y(t - \d t) = Y(t)\exp(\d Z_Y(\lambda,Y,t))$, which under permutation $Y\mapsto PY$ reads $PY(t - \d t) = PY(t)\exp(\d Z_Y(\lambda,PY,t))$. At the same time, multiplication of the former equation from the left side by $P$ implies $PY(t - \d t) = PY(t)\exp(\d Z_Y(\lambda,Y,t))$. Since $PY$ is invertible, these equations imply $\exp(\d Z_Y(\lambda,Y,t))=\exp(\d Z_Y(\lambda,PY,t))$. Since $\d Z_Y(\lambda,Y,t)\to 0$ as $\d t\to 0$, we can deduce that $\d Z_Y(\lambda,Y,t)=\d Z_Y(\lambda,PY,t)$, i.e., that the increment in the exponent (meaning the change phrased in terms of the Lie algebra) is permutation invariant. Since we learn the score on these Lie algebra increments, see Eq.~\eqref{EigenvectorInversionFormula}, this implies that the score is invariant under permutations $Y\to PY$. Crucially, this differs from the typical time reversal score as in Eq.~\eqref{eq:backward}, which would be permutation equivariant, $s(PYP^T,t)=Ps(Y,t)P^T$ (note that on the full matrix, the permutation changes rows and columns, i.e., $Y \to PYP^T$). We instead have invariance $s(PY,t)=s(Y,t)$, and no equivariance $s(PY,t)\ne Ps(Y,t)$. Note that this somewhat unexpected absence of equivariance is due to the choice to parametrize the score in the Lie algebra instead of the Lie group which is seen from the form $Y(t - \d t) = Y(t)\exp(\d Z_Y(\lambda,Y,t))$. Here, when updating $Y$ in a permuted frame, all the information about the permutation is already contained in the $Y(t)$ which enables that the exponent (containing the score) is unaffected by the permutation. This invariance opens the learning task to Set Transformers.

\subsection{Sign symmetry} \label{sec:app:eigenvecs:sign-symmetry}
We now show that the score obeys a sign symmetry law, which allows to reduce the space of the learning target by a factor of $2^d$. In particular, we show that the score obeys $\nabla \ln p_t(\bar \lambda, \bar X_t F)=F\nabla \ln p_t(\bar \lambda, \bar X_t)F$ for any diagonal matrix $F$ with entries $\pm 1$ representing sign flips.

To see this, compare the sign-flipped dynamics $\bar X_{t + \d t}F = \bar X_tF\exp(\d \bar Z_t(\lambda,\bar X_tF))$, to the original dynamics multiplied by $F$, i.e., $\bar X_{t + \d t}F = \bar X_t\exp(\d \bar Z_t(\lambda,\bar X_t))F$. Using $F=F^{-1}$, the latter yields $\bar X_{t + \d t}F = \bar X_tF\exp(F\d \bar Z_t(\lambda,\bar X_t)F)$, which since $\bar X_tF$ is invertible and $\exp$ can be linearized for small increments, implies $\d \bar Z_t(\lambda,\bar X_tF)=F\d \bar Z_t(\lambda,\bar X_t)F$. Since Wiener increments are symmetric in the sign, this is equivalent to $\nabla \ln p_t(\bar \lambda, \bar X_t F)=F\nabla \ln p_t(\bar \lambda, \bar X_t)F$.

Thus, the learning target is effectively reduced by a factor of $2^d$ -- i.e.\ an exponential reduction in the dimension. See App.~\ref{sec:app:eigenvecs-numerical-experiment} for a clear practical discussion on how this can be precisely exploited algorithmically.

\subsection{Numerical illustration}\label{sec:app:eigenvecs-numerical-experiment}
We demonstrate in \cref{fig:lie} our numerical scheme for the time reversal on $SO(3)$, the highest dimension fully representable in a plot. 
\begin{figure}[th]
    \centering
    \includegraphics[width=0.5\linewidth]{figs/eigenvectors-succesfully-generated.pdf}
    \caption{Numerical demonstration on $SO(3)$: Starting from the invariant distribution (left), a learned distribution is being generated (right) using \Cref{EigenvectorInversionFormula}.}
    \label{fig:lie}
\end{figure}

For higher dimensions, we note that for learning \eqref{eq:SDE-eigenvectors}, the $\lambda$-dependent coefficients pose a challenge. Yet, the SDE retains all its advantages such as staying on the orthogonal manifold $O(n)$, as well as the permutation invariance and sign symmetry properties when that prefactor is dropped, which we do in the following.
Our aim here is to show that, given that spectra have been learned, the eigenvectors are indeed accessible. In particular, we show that in the WL-equivalence dataset presented in \Cref{fig:wl-demonstration}, the eigenvector diffusion on $O(n)$ can be perfectly learned, taking the spectra from an oracle machine.
As proven in \Cref{lem:permutation-invariance-lie-algebra}, we open the learning problem to our best knowledge for the first time to the field of permutation \emph{in}variant architectures, such as Set transformers \cite{lee2019set}.
In particular, we train a set transformer to learn the score of the time reversal described in \ref{sec:dysondiff-eigenvector}, and present the results in Table~\ref{tab:lie-diffusion-best}. We also illustrate these results visually by showing in Fig.~\ref{fig:graphs-we-generated} explicitly the generated graphs. Note that generation of $n=1000$ such graphs takes only $17$ seconds.

\paragraph{Eigenvector score architecture.}
For the conditional eigenvector score learning of the Lie Algebra we use a row-invariance Set transformer~\citep{lee2019set}.
The model takes as input a diffusion time \(t\), the Laplacian spectrum
excluding the fixed first Laplacian eigenvalue, and an eigenvector matrix
\(X \in O(d)\). The output is a vector of
strict upper-triangular Lie algebra coordinates in \(\mathfrak{o}(d)\),
corresponding to the \(d(d-1)/2\) pairwise eigenvector rotation directions.

We treat the rows of \(X\) as an unordered set of tokens. Each row
is embedded into a width-512 token representation. The time variable is encoded
using Fourier features with embedding dimension \(256\) and frequency range
\([1,100]\). The spectrum and
time embeddings are combined into a width-512 conditioning vector. The row-token
trunk consists of three conditioned self-attention blocks with 16 attention
heads. Conditioning is injected by FiLM-style affine modulation
\citep{perezFilm18} before both the attention and token-MLP sublayers. We use
scaled SiLU activations.

For every pair \(k<\ell\), we form pair features from \(C_k\), \(C_\ell\),
their difference, absolute difference, elementwise product, the corresponding
spectral features
\[
    \lambda_k,\quad \lambda_\ell,\quad
    \lambda_k-\lambda_\ell,\quad |\lambda_k-\lambda_\ell|,
\]
and a global context vector. The global context is computed from the
conditioning vector and from row-token mean, variance, and smooth-max pooled
statistics. A width-512 pair MLP of depth 1 maps each pair feature to the
corresponding Lie algebra coordinate.

To apply our proven sign symmetry (see App.~\ref{sec:app:eigenvecs:sign-symmetry}), we proceed as follows: Eigenvector column signs are first
put into a deterministic canonical gauge, the scalar pair head is evaluated in
that gauge, and the output is transformed back to the original gauge using the corresponding sign flips, as detailed in App.~\ref{sec:app:eigenvecs:sign-symmetry}. This
enforces the correct equivariance under eigenvector sign flips while preserving
the expressivity of the unconstrained pair head, and reduces the learning problem by a factor of $2^d$ (see App.~\ref{sec:app:eigenvecs:sign-symmetry}).

As for the spectra, we use EMA (exponential moving averages) on the trained models. 

\paragraph{Training setup.}

We train on the time interval $[0,T]$ and use a fixed learning
grid with base spacing \(\d t = 0.05\) and refinement factors

\begin{center}
\begin{tabular}{l|l}
time interval & factor \\
\hline
$[0,0.25]$ & $0.03125$ \\
$[0.25,0.5]$ & $0.0625$ \\
$[0.5,1]$ & $0.125$ \\
$[1,2]$ & $0.25$ \\
$[2,3]$ & $0.5$ \\
$[3,8]$ & $1.0$
\end{tabular}
\end{center}

Thus, the earliest grid spacing is \(0.0015625\). We simulate 4000 forward
paths per epoch and use all fixed-grid intervals in the eigenvector loss. The
training batch size is 10, and the model is trained for 80 epochs. The reported model in \Cref{tab:lie-diffusion-best} used EMA with $\alpha = 0.95$ and averaged over the models from epochs $78$ to $80$.

Optimization uses AdamW through Optax. For the eigenvector model, we use global
gradient clipping at norm \(0.5\), initial learning rate \(7.5\cdot 10^{-4}\),
final learning rate \(1.2\cdot 10^{-4}\), weight decay \(10^{-5}\), and a
warmup fraction of \(0.07\) with a cosine decay schedule.

\subsubsection{Compute Resources}\label{sec:app:lie:resources}

\begin{figure}[t!!]
    \centering
    \includegraphics[width=\linewidth]{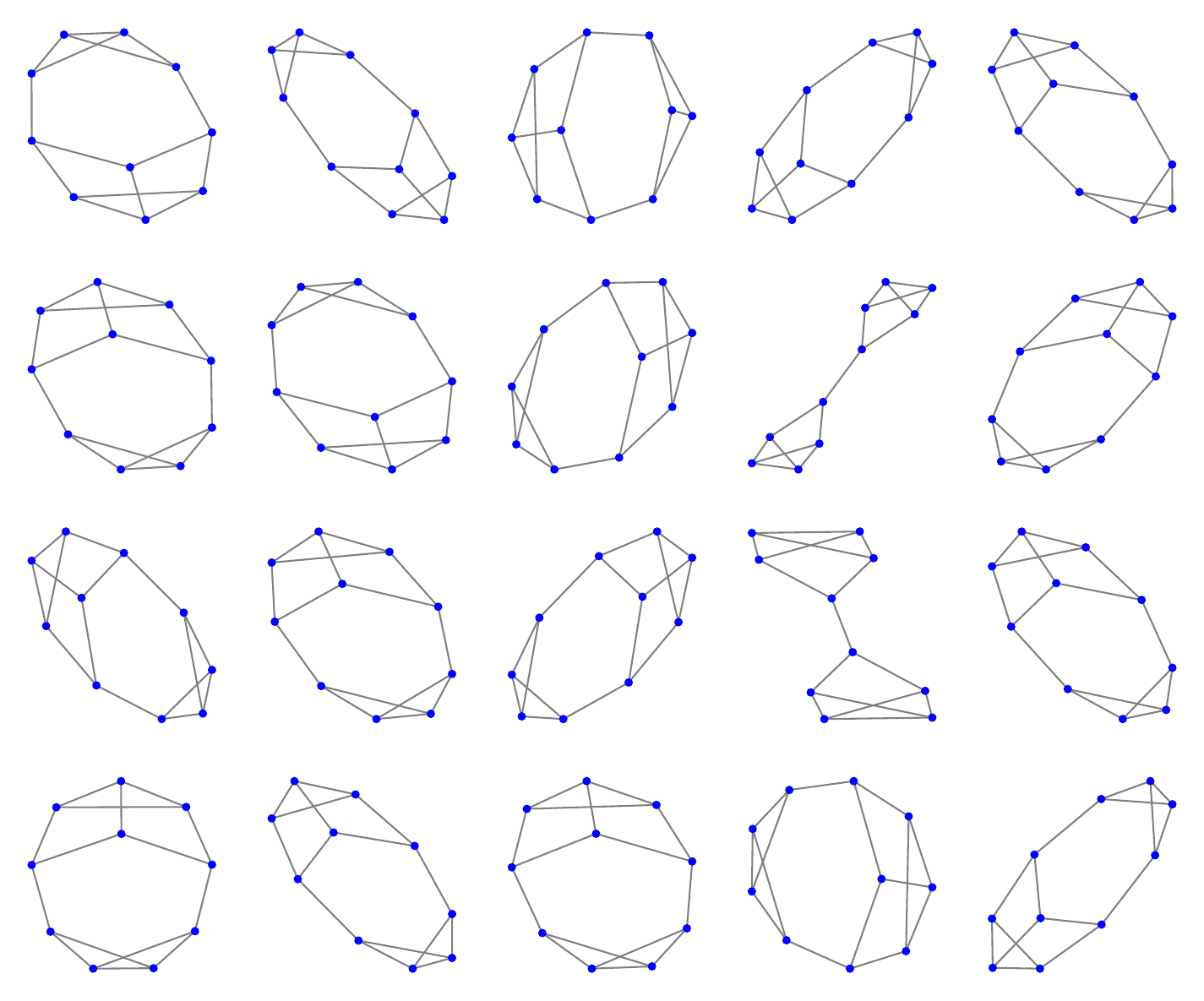}
    \caption{Graphs generated using the Lie Group diffusion: Each generated graph is isomorphic to one of the WL equivalent graphs, and as evident from \Cref{tab:lie-diffusion-best}, it learns the distribution very well. Sampling of $1{,}000$ such graphs takes $17$ seconds on an A100 GPU.}
    \label{fig:graphs-we-generated}
\end{figure}

On one $A100$ gpu, Sampling of $1000$ graphs from our model takes $17$ seconds. Training of the eigenvector model takes $20$ minutes. We trained in total $5$ models with different seeds and report their average and empirical standard deviation in Table~\ref{tab:app:our-eigenvector-lie-group-is-best}. Consistent with our discussion of the challenges that WL-equivalent graphs pose to GNNs, we see that models like EDP-GNN are very good at learning the degree, since all graphs in the same equivalence class have the same degree distribution. Indeed, all graphs of the WL-Bimodal dataset have the same degree distribution. However, distances which capture more complexity of the graph like Clustering Coefficient, Spectral or Orbit MMD distance show clearly the advantage of our proposed spectrum-first method, and they show evidently that our Lie group diffusion learns accurately. We depict in Fig.~\ref{fig:graphs-we-generated} graphs that were generated using the Lie group diffusion.

\begin{table}[t]
\centering
\vspace{0.3cm}
\caption{MMD metrics of generated graphs trained on the WL-Bimodal dataset. Consistently with the WL-blindness of GNNs, models like EDP-GNN are very good at learning the WL-equivalence class of $3$-regular graphs (degree distance is $0$), but they get outperformed by the DyDM + Lie group diffusion approach on the other distances. Top table: Average of multiple training runs, bottom table: empirical standard deviation for completeness.}
\label{tab:lie-diffusion-best}
\begin{tabular}{|l|l|l|l|l|}
\hline
Mean Distances & Degree & Clustering Coefficient & Spectral & Orbit \\
\hline
\textbf{Lie Diffusion Eigenvectors} & 0.00013 & \best{ 0.00041 } & \best{ 0.00011 } & \best{ 0.00001 } \\
EDP-GNN & \best{ 0.00000 } & 0.16590 & 0.05011 & 0.00189 \\
GDSS & \almostbest{ 0.00000 } & 0.33342 & 0.12303 & 0.00025 \\
ConGress & 0.07759 & 0.45267 & 0.15022 & 0.01024 \\
DiGress (no trick) & 0.37712 & 0.70940 & 0.23426 & 0.04024 \\
DiGress (trick) & 0.00000 & \almostbest{ 0.00189 } & \almostbest{ 0.00050 } & \almostbest{ 0.00004 } \\
\hline
\end{tabular}
\vspace{0.6cm}

\begin{tabular}{|l|l|l|l|l|}
\hline
Empirical Standard Deviations & Degree & Clustering Coefficient & Spectral & Orbit \\
\hline
\textbf{Lie Diffusion Eigenvectors} & 6.3e-05 & 6.5e-05 & 4.1e-05 & 9.7e-06 \\
EDP-GNN & 6e-08 & 0.11 & 0.035 & 0.0007 \\
GDSS & 6.5e-08 & 0.028 & 0.009 & 0.00011 \\
ConGress & 0.17 & 0.17 & 0.053 & 0.018 \\
DiGress (no trick) & 0.004 & 0.0023 & 0.0028 & 0.011 \\
DiGress (trick) & 3.3e-07 & 0.0016 & 0.00042 & 2.6e-05 \\
\hline
\end{tabular}\label{tab:app:our-eigenvector-lie-group-is-best}
\end{table}

\newpage

\end{document}